\documentclass{article}

\PassOptionsToPackage{numbers, sort, compress}{natbib}

 \usepackage[preprint]{neurips_2023}

\usepackage[utf8]{inputenc} 
\usepackage[T1]{fontenc}    
\usepackage[colorlinks=true,urlcolor=blue]{hyperref} 
\usepackage{url}            
\usepackage{booktabs}       
\usepackage{amsfonts}       
\usepackage{nicefrac}       
\usepackage{microtype}      
\usepackage{xcolor}         

\usepackage{multirow}
\usepackage{multicol}
\usepackage{amsmath}
\usepackage{graphicx}
\usepackage{graphics}
\usepackage{tablefootnote}
\usepackage{wrapfig}
\usepackage{ifthen}
\usepackage{enumitem}
\usepackage{alphalph}
\usepackage{adjustbox}
\usepackage{scalefnt}
\usepackage{colortbl}
\usepackage{xcolor}
\usepackage{subcaption}
\usepackage{caption}
\usepackage{threeparttable}
\definecolor{VeryLightPink}{RGB}{255, 248, 248}
\definecolor{VeryLightBlue}{RGB}{248, 248, 255}
\definecolor{VeryLightGreen}{RGB}{248, 255, 248}

\title{StreetSurf: Extending Multi-view Implicit Surface Reconstruction to Street Views}

\author{
\parbox{\linewidth}{
	\centering
	Jianfei Guo$^{1}$ 
	\quad Nianchen Deng$^{1}$ 
	\quad Xinyang Li$^{1}$ 
	\quad Yeqi Bai$^{1}$ 
	\quad Botian Shi$^{1}$ \\
	Chiyu Wang$^{2}$ 
	\quad Chenjing Ding$^{2}$ 
	\quad Dongliang Wang$^{1,2}$ 
	\quad Yikang Li$^{1}$\thanks{Corresponding author} 
}
\vspace{8pt}\\
$^{1}$Shanghai Artificial Intelligence Laboratory \qquad 
$^{2}$Sensetime Research
\vspace{8pt}\\
\href{https://ventusff.github.io/streetsurf_web/}{https://ventusff.github.io/streetsurf\_web}
}

\begin{document}

\maketitle

\begin{abstract}


We present a novel multi-view implicit surface reconstruction technique, termed \textbf{StreetSurf}, that is readily applicable to street view images in widely-used autonomous driving datasets, such as Waymo-perception sequences, 
without necessarily requiring LiDAR data. 
As neural rendering research expands rapidly, its integration into street views has started to draw interests.
Existing approaches on street views either mainly focus on novel view synthesis with little exploration of the scene geometry, or rely heavily on dense LiDAR data when investigating reconstruction. Neither of them investigates multi-view implicit surface reconstruction, especially under settings without LiDAR data.
Our method extends prior object-centric neural surface reconstruction techniques to address the unique challenges posed by the unbounded street views that are captured with non-object-centric, long and narrow camera trajectories.
We delimit the unbounded space into three parts, close-range, distant-view and sky, with aligned cuboid boundaries, and adapt cuboid/hyper-cuboid hash-grids along with road-surface initialization scheme for finer and disentangled representation.
To further address the geometric errors arising from textureless regions and insufficient viewing angles, we adopt geometric priors that are estimated using general purpose monocular models.
Coupled with our implementation of efficient and fine-grained multi-stage ray marching strategy, we achieve state of the art reconstruction quality in both geometry and appearance within only one to two hours of training time with a single RTX3090 GPU for each street view sequence.
Furthermore, we demonstrate that the reconstructed implicit surfaces have rich potential for various downstream tasks, including ray tracing and LiDAR simulation.

\end{abstract}

\section{Introduction}
\label{sec:intro}




Neural rendering has seen rapid growth in recent years since the emergence of the NeRF~(Neural Radiance Field)~\citep{mildenhall2021nerf}. 
With a collection of posed images as inputs or supervisions, NeRF is capable of providing high-fidelity novel view synthesis results with simple yet effective assumptions and structures, which also makes it a flexible technique that sparks a flourishing of following research.
Recent research~\citep{ost2021neural,tancik2022blocknerf,rematas2022urban,turki2023suds,wang2023fegr,xie2023street-nerf,li2022read} starts to apply NeRFs to street views or large-scale urban scenes. 
Among them, there are some works~\citep{ost2021neural,KunduCVPR2022PNF,fu2022panoptic,xie2023street-nerf} that focus on the relatively smaller-scale scenes readily available in autonomous driving datasets, such as KITTI~\citep{geiger2012we}, Waymo Open Dataset~\citep{waymo}, or NuScenes~\citep{caesar2020nuscenes}.
Reconstruction and rendering in these datasets are helpful for various downstream tasks including data augmentation, 3D/multi-modal perception and AR/VR.
However, existing approaches on street views either mainly focus on novel view synthesis~\citep{ost2021neural,wang2023f2nerf,xie2023street-nerf,li2022read} with little exploration of the scene geometry, or rely heavily on dense LiDAR data when investigating reconstruction~\citep{rematas2022urban,wang2023fegr}. Neither of them investigates multi-view implicit surface reconstruction, especially under settings without LiDAR data.

3D reconstruction from street views are much more challenging compared to previous works under bounded or object-centric multi-view settings~\citep{wang2021neus,yariv2021volume,oechsle2021unisurf}, since  
scenes in street views are typically unbounded and captured with non-object-centric, long and narrow camera trajectories. 


\begin{wrapfigure}{r}{0.45\textwidth}
\vspace{-12pt}
\centering
\begin{subfigure}[b]{0.42\textwidth}
\includegraphics[width=\textwidth]{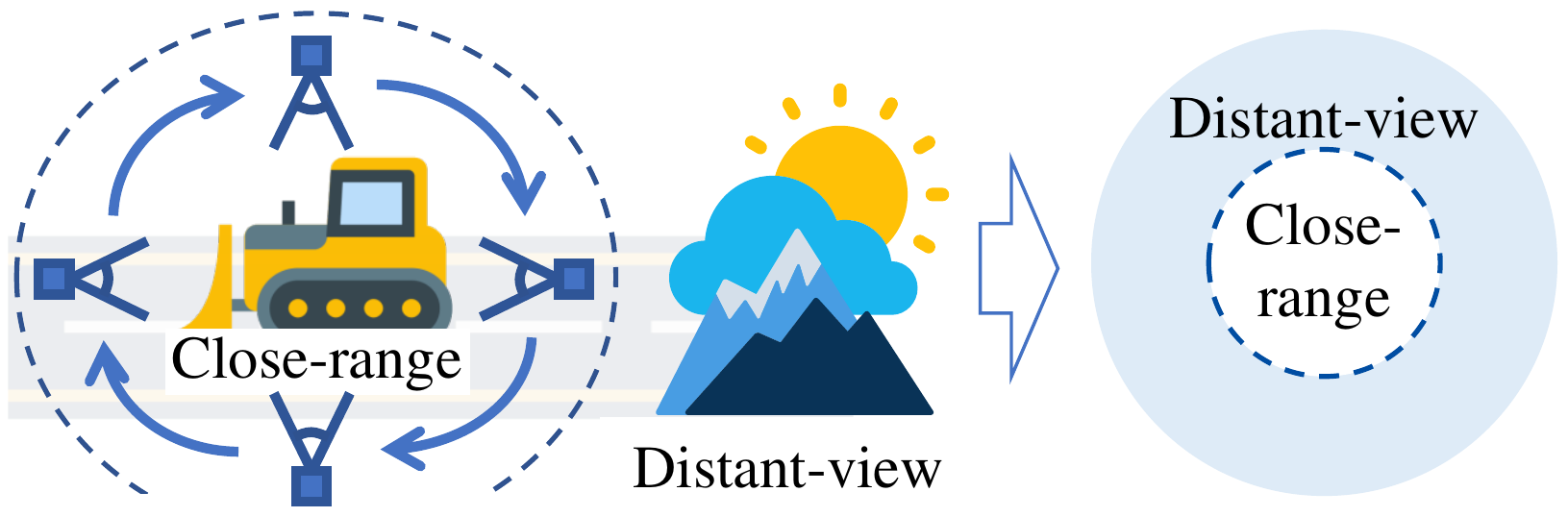}
\caption{NeRF-360: Unbounded scene with object-centric camera trajectories. Using spherical bounds and spherical/cubic model.}
\label{fig:cuboid_space:360}
\end{subfigure}
\par
\vspace{0.1cm}
\begin{subfigure}[b]{0.42\textwidth}
\includegraphics[width=\textwidth]{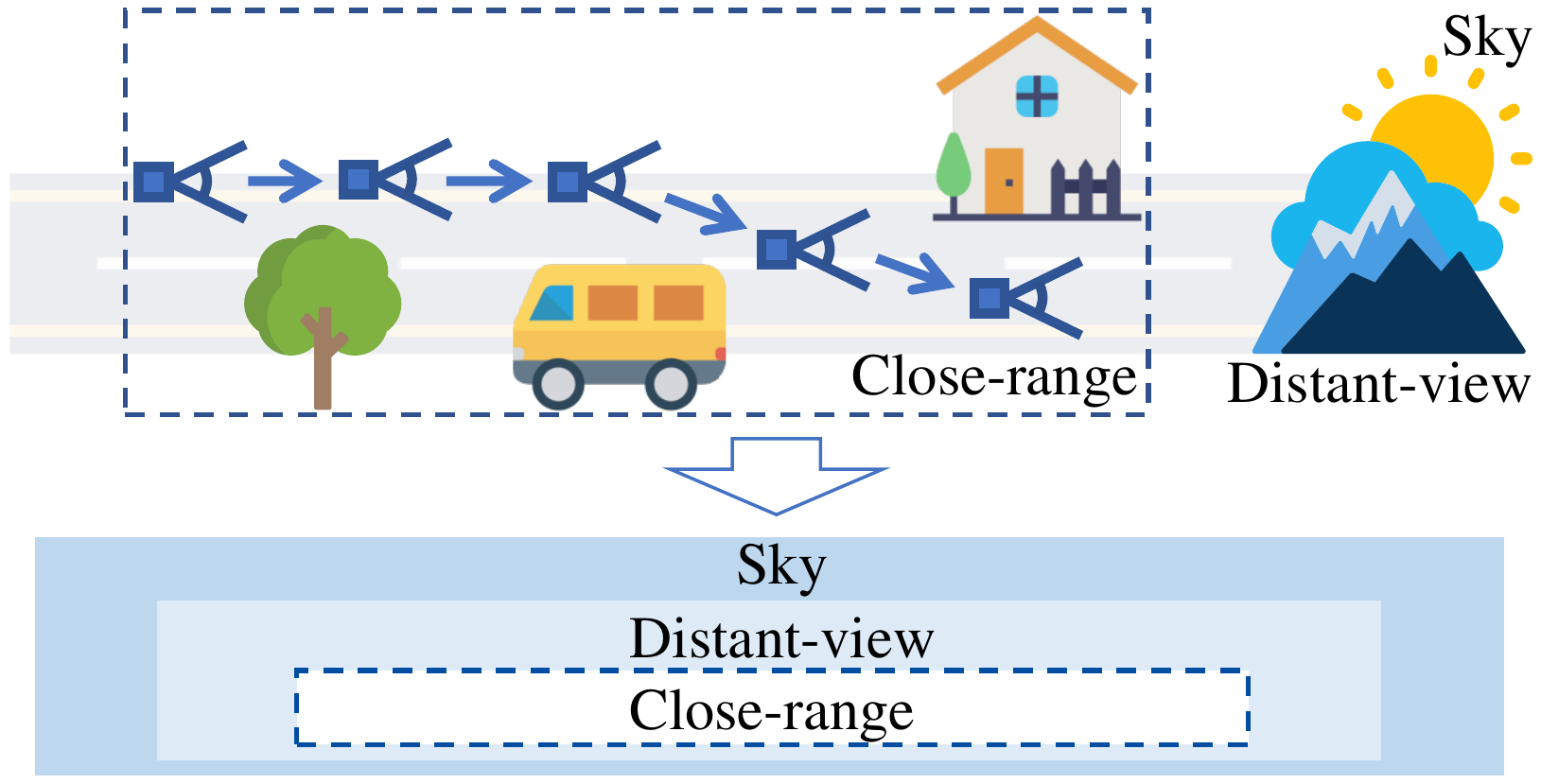}
\caption{Street views: Unbounded scene captured with non-object-centric, long and narrow camera trajectories. We propose to use aligned cuboid bounds and cuboid model.}
\label{fig:cuboid_space:street_views}
\end{subfigure}
\caption{The challenges of street views}
\label{fig:cuboid_space}
\vspace{-12pt}
\end{wrapfigure}

To deal with unbounded scenes, previous research~\citep{wang2021neus,zhang2020nerf++,barron2021mipnerf,barron2022mipnerf360} leverage inverse-spherical warping for object-centric camera settings, as shown in Fig.~\ref{fig:cuboid_space:360}. 
NeuS~\citep{wang2021neus} and other concurrent works~\citep{oechsle2021unisurf,yariv2021volume} introduce this idea to multi-view implicit surface reconstruction, in which they represent the close-range scene with SDFs~(Signed Distance Functions) and the distant-view with NeRF++~\citep{zhang2020nerf++}.
However, when it comes to non-object-centric settings like street views, inverse-spherical warping becomes wasteful of capacity and thus no longer suitable.
This is evidenced by the very recent works~\citep{wang2023f2nerf,meuleman2023progressively}, who subsequently introduce perspective warping or progressive warping that warp the full space according to different camera trajectories to better allocate network capacity.

However, we have found that these full-space warping strategies are not well-suited for geometry represented by SDFs as they break the definition of distance fields and sphere tracing. 
In addition, although effective for the novel view synthesis task, modeling the entire space with one single model does not match the different needs of different viewing spaces when it comes to the multi-view reconstruction task. 
Specifically, the close-range space requires complete and accurate surface geometry, whereas the distant-view and sky only need their appearance to be properly fit and should not interfere with the geometry of the close-range model.
For this purpose, we propose to delimit the close-range and distant-view spaces with long and narrow cuboid boundaries that are aligned with the camera trajectories, as shown in Fig.~\ref{fig:cuboid_space:street_views}. 
Then we assign a cuboid NeuS model and a hyper-cuboid NeRF++ model to the close-range and  distant-view spaces, respectively. 
If an optional set of sky masks are provided, we further distinguish the sky part out of the distant-view by using a directional MLP.

Following the above discussions, we are led to the disentanglement of close-range and distant-view models, which is an unsupervised and ill-posed problem with almost no constraints. 
Its impact is negligible in setups with object-centric camera trajectories \citep{wang2021neus,zhang2020nerf++}.
However, in our non-object-centric multi-view setup, lack of sufficiently varying view angles often leads to ambiguous solutions as two models are in competition with each other.
To address this issue, we first examine different initialization schemes of the SDF field in our close-range model. 
We have found that enclosed initial shapes like inside-out spheres~\citep{guo2022neural,Yu2022MonoSDF} or ellipsoids/capsules can hinder the disentanglement since geometries in street-view scenes are usually open and not enclosed. 
Therefore, we propose a novel road-surface initialization scheme that pre-train the SDF to be roughly aligned with the road surface.
We also apply an entropy regularization to further encourage crisp disentanglement.

\begin{figure}[htbp]
\centering
\setlength\tabcolsep{1pt}
\begin{tabular}{cccc}

\raisebox{0.05\textwidth}[0pt][0pt]{\rotatebox[origin=c]{90}{\small{Images}}}
& \includegraphics[width=0.32\textwidth]{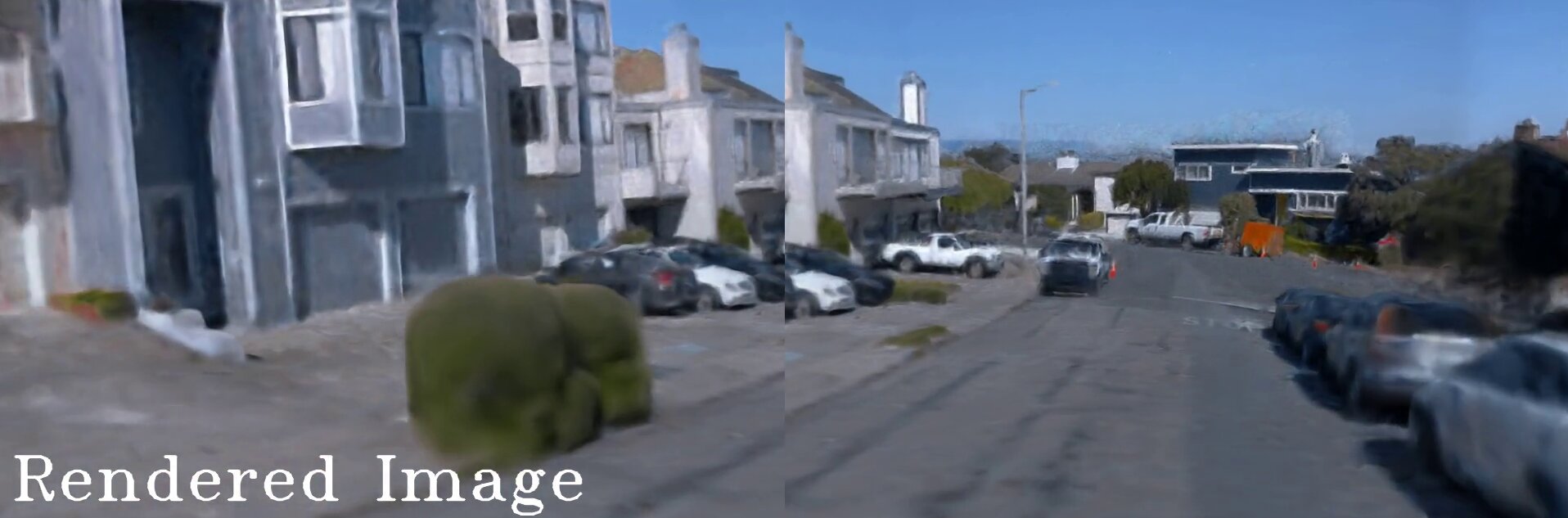}  
& \includegraphics[width=0.32\textwidth]{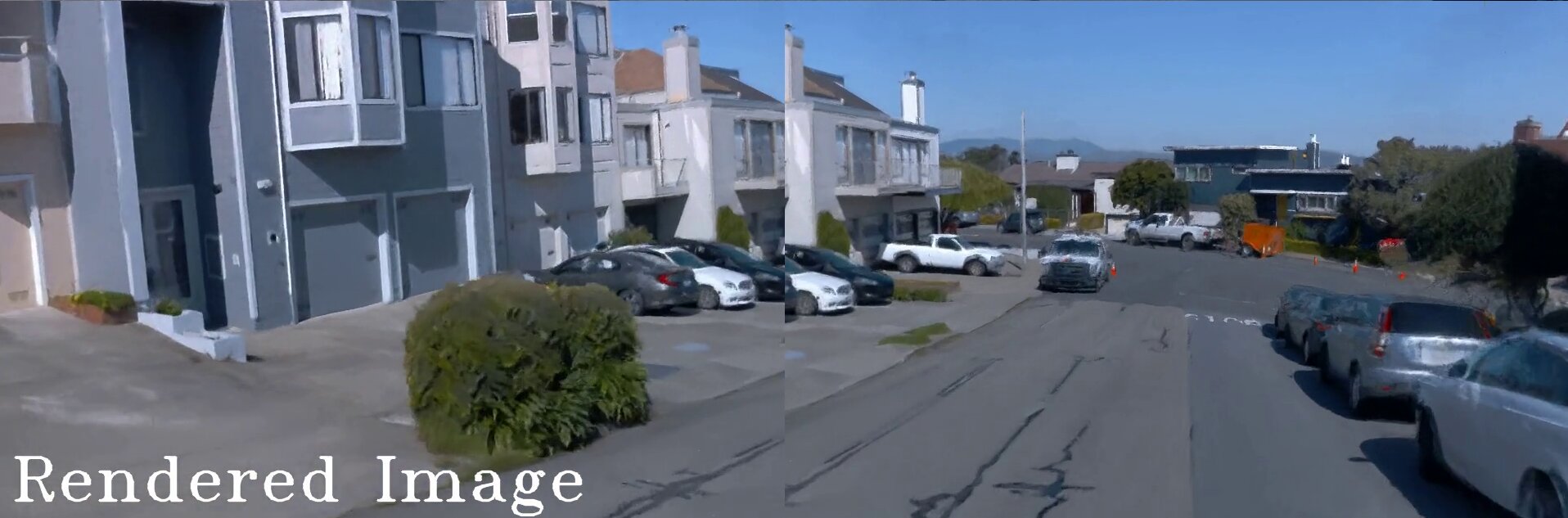}  
& \includegraphics[width=0.32\textwidth]{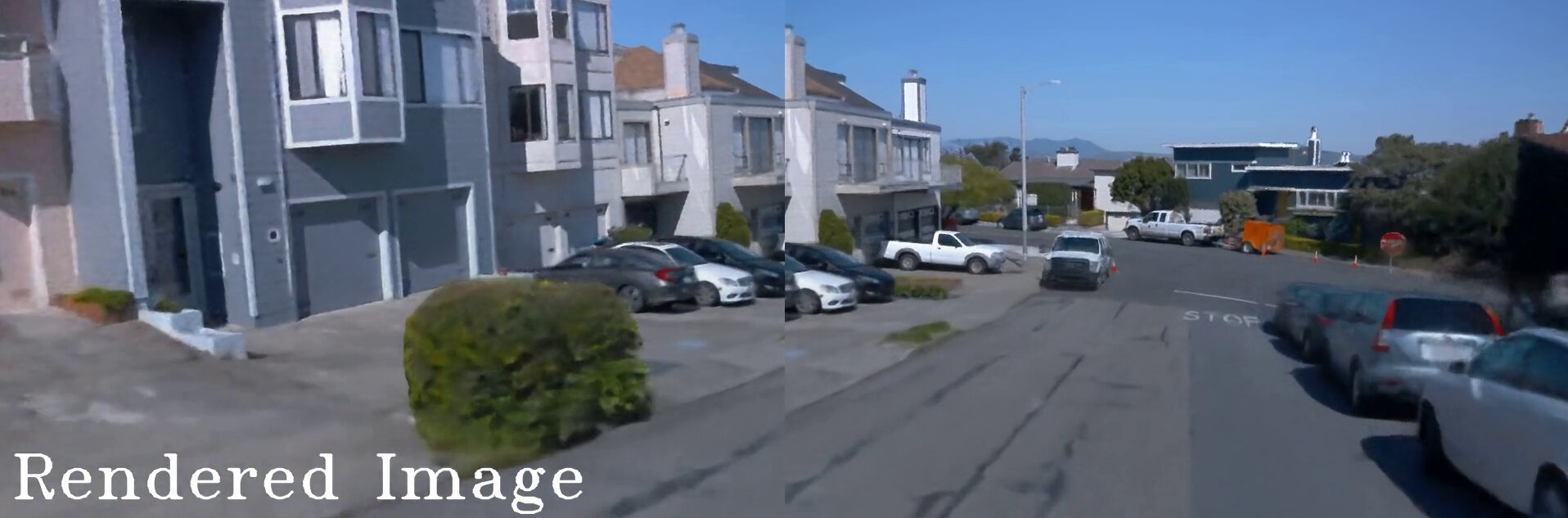}   \\[-2pt]

\raisebox{0.05\textwidth}[0pt][0pt]{\rotatebox[origin=c]{90}{\small{Depths}}}
& \includegraphics[width=0.32\textwidth]{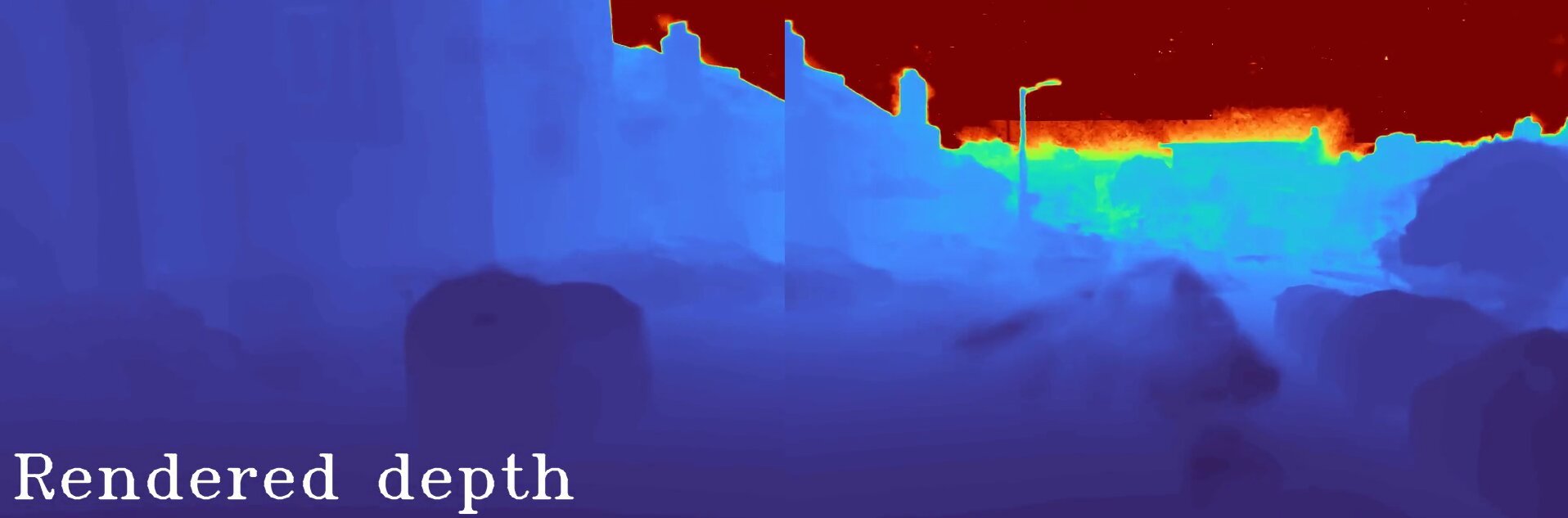}  
& \includegraphics[width=0.32\textwidth]{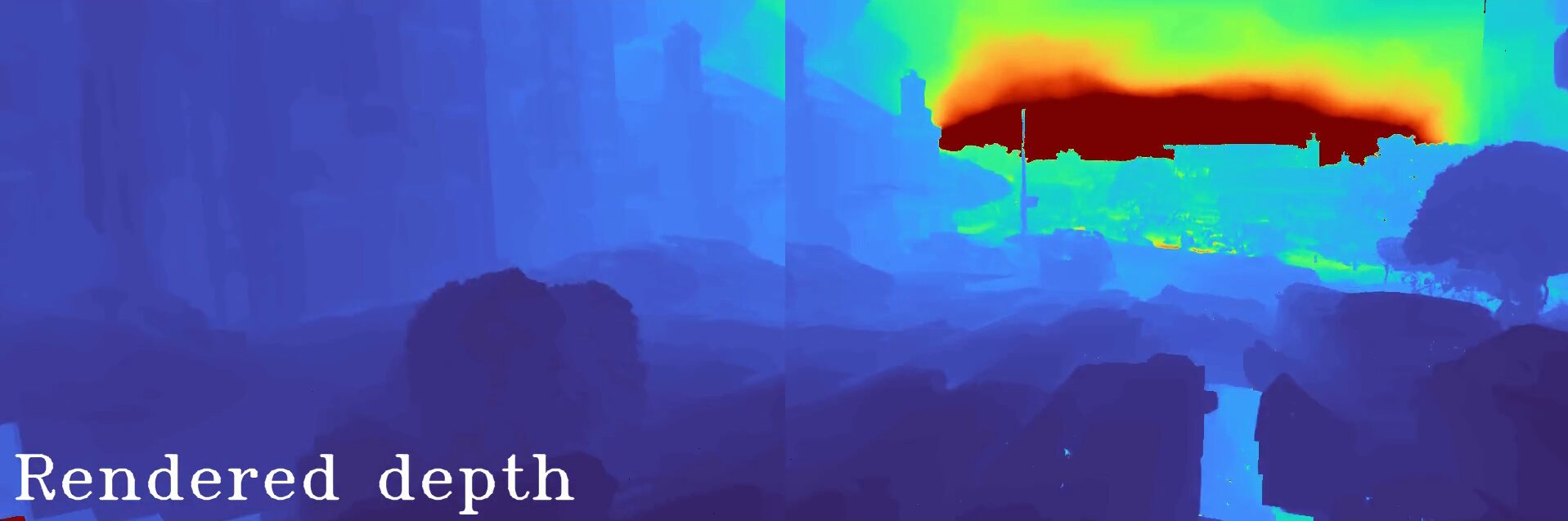}  
& \includegraphics[width=0.32\textwidth]{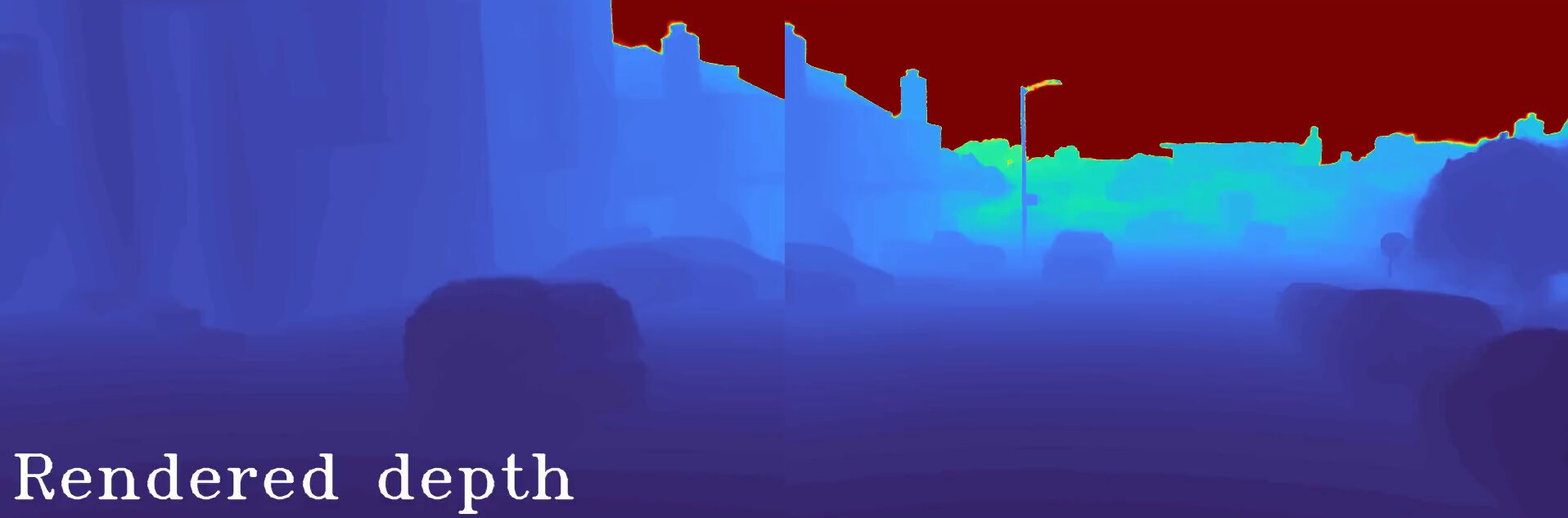}   \\[-2pt]

\raisebox{0.11\textwidth}[0pt][0pt]{\rotatebox[origin=c]{90}{\small{Reconstructed surfaces
}}}
& \includegraphics[width=0.32\textwidth]{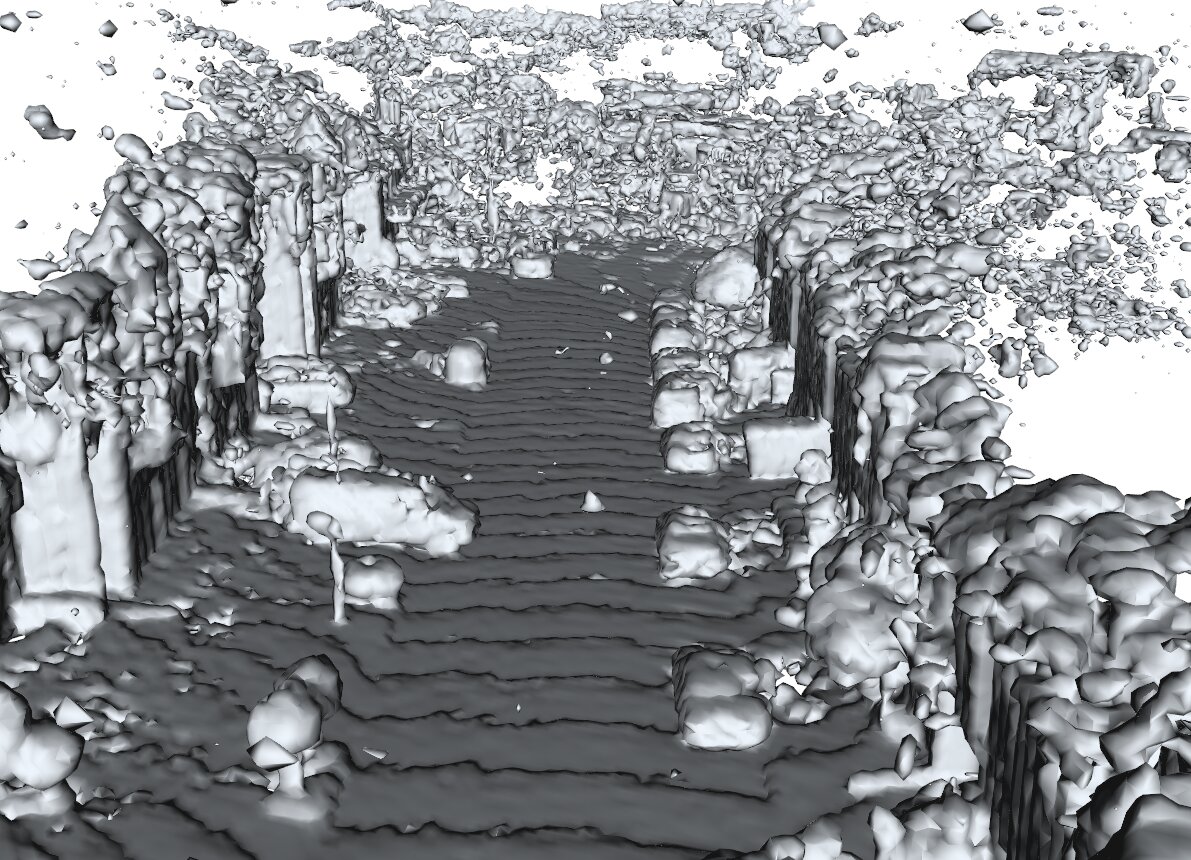}  
& \includegraphics[width=0.32\textwidth]{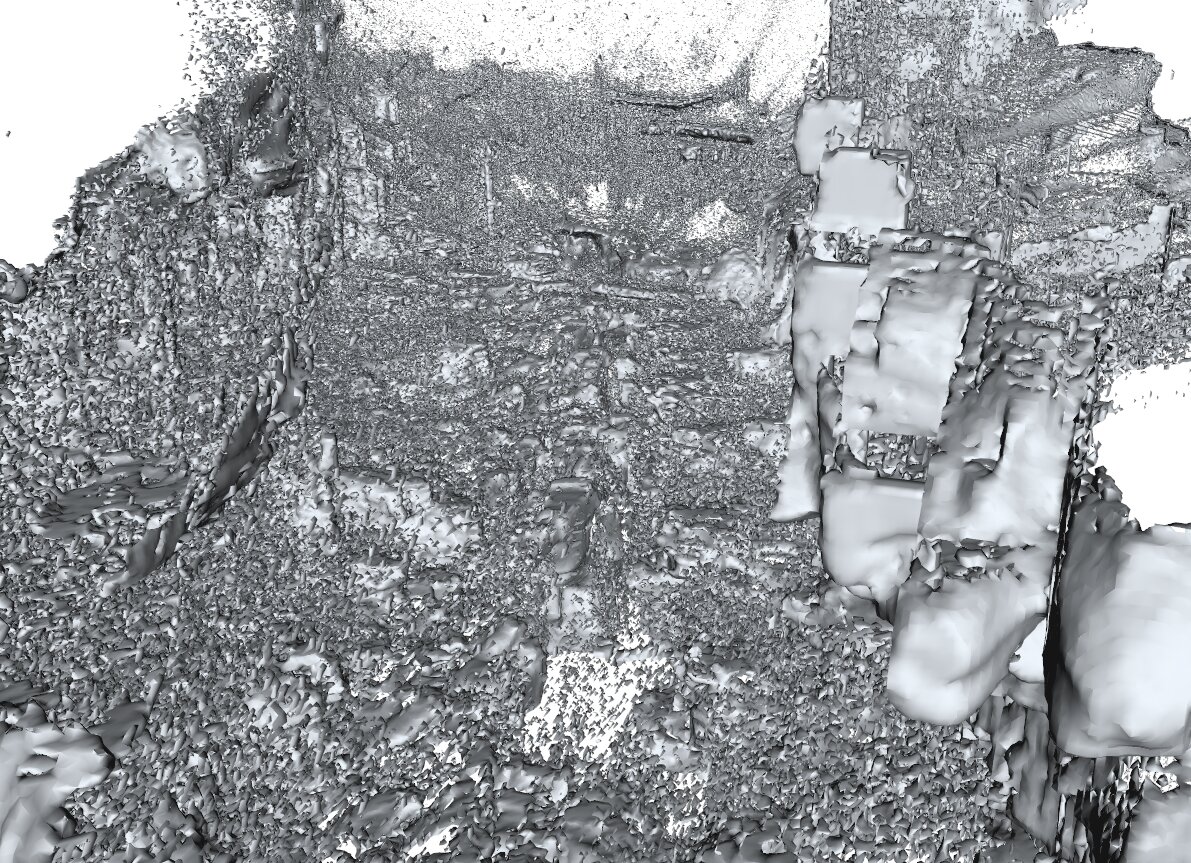}  
& \includegraphics[width=0.32\textwidth]{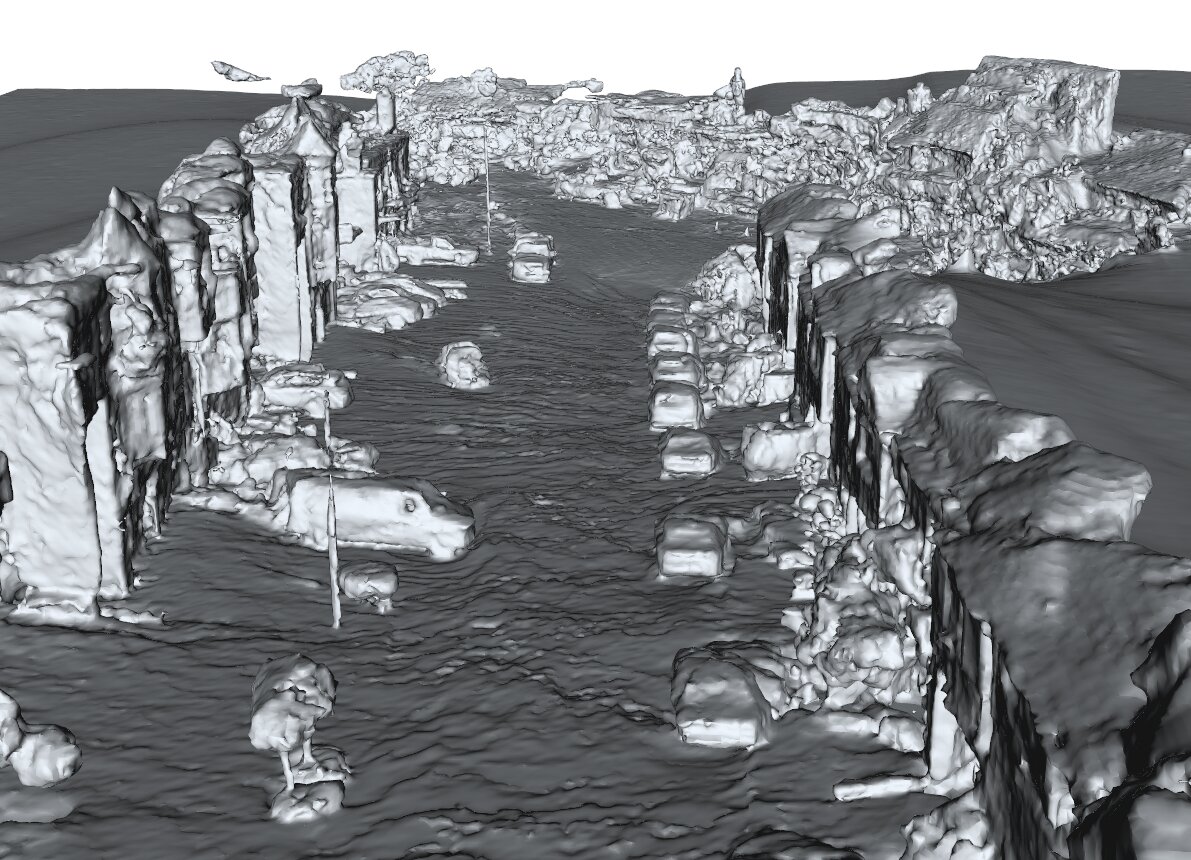}   \\[-2pt]

& \multicolumn{3}{c}{Waymo Open Dataset~\citep{waymo}, seg1006130\dots} \\

& & & \\

\raisebox{0.05\textwidth}[0pt][0pt]{\rotatebox[origin=c]{90}{\small{Images}}}
& \includegraphics[width=0.32\textwidth]{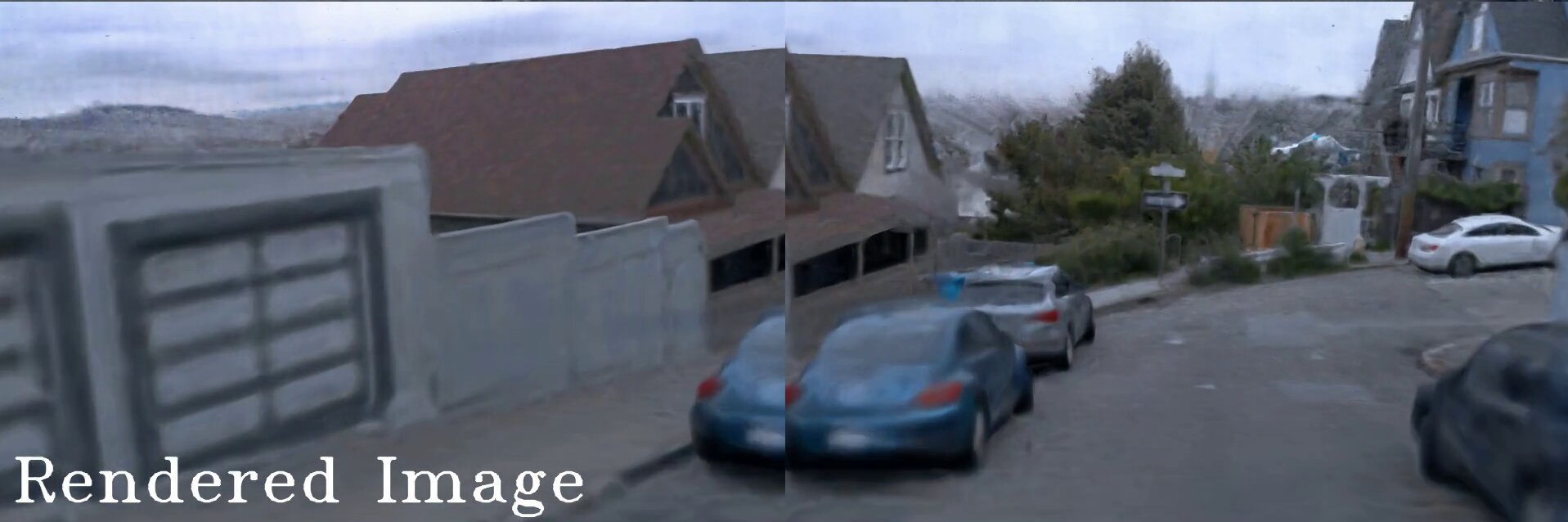}  
& \includegraphics[width=0.32\textwidth]{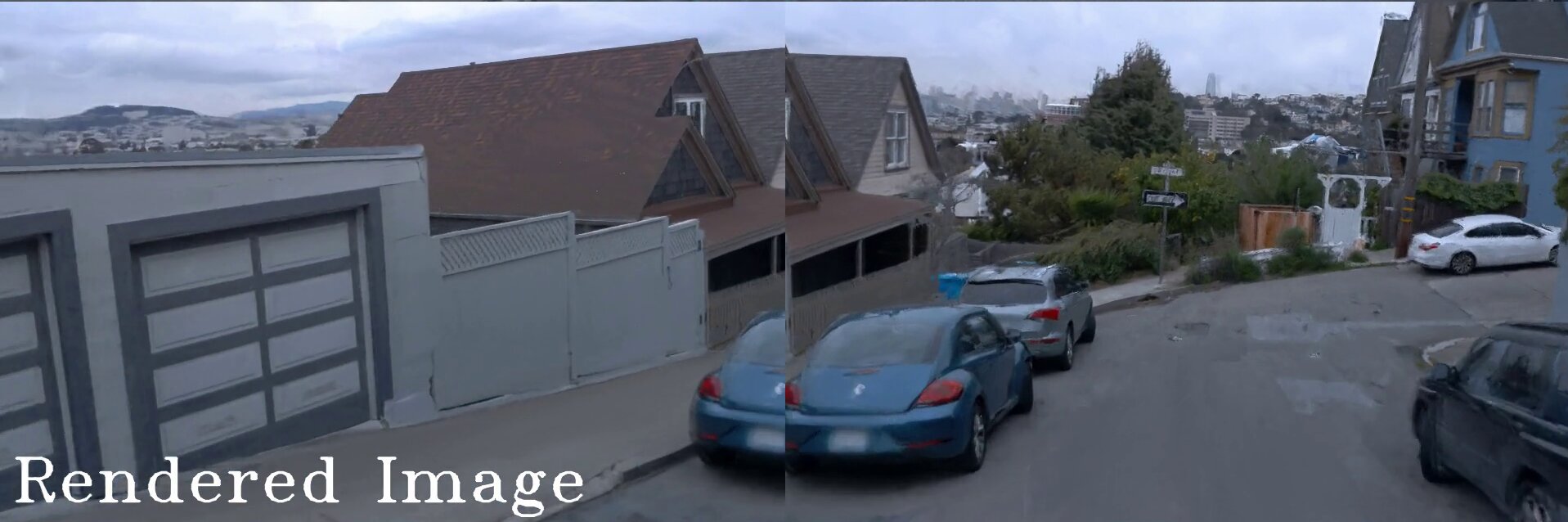}  
& \includegraphics[width=0.32\textwidth]{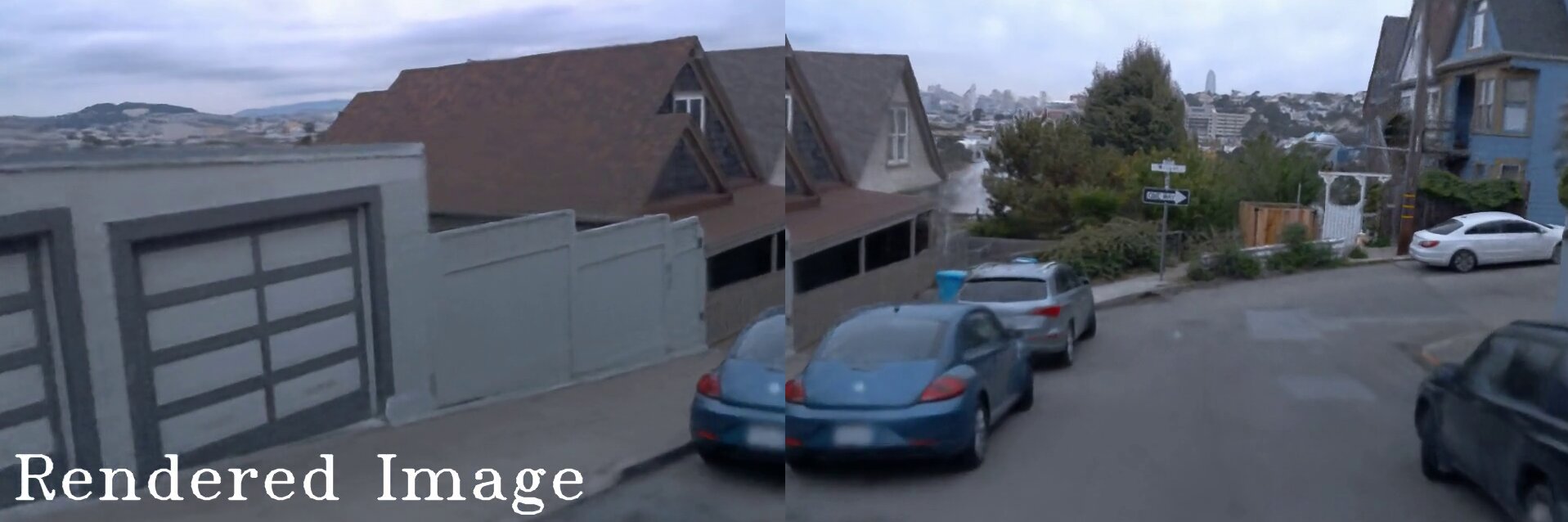}   \\[-2pt]

\raisebox{0.05\textwidth}[0pt][0pt]{\rotatebox[origin=c]{90}{\small{Depths}}}
& \includegraphics[width=0.32\textwidth]{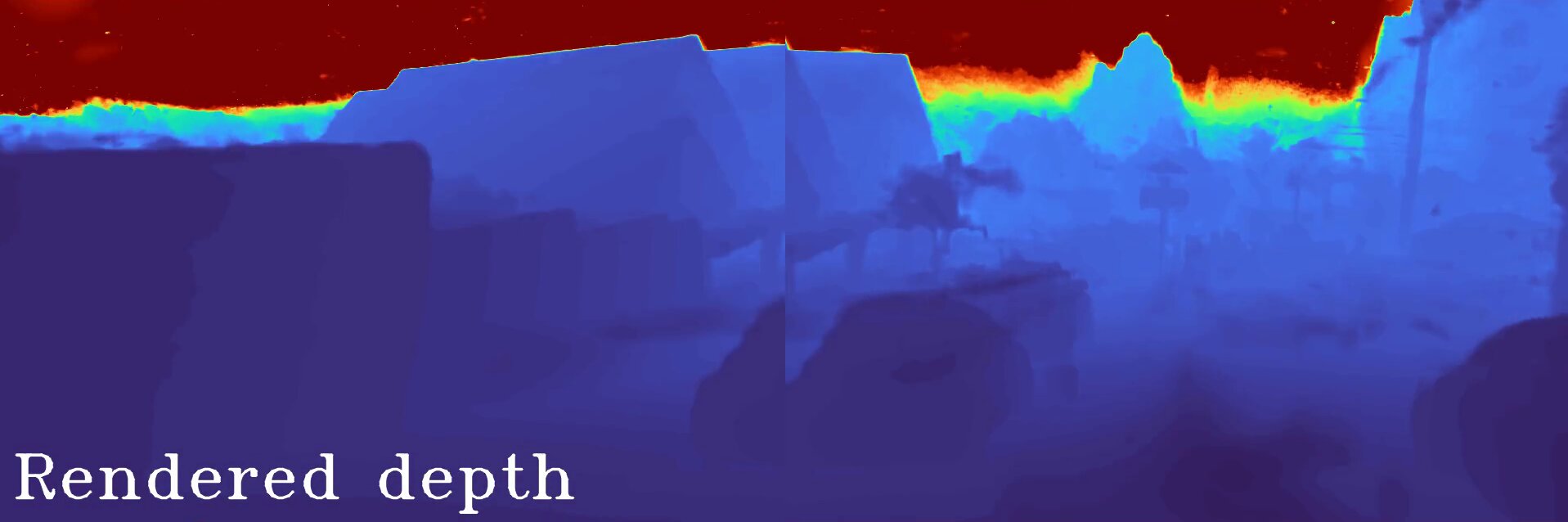}  
& \includegraphics[width=0.32\textwidth]{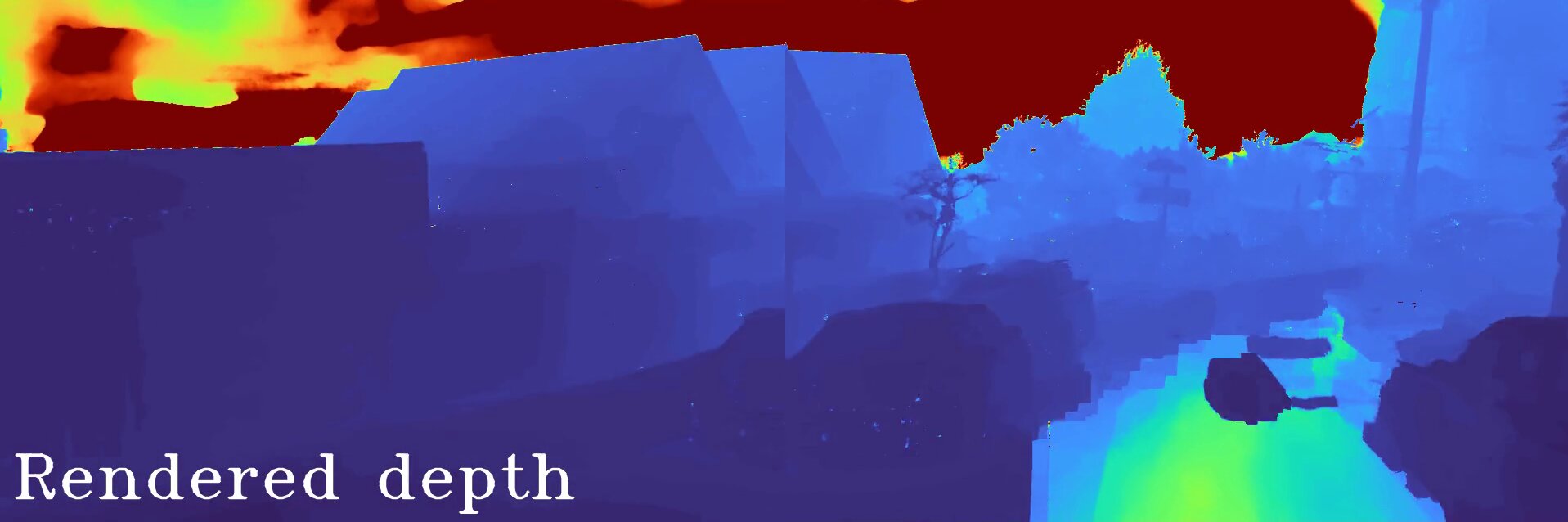}  
& \includegraphics[width=0.32\textwidth]{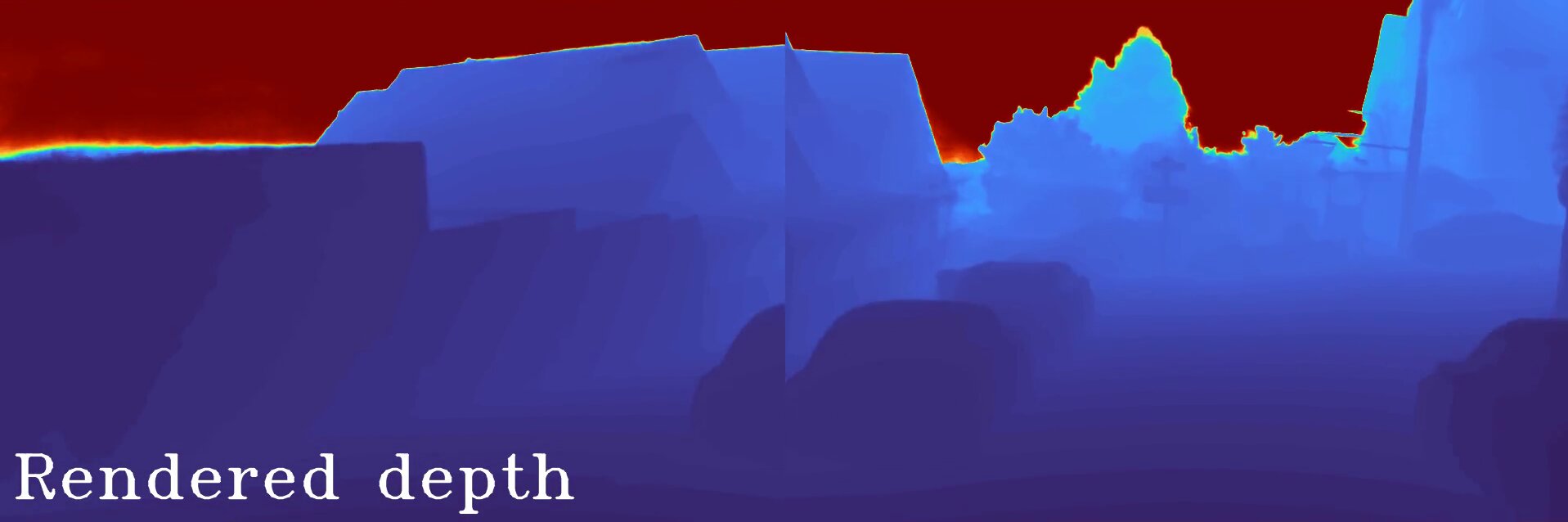}   \\[-2pt]

\raisebox{0.11\textwidth}[0pt][0pt]{\rotatebox[origin=c]{90}{\small{Reconstructed surfaces
}}}
& \includegraphics[width=0.32\textwidth]{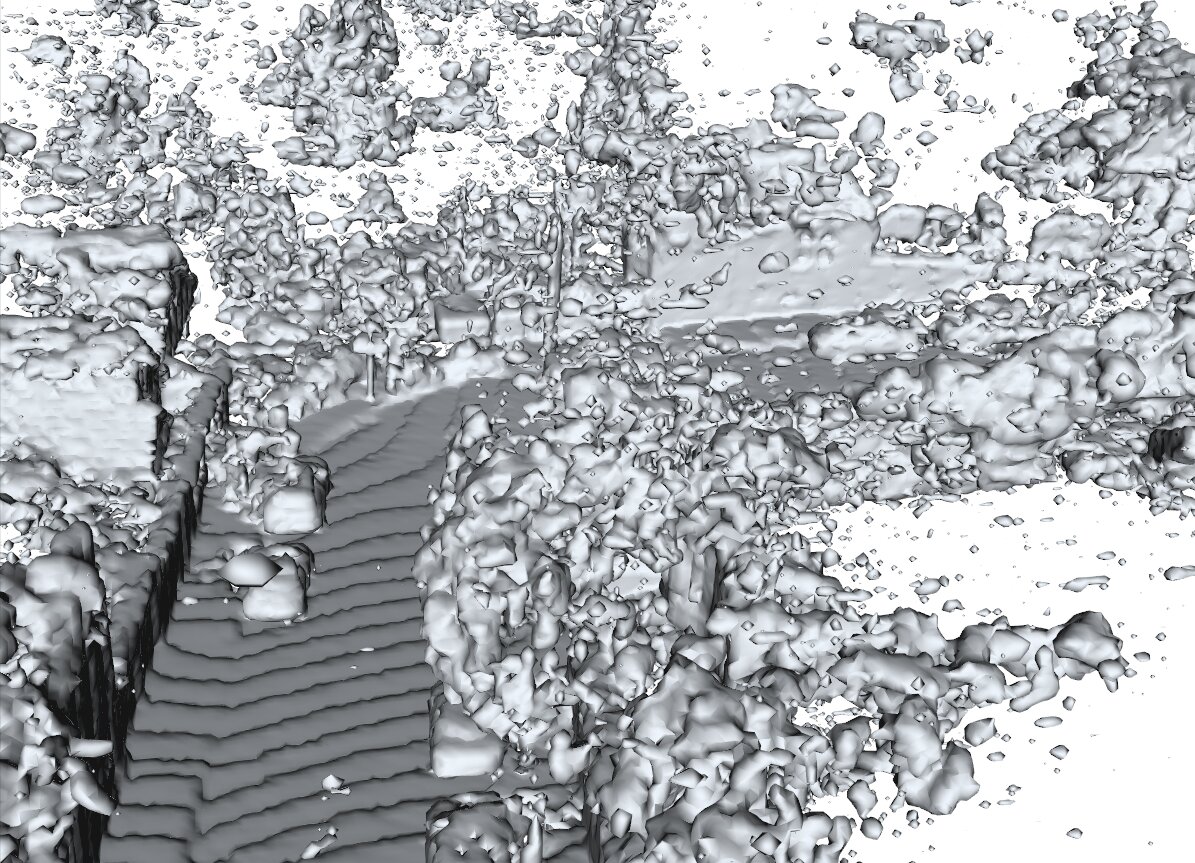}  
& \includegraphics[width=0.32\textwidth]{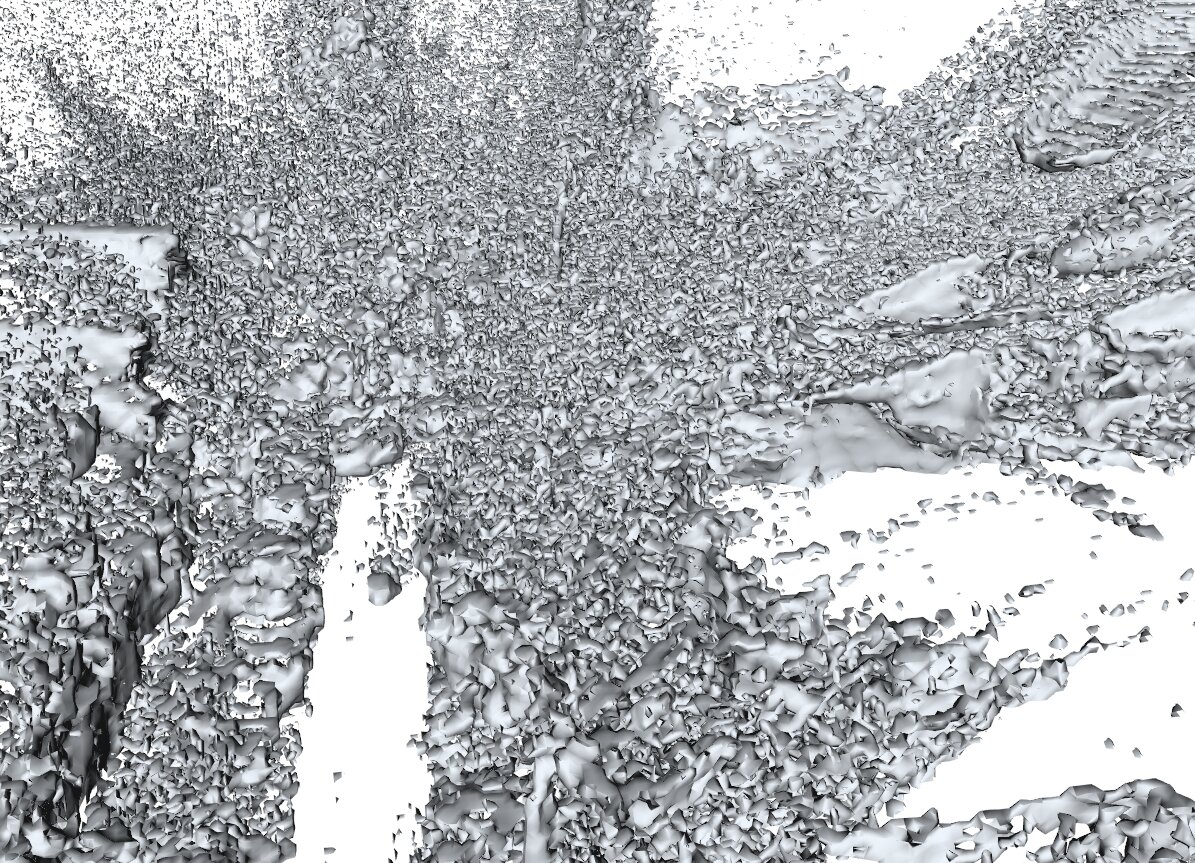}  
& \includegraphics[width=0.32\textwidth]{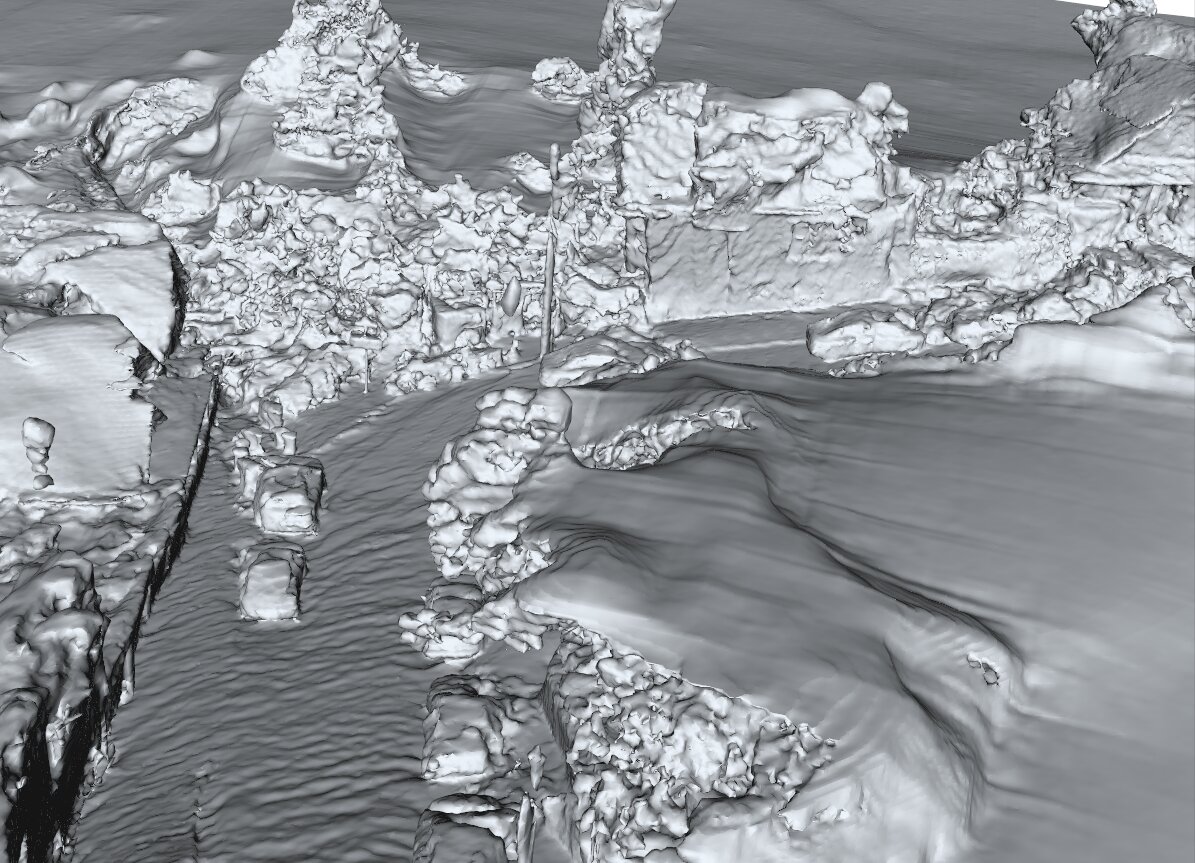}   \\[-2pt]

& \multicolumn{3}{c}{Waymo Open Dataset~\citep{waymo}, seg1347637\dots} \\

& & & \\

\raisebox{0.05\textwidth}[0pt][0pt]{\rotatebox[origin=c]{90}{\small{Images}}}
& \includegraphics[width=0.32\textwidth]{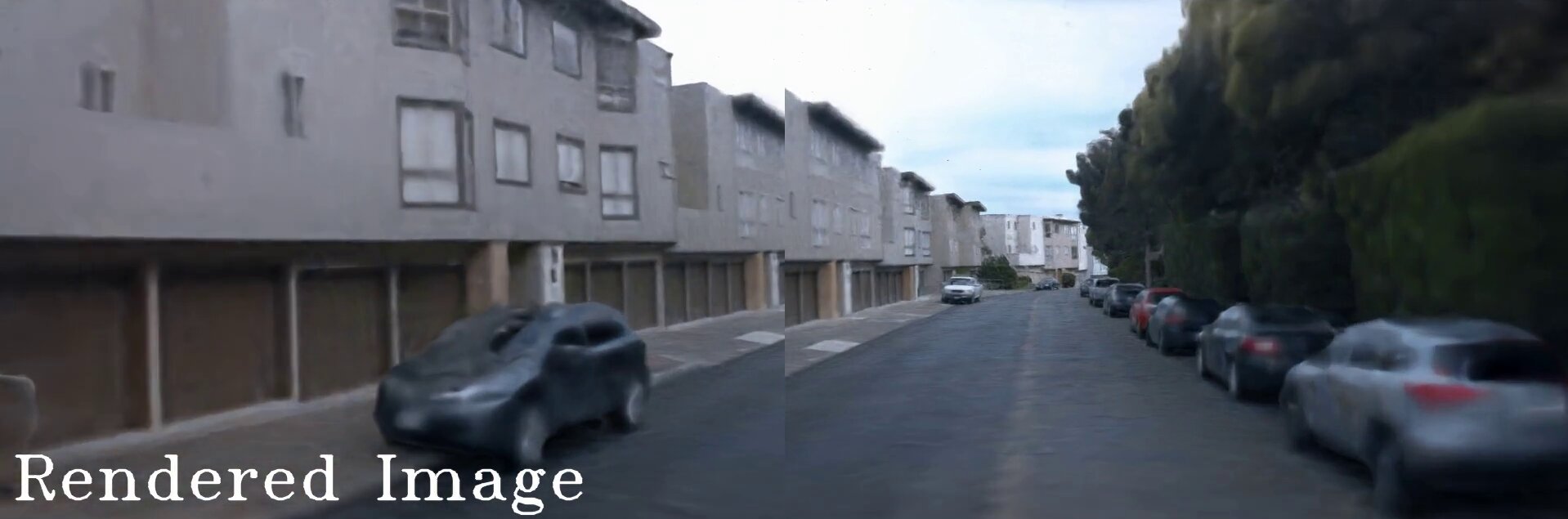}  
& \includegraphics[width=0.32\textwidth]{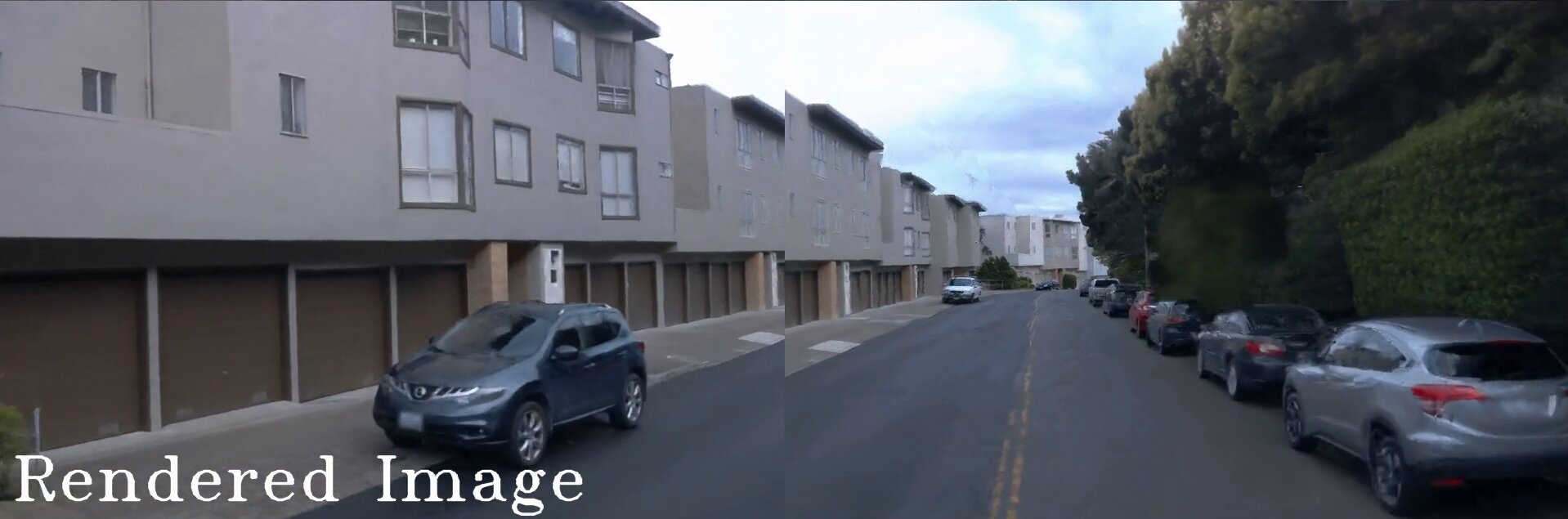}  
& \includegraphics[width=0.32\textwidth]{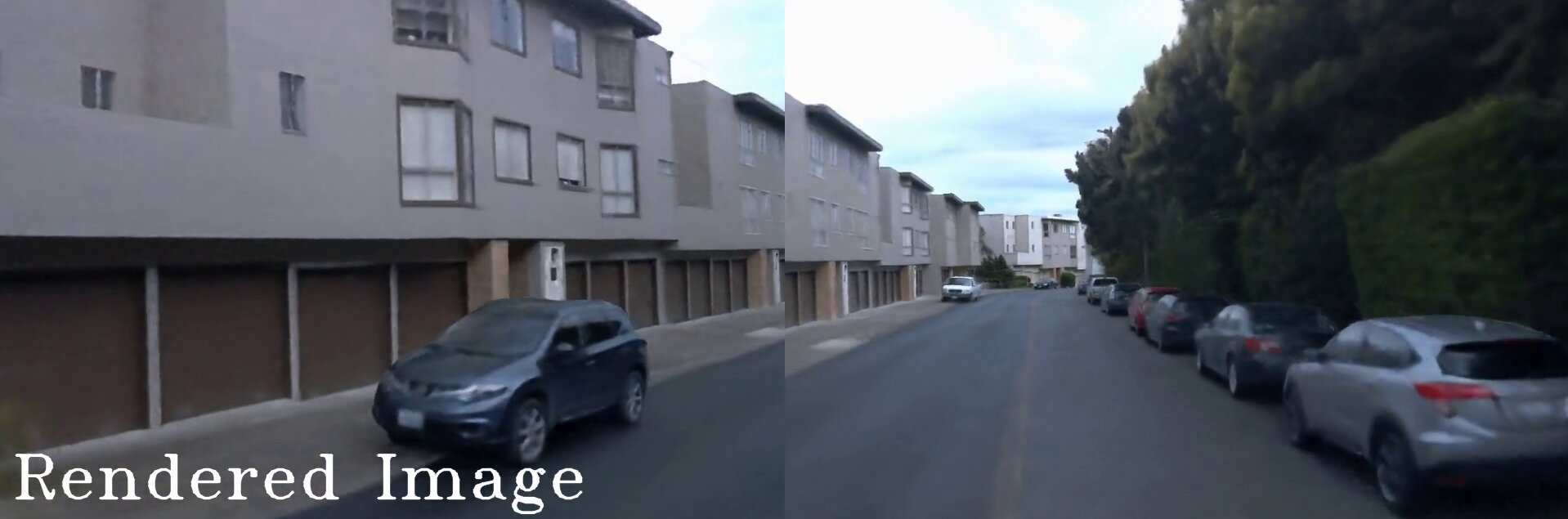}   \\[-2pt]

\raisebox{0.05\textwidth}[0pt][0pt]{\rotatebox[origin=c]{90}{\small{Depths}}}
& \includegraphics[width=0.32\textwidth]{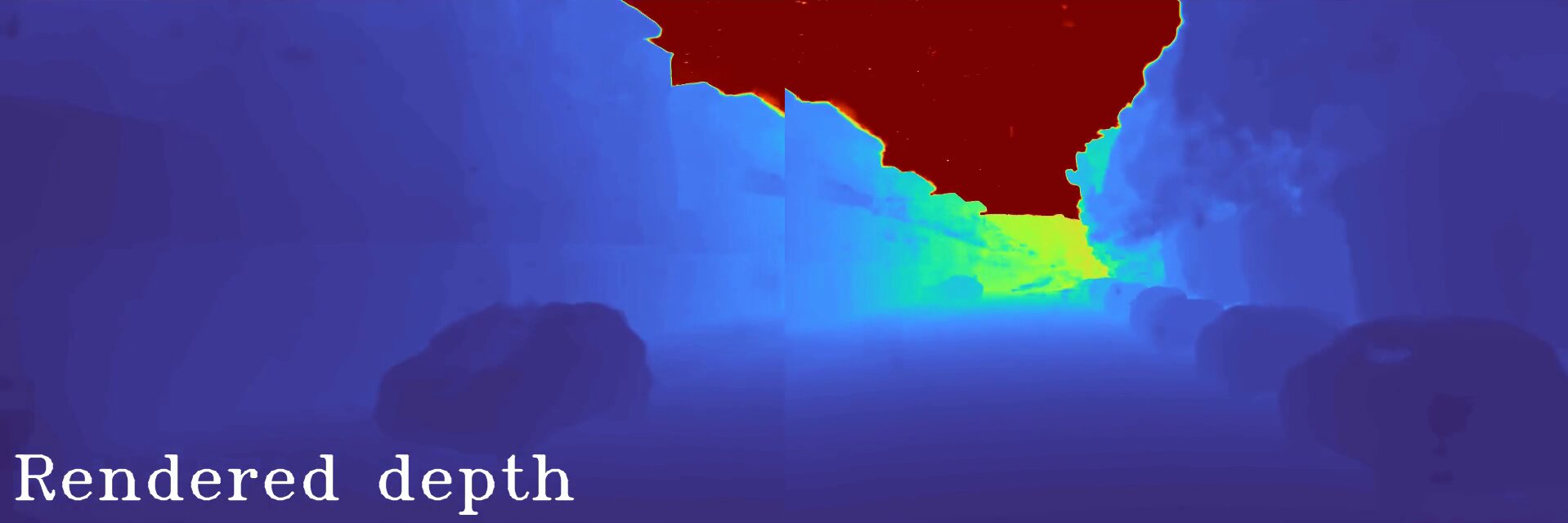}  
& \includegraphics[width=0.32\textwidth]{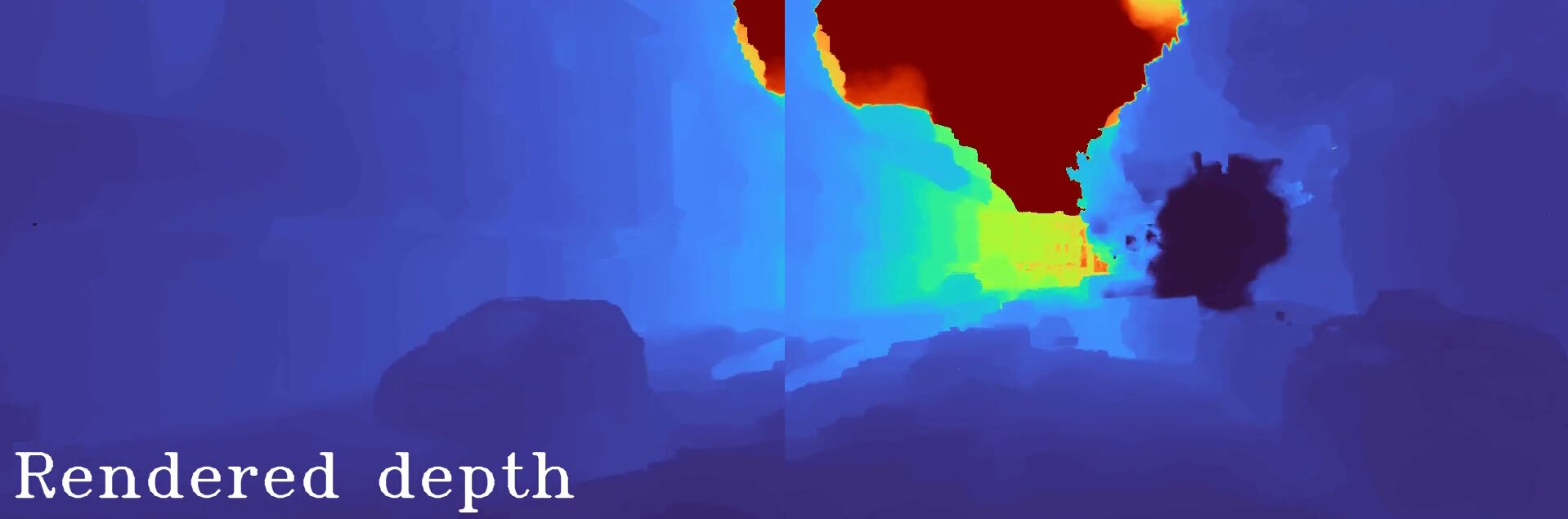}  
& \includegraphics[width=0.32\textwidth]{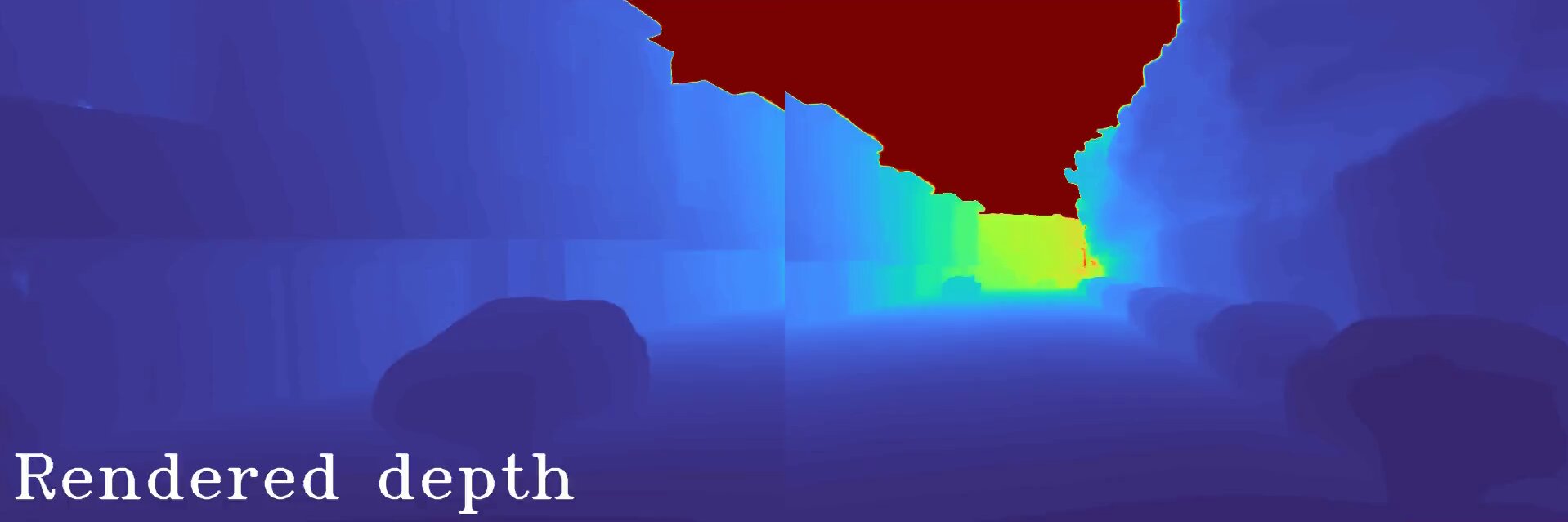}   \\[-2pt]

\raisebox{0.11\textwidth}[0pt][0pt]{\rotatebox[origin=c]{90}{\small{Reconstructed surfaces
}}}
& \includegraphics[width=0.32\textwidth]{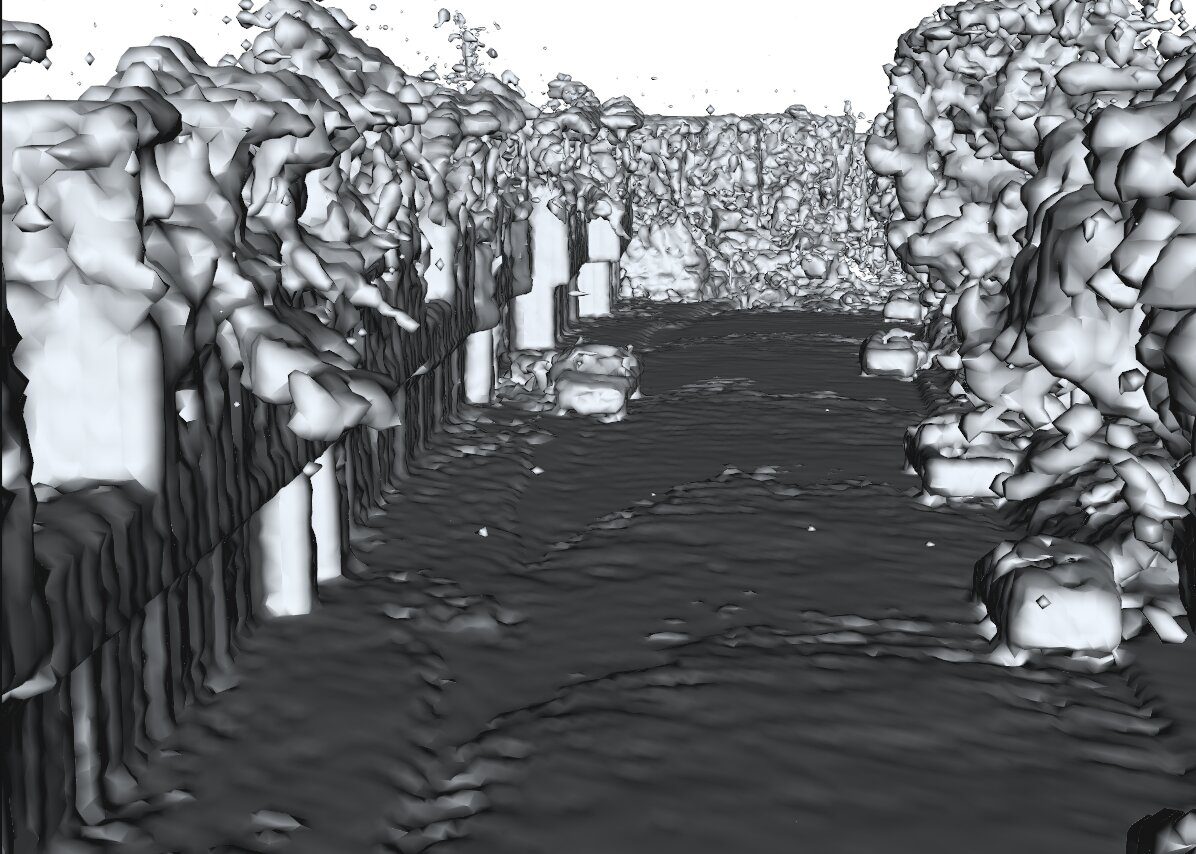}  
& \includegraphics[width=0.32\textwidth]{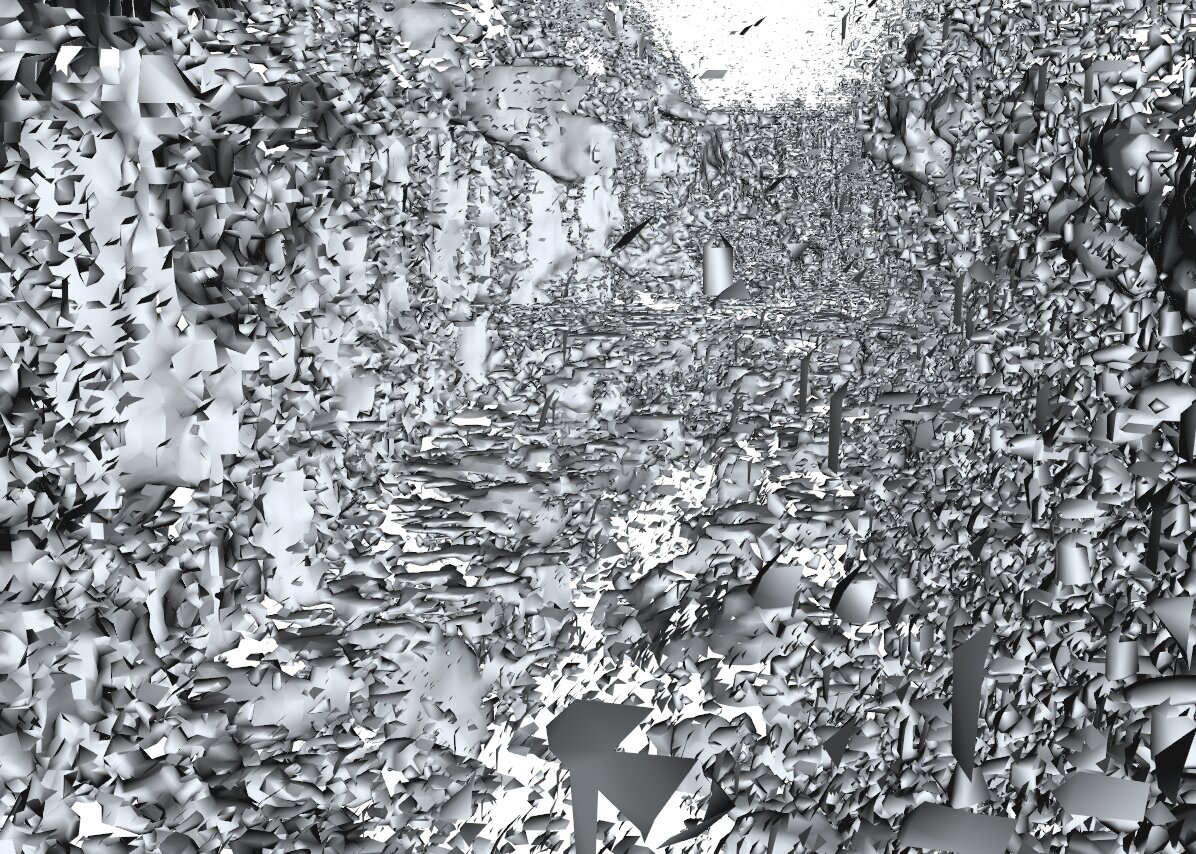}  
& \includegraphics[width=0.32\textwidth]{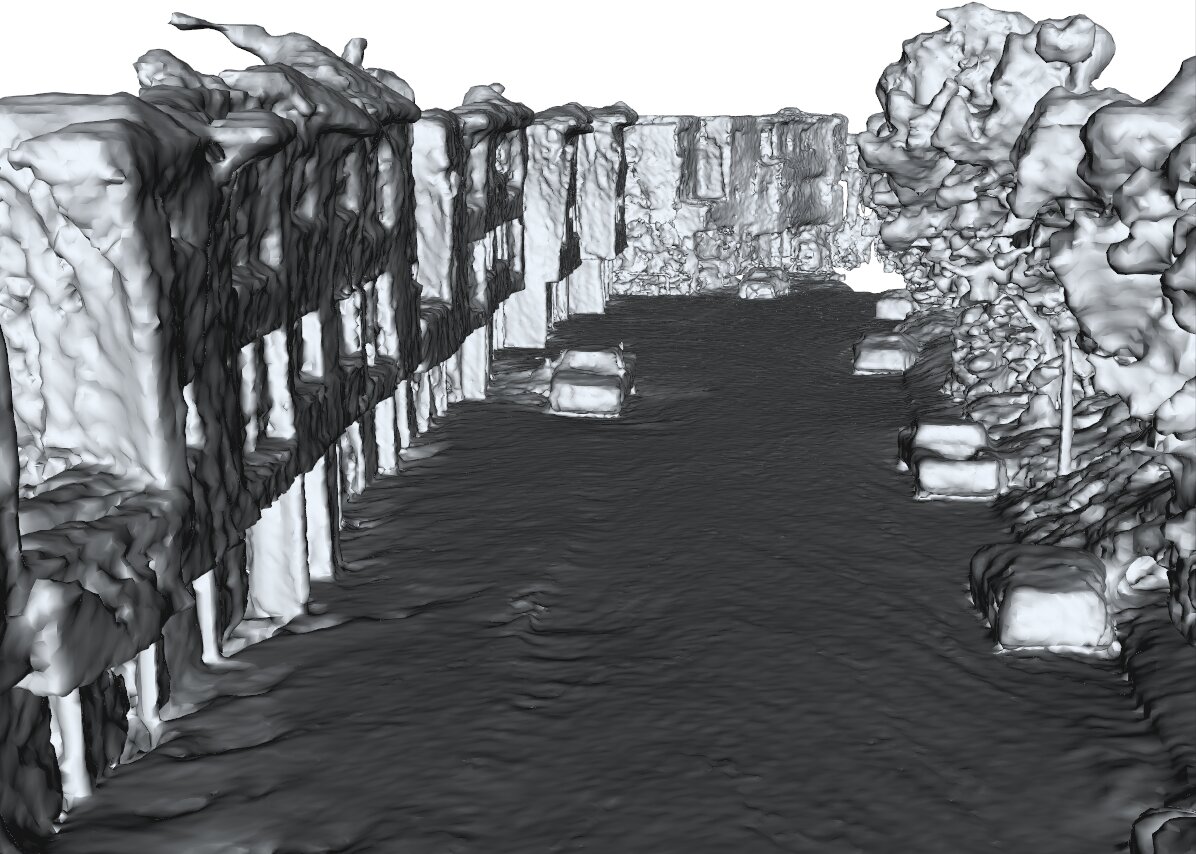}   \\[-2pt]

& \multicolumn{3}{c}{Waymo Open Dataset~\citep{waymo}, seg1527063\dots} \\

&
\multicolumn{1}{c}{
\begin{subfigure}{0.32\textwidth}
	\caption{NGP~\citep{muller2022instantngp}, with LiDAR~\citep{rematas2022urban}}
	\label{fig:urbanngp}
\end{subfigure}
}
& \multicolumn{1}{c}{
\begin{subfigure}{0.32\textwidth}
	\caption{$\text{F}^2$-NeRF~\citep{wang2023f2nerf}, without LiDAR}
	\label{fig:f2nerf}
\end{subfigure}
}
& \multicolumn{1}{c}{
\begin{subfigure}{0.32\textwidth}
	\caption{Ours, \textbf{without LiDAR}}
	\label{fig:ours}
\end{subfigure}
}
\end{tabular}

\caption{
Qualitative comparison of our work with recent studies, including one that requires LiDAR data input and another that does not. For NGP~\citep{muller2022instantngp} and our work, we apply Marching Cubes~\citep{marchingcubes} to extract meshes. For $\text{F}^2$-NeRF~\citep{wang2023f2nerf}, we leverage vdbfusion~\citep{vdbfusion} to the rendered camera depth since we are temporarily unable to extract mesh using their recently released code-base. 
}
\end{figure}

Another challenge posed by the non-object-centric views is the geometric errors arising from the insufficient coverage of viewing angles. 
It is even worse regarding street-view images in autonomous driving datasets, since each data sequence in these datasets only cover one single street with no repeated observations.
Fig.~\ref{fig:f2nerf} shows that while the state of the art novel view synthesis method~($\text{F}^2$-NeRF~\citep{wang2023f2nerf}) already aims at non-object-centric camera trajectories, they still struggle at the geometry of textureless objects and surfaces, which are often mistakenly interpreted as large cavity or holes. 
To tackle this challenge, we follow MonoSDF~\citep{Yu2022MonoSDF} and leverage monocular estimations to guide the volume-rendered normals and depths of the close-range NeuS model.


Apart from disentanglement and geometric priors, we also implement a novel ray marching strategy to achieve efficient and fine-grained ray sampling in the long and narrow street-view spaces with complex occlusions. We combine the previous idea of occupancy-grid-based acceleration~\citep{muller2022instantngp} with multi-stage hierarchical sampling~\citep{mildenhall2021nerf,wang2021neus}, and implement the atomic pack-like operations under CUDA, which ensures granularity without sacrificing efficiency.

In summary, we make the following contributions:
\begin{itemize}[leftmargin=*]
	\item We present \textit{StreetSurf}, a novel multi-view implicit surface reconstruction framework that achieves state of the art geometry and appearance quality in street views within an hour or two with a single RTX3090 GPU, without necessarily requiring LiDAR data.
	\item We propose a close-range-vs-distant-view disentanglement approach suitable for unbounded scenes captured with non-object-centric, long and narrow camera trajectories.
	\item We conduct experiments in challenging real-world autonomous driving datasets such as Waymo Open Dataset~\citep{waymo} and provide a benchmark for the reconstructed geometry and appearance.
\end{itemize}

\begin{figure}[htbp]
\includegraphics[width=\textwidth]{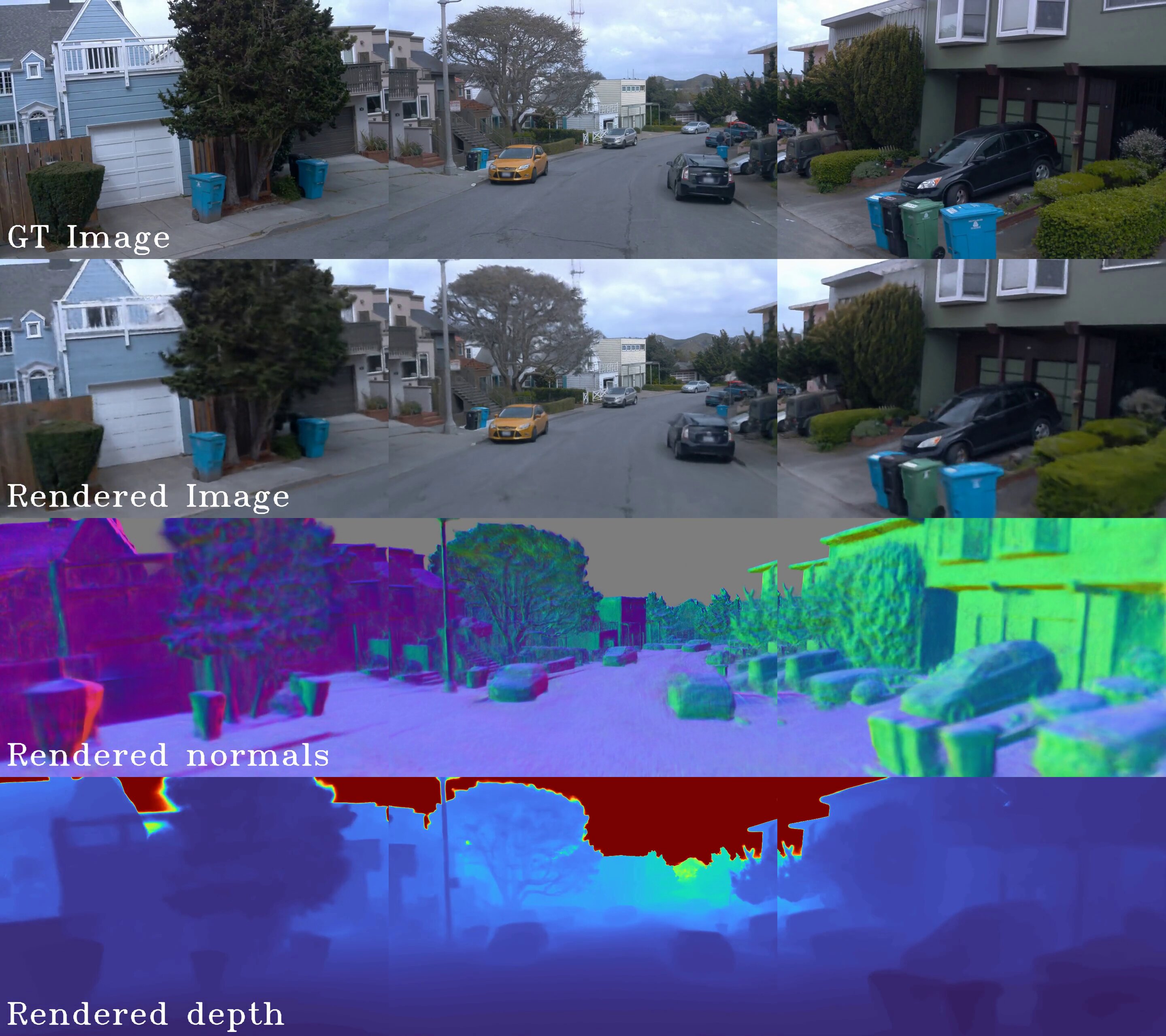}
\caption{Qualitative results of \textbf{reconstruction without LiDAR data}. We apply our method to Waymo Open Dataset~\citep{waymo} seg1534950\dots with a trajectory length of 112.1 (m).}
\label{fig:head}
\end{figure}


%
%
%
%
%

\section{Related work}

\textbf{NeRF on street views}. 
Research on integrating NeRF~\citep{mildenhall2021nerf} to street views or urban scenes has recently started to gain attention. 
Some addresses the challenges of large-scale outdoor scenes by using local blocks \citep{tancik2022blocknerf,Turki_2022_CVPR,turki2023suds} or other scalable representations~\citep{li2022read, rematas2022urban}.
Others~\citep{ost2021neural,xie2023street-nerf,KunduCVPR2022PNF,fu2022panoptic} instead, propose to handle relatively smaller street-view scenes available in widely-used autonomous driving datasets, such as KITTI~\citep{geiger2012we}, Waymo~\citep{waymo}, or NuScenes~\citep{caesar2020nuscenes}.
In this paper, we also focus on these autonomous driving scenarios since we expect our reconstruction results to be directly useful for perception methods trained on these datasets.
Many research in this field studies compositional scene representations, either under multi-object settings~\citep{ost2021neural,xie2023street-nerf} or static-dynamic settings~\citep{turki2023suds, martin2021nerfw, sun2022neural-recon-w}. 
Without explicit multi-object decomposition, ~\citep{KunduCVPR2022PNF,fu2022panoptic} model semantic fields using panoptic segmentation.
Another line of research~\citep{tancik2022blocknerf,li2022read,wang2023f2nerf,rematas2022urban,wang2023fegr} focuses on static street backgrounds and ignores dynamic foreground objects. 
Most of them aim only at the task of novel view synthesis.
While UrbanNeRF~\citep{rematas2022urban} and FEGR~\citep{wang2023fegr} are capable of reconstructing decent surfaces, they rely on dense LiDAR inputs
and they do not handle the distant-view that can interfere with the close-range reconstruction.
In this paper, we focus on the task of multi-view implicit surface reconstruction for static close-range street-view backgrounds without necessarily requiring LiDAR data. 

\textbf{Spatial scene representations}.
Neural fields~\citep{xie2022neural} or implicit scene representations are experiencing a surge of interest.
\citep{fathony2021multiplicative, lindell2022bacon, mildenhall2021nerf, wang2021neus, ramasinghe2022beyond, sitzmann2020implicit} are based on different modification of multi layer perceptrons~(MLPs).
Another line of research establish their scene representation on spatial grids or voxels, known as local implicits or hybrid representations.
\citep{sitzmann2019deepvoxels,sun2022direct} and earlier works use dense feature grids.
To improve spatial efficiency, voxel-pruning~\citep{liu2020neural,takikawa2021nglod,yu2021plenoctrees,yu2022plenoxels,li2022voxsurf}, hash-indexing~\citep{muller2022instantngp}, tensor-decomposition~\citep{chen2022tensorf,chan2022efficient}, or multi-scale voxels \citep{martel2021acorn} are introduced.
Instead of cubic voxels, other works utilize special forms of voxels like tetrahedra~\citep{Munkberg_2022_CVPR} or permutohedra~\citep{rosu2023permutosdf}.
In this paper, we establish 3D/4D cuboid/hyper-cuboid hash-grids for the close-range/distant-view spaces respectively.
To due with unbounded scenes, two spatial warping functions are commonly used, namely inverse-spherical warping~\citep{barron2021mipnerf,barron2022mipnerf360,zhang2020nerf++} and NDC warping~\citep{mildenhall2021nerf}. 
$\text{F}^2$-NeRF~\citep{wang2023f2nerf} and \citep{meuleman2023progressively} that warps the full space according to certain rules to further reduce capacity waste.
In this paper, instead of full-space warping, we explicitly delimit the close-range and distant-view spaces with a long and narrow cuboid boundary that is aligned with the camera trajectory.
\textbf{Multi-view implicit surface reconstruction}. 
3D Reconstruction from multi-view images has been a long-time aim in the field of computer vision.
Recently, instead of modeling the scene with explicit shape representation as traditional multi-view stereo pipelines like COLMAP~\citep{schoenberger2016sfm} did, using neural representations for implicit surfaces~\citep{yariv2020idr,wang2021neus,yariv2021volume,oechsle2021unisurf} has begun to show extraordinary results using object-centric views.
NeuS~\citep{wang2021neus} or VolSDF~\citep{yariv2021volume} models the underlying 3D geometry with SDF fields and utilizes volume rendering to produce differentiable colors.
Follow-up works replace the MLP network with local implicit grids~\citep{wang2022neus2,rosu2023permutosdf}, sparse voxels~\citep{li2022voxsurf}, MLP blocks~\citep{esposito2022kiloneus} or displacement fields~\citep{wang2022hf} for better efficiency or local details.
Multi-view 3D reconstruction is typically an under-constrained problem, driving researchers to  introduce different priors or regularizations, such as multi-view stereo prior~\citep{fu2022geo}, geometric priors~\citep{Yu2022MonoSDF,guo2022neural,mu2023neural,wang2022neuralroom}, shadow information~\citep{ling2022shadowneus} and sparse regularization~\citep{zhang2022critical}.
After reconstruction, further research \citep{qiu2023looking,chen2022tracing} conduct study on decouple and dealing with ambient light or specular effect.
Besides single objects, \citep{guo2022neural,wang2022neuris,Yu2022MonoSDF,wang2022neuralroom,liang2023helixsurf} have been applied to indoor datasets.
Meanwhile, \citep{sun2022neural-recon-w,zhang2022critical} extends to outdoor datasets. 
However, they are still using data captured with object-centric views.
In this paper, we extend this line of research to handle the unique challenges in street views.


%

\section{Method}







\begin{figure}[htbp]
\centering
\includegraphics[width=\textwidth]{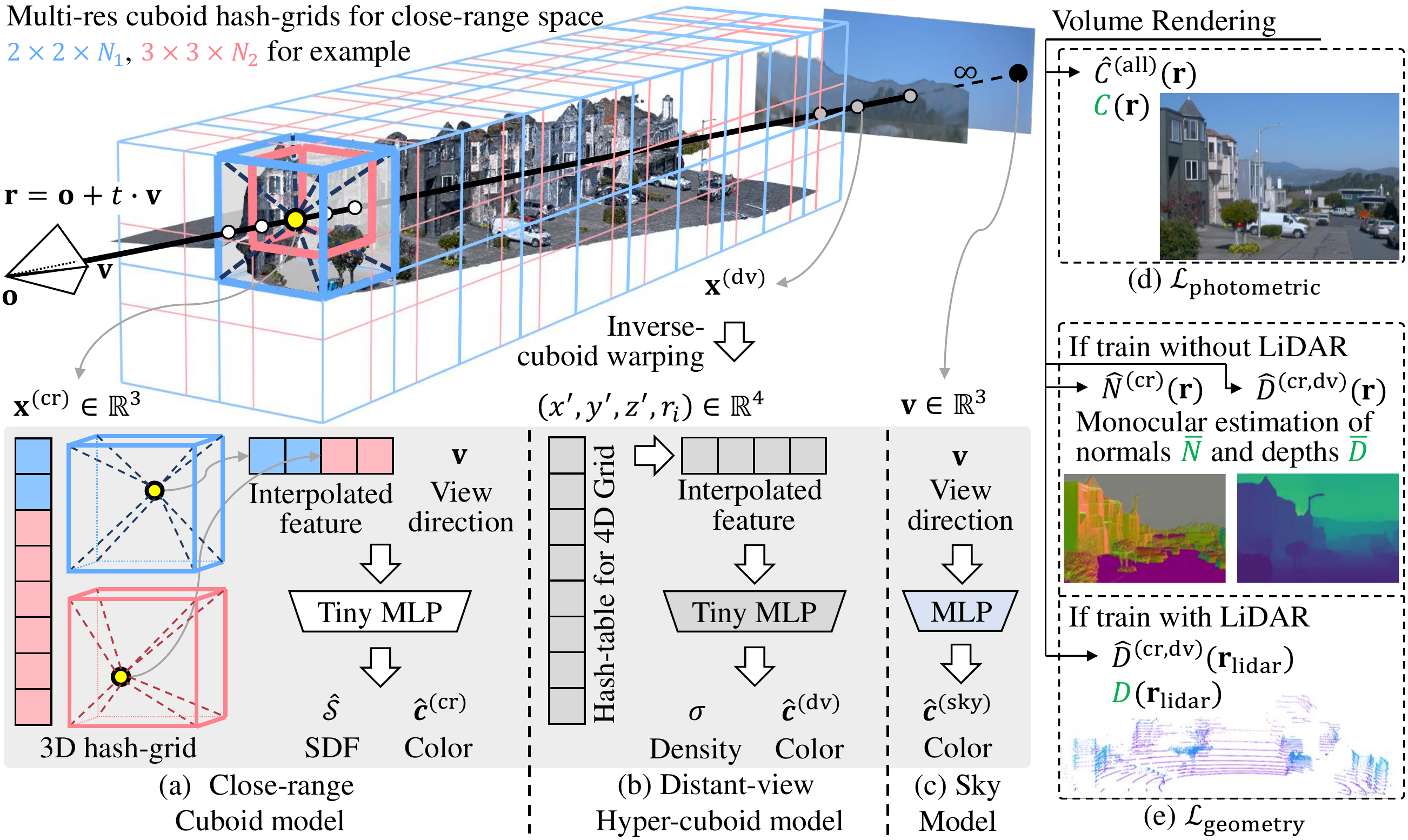}
\par
\vspace{-5pt}
\caption{StreetSurf overview. By using a collection of posed images, along with an optional set of LiDAR data or monocular cues as inputs (or supervisions), we reconstruct the street-view scene through an optimization process. Each scene is divided into three parts according to the viewing distance, namely close-range~(cr), distant-view~(dv) and sky.}
\label{fig:overview}
\vspace{-10pt}
\end{figure}

%

\subsection{Multi-shell networks}

To tackle the challenge of unbounded spaces and significantly varying viewing distance in street-views, let us first consider an observer standing on a street, whose view can be divided into three parts according to the distance of objects.
Buildings and roads can be located within several to tens of meters, while landscapes and objects that extend beyond the end of the trajectory may be hundreds or thousands of meters away, and the sky seemingly infinitely far off.
In the following, we refer to these three parts as the \textbf{close-range}~(cr), \textbf{distant-view}~(dv), and \textbf{sky} parts, respectively.

We employ a cuboid NeuS model (modified from \citep{wang2021neus}), a hyper-cuboid NeRF++ model (modified from \citep{zhang2020nerf++}), and a directional MLP for the close-range scene, distant-view, and sky, respectively.

For each ray $\mathbf{r}$, we query all three models and jointly render a differentiable pixel color $\widehat{C}^{(\text{all})}(\mathbf{r})$.

Specifically, we first merge queried samples of close-range and distant-view models on the ray $\mathbf{r}$ from near to far, $\{t^{(\text{cr,dv})}_i\}_{i=1}^{k} = \{t^{(\text{cr})}_{1}, t^{(\text{cr})}_2, \dots, t^{(\text{cr})}_{\textrm{n}_\text{cr}}, t^{(\text{dv})}_1, t^{(\text{dv})}_2, \dots, t^{(\text{dv})}_{\mathrm{n}_\text{dv}}\}$, where $t^\text{(cr)}$ stands for depth samples on of the close-range model and $t^\text{(dv)}$ for the distant-view model. 

Then, we perform volume rendering on the combined samples of depth $t^{(\text{cr,dv})}$ and color $c^{(\text{cr,dv})}$ (we will omit the superscript of $(\text{cr,dv})$ on
${\{T_i,\alpha_i,t_i,c_i\}}^{(\text{cr,dv})}$ 
for simplicity, unless otherwise specified), 

\begin{equation}
	\label{equ:volume_render_crdv}
	\widehat{O}^{(\text{cr,dv})}(\mathbf{r}) = \sum_{i=1}^{k} T_i \alpha_i
	\qquad
	\widehat{D}^{(\text{cr,dv})}(\mathbf{r}) = \sum_{i=1}^{k} T_i \alpha_i t_i
	\qquad
	\widehat{C}^{(\text{cr,dv})}(\mathbf{r}) = \sum_{i=1}^{k} T_i \alpha_i c_i
\end{equation}

where $T_i=\prod_{j=1}^{i-1} (1-\alpha_{j})$ is the accumulated transmittance and $\alpha_i$ is calculated the same way as NeuS~\citep{wang2021neus} for the close-range model. $\widehat{O}^{(\text{cr,dv})}, \widehat{C}^{(\text{cr,dv})}, \widehat{D}^{(\text{cr,dv})}$ are the predicted opacity, color and depth, respectively, volume-rendered by the close-range and distant-view models jointly.


%

Finally, as shown in Equ.~\ref{equ:final_color}, we use $\widehat{c}^{(\text{sky})}(\mathbf{v})$, the colors queried from the sky model on ray direction vectors $\mathbf{v}$, as the radiance of the last and infinitely long ray marching interval, whose $\alpha$-value equals to $(1-\widehat{O}^{(\text{cr,dv})}(\mathbf{r}))$. 
In addition, the sky model is useful only if an optional set sky masks is given. Otherwise, the distant-view model can inclusively represent the sky and in this case $\widehat{C}^{(\mathrm{all})}=\widehat{C}^{(\mathrm{cr,dv})}$.


\begin{equation}
	\label{equ:final_color}
	\widehat{C}^{(\text{all})}(\mathbf{r}) = 
	\begin{cases}
		\widehat{C}^{(\text{cr,dv})}(\mathbf{r})
		+ \left(1-\widehat{O}^{(\text{cr,dv})}(\mathbf{r})\right) 	\widehat{c}^{(\text{sky})}(\mathbf{v}) & \text{w/ sky} \\
		\widehat{C}^{(\text{cr,dv})}(\mathbf{r}) & \text{w/o sky}
	\end{cases}
\end{equation}

\subsection{Cuboid space, cuboid hash-grids and hyper-cuboid distant-view models}
\label{sec:cuboid}

\begin{figure}[htbp]
	\centering
	\vspace{-5pt}
	\includegraphics[width=0.7\textwidth]{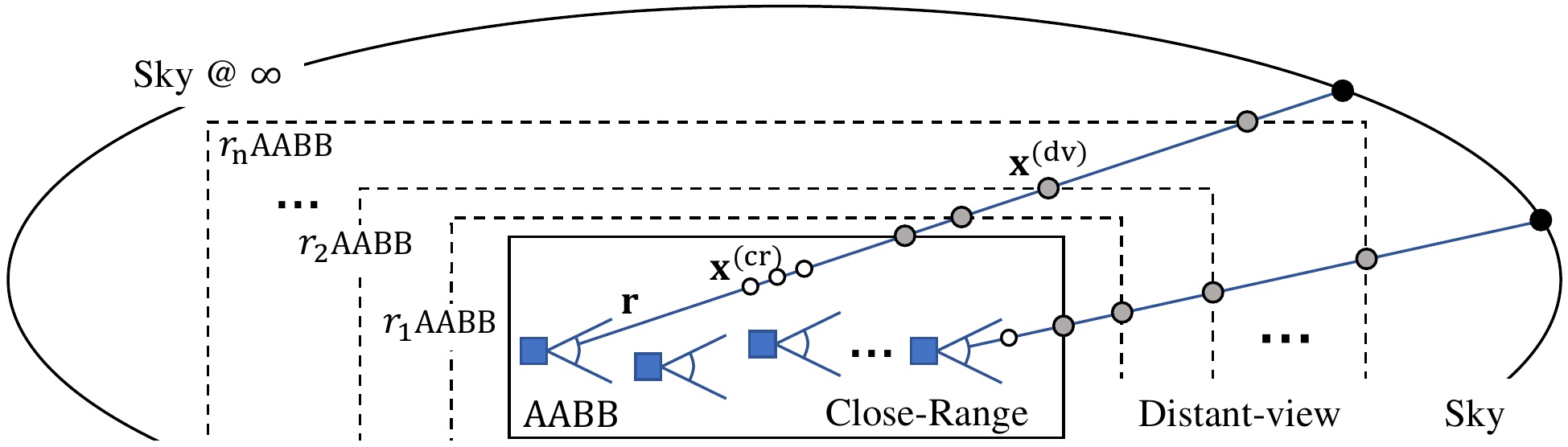}
	\par
	\vspace{-5pt}
	\caption{
		Ray sampling pattern for the close-range, distant-view and sky models. 
		We delimit the close-range space with long and narrow cuboid shapes using the camera trajectories.
		As shown by the gray points $\mathbf{x}^{(\text{dv})}$, the samples of our distant-view model lies on cuboid shells (perturbed in between when training). 
		The cuboid shells are scaled proportionally from the AABB~(Axis-Aligned Bounding Box) of the close-range space with inverse-proportionally increasing scales $\{ r_i \mid r_i=\frac{1}{ (1-i/\textrm{n}_{\text{dv}}) \cdot 1 + (i/\textrm{n}_{\text{dv}}) \cdot (1/r_{\max}) }, i \in [0, \textrm{n}_{\text{dv}}] \}$, where $r_{\max}=1000$ typically.
	}
	\label{fig:ray_sampling_all_three}
	\vspace{-15pt}
\end{figure}


Since most streets occupy long and narrow space with relatively low heights, 
defining cubic or spherical close-range boundaries along with cubic hash-grids may lead to significant capacity waste, especially for the distant-view NeRF++ model~\citep{wang2023f2nerf}. 
Therefore, we propose to delimit the close-range scene with long and narrow cuboid-shaped space that is aligned with the camera trajectories, and introduce cuboid-shell sampling and inverse cuboid warping for the distant-view model. We adapt multi-resolution 3D cuboid hash-grids for the close-range model and multi-resolution 4D hyper-cuboid hash-grids for the distant-view model.


\textbf{Cuboid space}. The cuboid close-range volume is delimited automatically according to the motion trajectory of ego car, without the need of manual specification.
To fully exploits network capacity, we first define our coordinate system that aligns its orientation about the vertical axis with that of the close-range scene.
This orientation is estimated by the average of ego motion, which roughly indicates the orientation of the street.
Then, we extend all camera frustums in each frame by a certain length, typically tens of meters, and define the AABB~(Axis-Aligned Bounding Box) of the close-range scene with the minimum and maximum points of the vertex set that contains all the vertices of the extended frustums in the aligned coordinate system.

\textbf{Cuboid close-range hash-grid}. 
Within the cuboid boundary, we assign a NeuS~\citep{wang2021neus} model to represent the scene's geometry and appearance.
For its SDF field $\widehat{\mathcal{S}}(\mathbf{x})$, we adapt a 3D hash-grid with cuboid resolution that is proportional to the size of the AABB. 
This ensures that grid voxels are equally stretched in all three dimensions, which is more appropriate for the SDF representation and its gradients (i.e. normals), and can achieve finer granularity with the same amount of parameters compared to cubic ones.
In addition, despite the presence of diagonal angles or curves in some streets, the hash-grid representation remains adaptive~\citep{muller2022instantngp} to this and incurs minimal representational waste.

\textbf{Hyper-cuboid distant-view model}. Different from NeRF++~\citep{zhang2020nerf++} that samples on spherical shells for ray marching and apply inverse spherical warping to form 4D inputs, we sample on cuboid shells, as shown in Fig.~\ref{fig:ray_sampling_all_three}, and apply inverse cuboid warping. 
Then, on top of the inversely warped 4D input space, we establish a 4D hyper-cuboid hash-grid with the first three elements of its resolution to be proportional to the size of the AABB of the close-range scene.


\subsection{Optimization}

\subsubsection{Disentanglement of close-range, distant-view and sky}
\label{sec:method:dientanglement}

\textbf{Road-surface initialization}. 
The disentanglement of close-range and distant-view models is an unsupervised and ill-posed problem with almost no constraints. 
To address this issue, we first examine different initialization schemes of the SDF field in our close-range NeuS model.

\begin{wrapfigure}{r}{0.6\textwidth}
\vspace{-15pt}
\centering
\begin{subfigure}[b]{0.28\textwidth}
	\includegraphics[width=\textwidth]{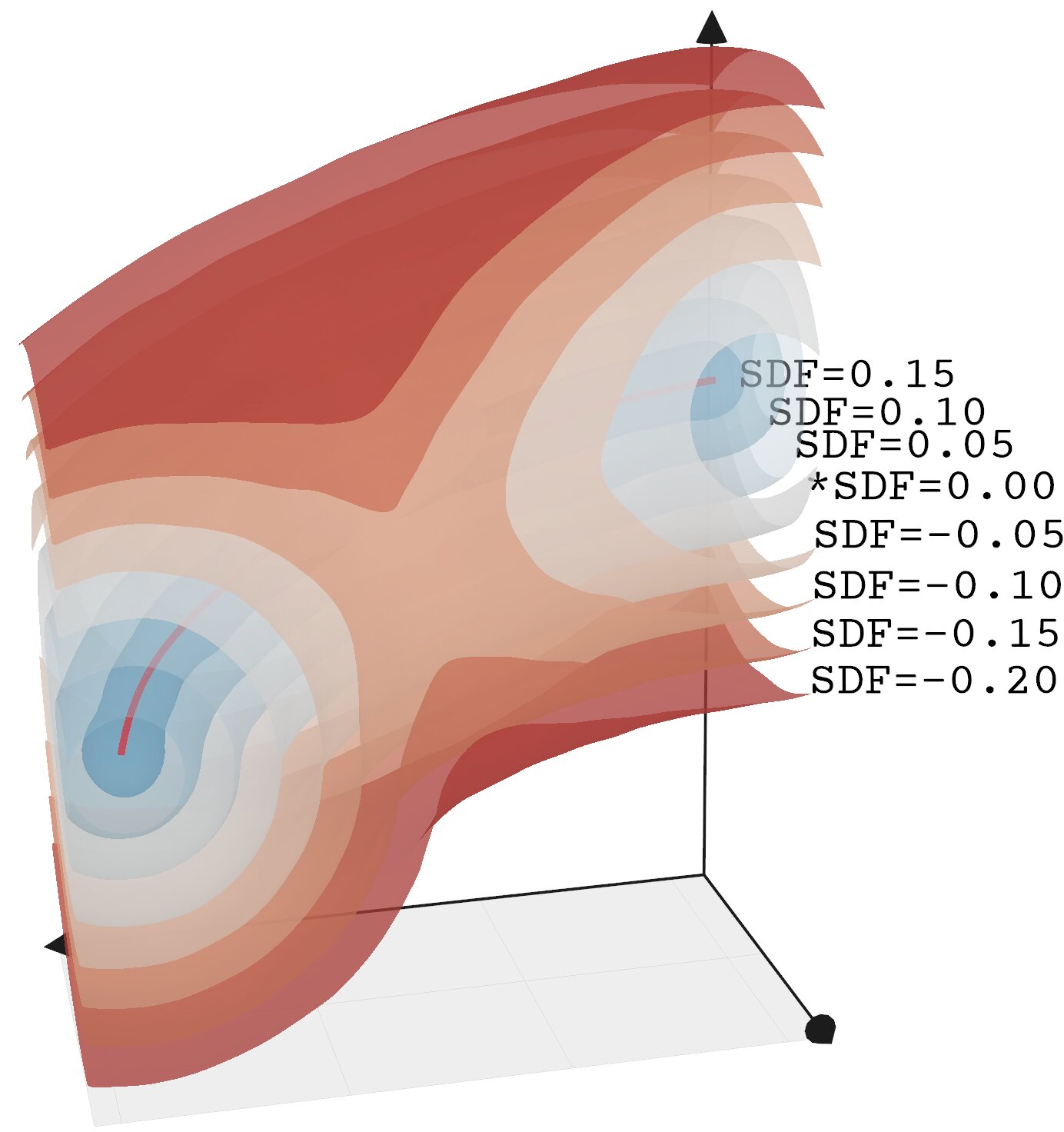}
	\caption{Capsule initialization}
	\label{fig:init_capsule}
\end{subfigure}
\hfill
\begin{subfigure}[b]{0.28\textwidth}
	\begin{minipage}[b]{\textwidth}
		\centering
		\includegraphics[width=0.7\textwidth]{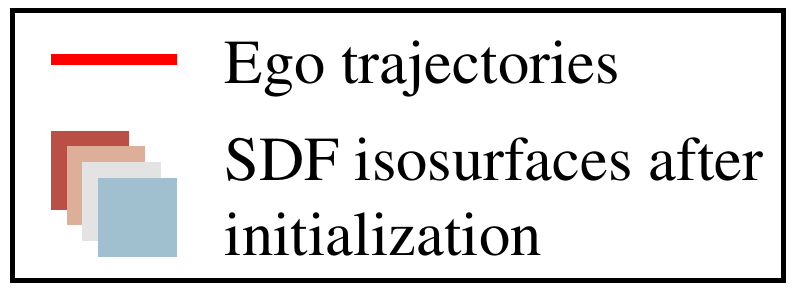} 
		\includegraphics[width=\textwidth]{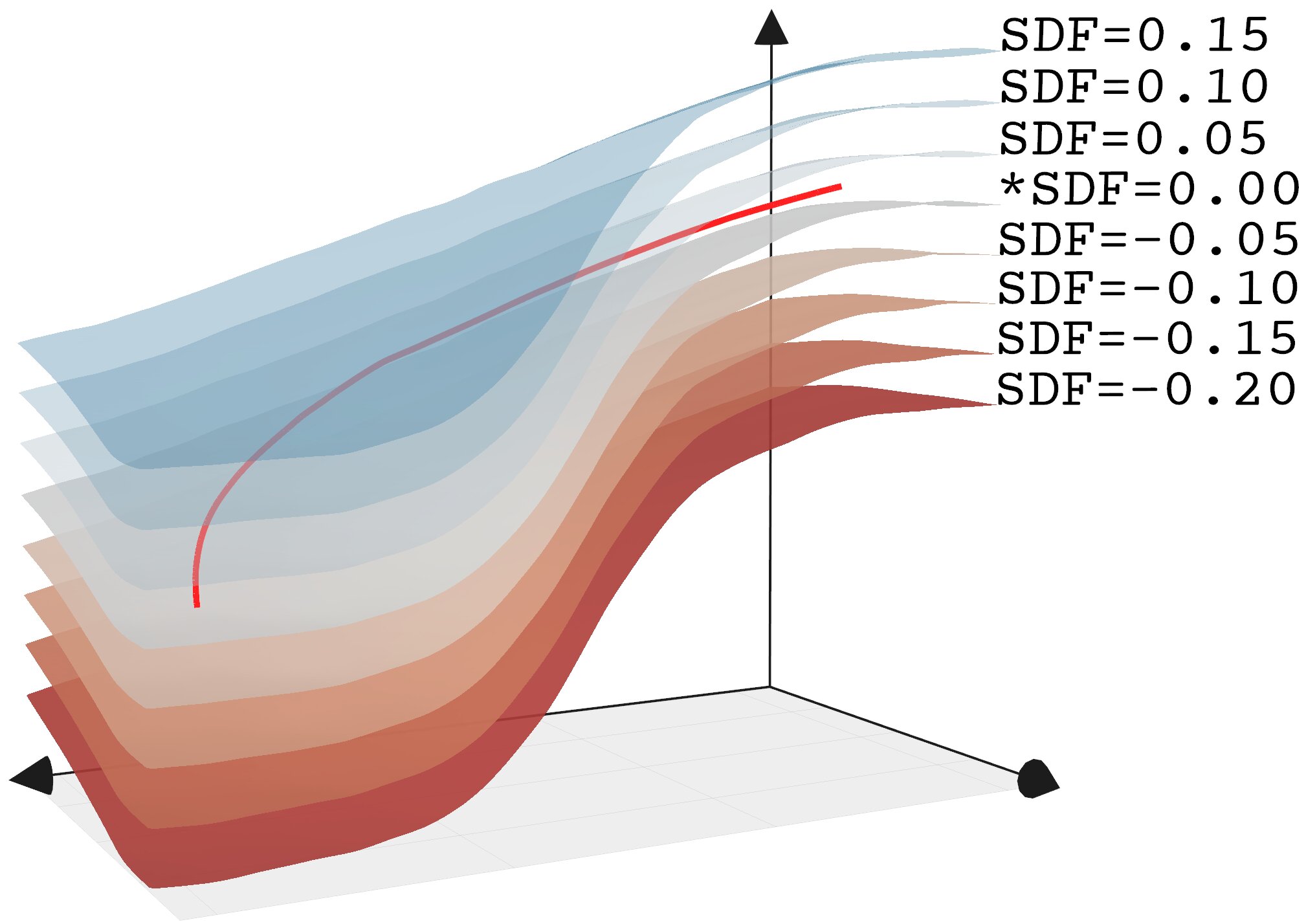}
	\end{minipage}
	\caption{Road-surface initialization}
	\label{fig:init_road_surface}
\end{subfigure}
\vspace{-5pt}
\par
\caption{
Different close-range SDF initialization schemes, taking Waymo-perception sequence \texttt{seg1434813\dots} for example. 
We mark the zero-level set with \texttt{*}. The vertical scale has been 5$\times$ exaggerated for better illustration.
}
\label{fig:init}
\vspace{-10pt}
\end{wrapfigure}	

Previous research \citep{wang2021neus,yariv2020idr} has shown that in order to achieve a proper SDF representation, it is essential to properly initialize the geometry network so that its zero-level set forms a sphere shape.
While this initialization is designed for MLPs and not directly applicable to SDFs represented by local implicits such as hash-tables, we can simply pre-train the SDF network with manually constructed pseudo ground truths following the same intuition.
Furthermore, for long and narrow camera trajectories in street views, we can construct pseudo SDF ground truths that are shaped as inside-out capsules or bent ellipsoids, which encases the ego motion tracks inside, as shown in Fig.~\ref{fig:init_capsule}. 

However, we have found that initialization schemes that form enclosed isosurfaces are not well-suited for street views, since geometries in these scenarios are usually expected to be open and not enclosed.
Subsequently, enclosed initial shapes can hinder the disentanglement of close-range and distant-view models, or often leads to failures in the convergence of sky mask.
%

Therefore, as shown in Fig.~\ref{fig:init_road_surface}, we propose a novel road-surface initialization scheme for street view reconstruction, in which the close-range SDF model is pre-trained to have a zero-level set that is roughly aligned with the road surface. This surface is automatically estimated by first horizontally extending ego motion tracks and then lowering it by an amount roughly equivalent to ego height.


\textbf{Entropy regularization loss}. 
To further eliminate ambiguous occupancy of the close-range and distant-view models, we apply an entropy regularization on $\widehat{O}^{(\text{cr})}$, the predicted occupancy of the close-range model, that encourages opaque renderings and penalizes translucent ones:

\begin{equation}
	\label{equ:loss_entropy}
	\mathcal{L}_{\text{entropy}} = f_\text{entropy}\left(\widehat{O}^{(\text{cr})}(\mathbf{r})\right), \text{where} \:  f_\text{entropy}(x) = - \left( x \ln{x} + (1-x) \ln{(1-x)} \right) 
\end{equation}

\textbf{Optional sky model and sky mask loss}. 
Given sky masks $M^{(\text{sky})}$, we explicit distinguish the sky model out of distant-view. $M^{(\text{sky})}$ is used to supervise the rendered occupancy $\widehat{O}^{(\text{cr,dv})}$ 
with $\mathcal{L}_{\text{mask}}=\text{BCE}(\widehat{O}^{(\text{cr,dv})}, 1-M^{(\text{sky})})$, 
where BCE stands for the binary cross entropy loss. 
We can efficiently use SegFormer~\citep{xie2021segformer} to extract sky masks if sky annotations are not provided.
In addition, thanks to our unsupervised disentangled design, we also support not requiring sky masks and sky model. See supplementary for more details.

\subsubsection{All supervisions}

\textbf{Photometric loss}. We use the $L_1$ photometric loss to minimize the differences between the rendered pixel color $\widehat{C}^{(\text{all})}$ and the ground truth color $C$. 
In our observation, $L_1$ loss results in less noisy reconstruction and is more beneficial for pose refinement while sacrificing a little performance in appearance, compared to the other two commonly used losses, namely MSE and Huber losses.

\begin{equation}
	\mathcal{L}_\text{photometric} = \lVert \widehat{C}^{(\text{all})} - C \rVert_1
\end{equation}

\label{sec:method:geometry_loss}
\textbf{Geometry loss}. 
To address the geometric errors arising from textureless regions and insufficient viewing angles, we develop two strategies, one for when LiDAR is available and the other for not.

If LiDAR data is available, we use the range measurements as ground truth sparse depths.
On each sampled LiDAR beam $\mathbf{r}_\text{lidar}$, we apply a logarithm $L_1$ loss on $\widehat{D}^{(\text{cr,dv})}$, the rendered depth of the combined close-range and distant-view model, as shown in Equ~\ref{equ:loss_geometry}. 

If LiDAR data is not provided, we can take inspiration from MonoSDF~\citep{Yu2022MonoSDF} and use Omnidata~\citep{eftekhar2021omnidata} to estimate the monocular normals $\bar{N}$ and depths $\bar{D}$ for each camera image. 
$\bar{N}$ and $\bar{D}$ are then used as a guide for the rendered normals $\widehat{N}^{(\text{cr})}$ and depths $\widehat{D}^{(\text{cr,dv})}$, as shown in Equ.~\ref{equ:loss_geometry},
where $\mathcal{L}_{\text{mono\_normal}}$ and $\mathcal{L}_{\text{mono\_depth}}$ is defined the same way as MonoSDF~\citep{Yu2022MonoSDF}.

\begin{equation}
	\label{equ:loss_geometry}
	\mathcal{L}_\text{geometry} = 
	\begin{cases}
		\ln ( | \widehat{D}^{(\text{cr,dv})}(\mathbf{r}_{\text{lidar}}) - D(\mathbf{r}_{\text{lidar}}) | + 1 ) & \text{if LiDAR data is provided} \\
		
		\mathcal{L}_{\text{mono\_normal}} + \lambda \mathcal{L}_{\text{mono\_depth}}
		& \text{if no LiDAR}
	\end{cases}
\end{equation}

%

\textbf{Optional sky mask loss}. 
As aforementioned in section~\ref{sec:method:dientanglement}, if an optional set of sky mask annotations or segmentation is provided, we apply a BCE mask loss $\mathcal{L}_{\text{mask}}$.

\textbf{Entropy regularization loss}. 
As aforementioned in section~\ref{sec:method:dientanglement}, we adopt an entropy regularization  $\mathcal{L}_\text{entropy}$ in Equ.~\ref{equ:loss_entropy} to encourage crisp disentanglement of the close-range and distant-view models.

\textbf{Eikonal regularization loss}. 
Following \citep{gropp2020implicit,wang2021neus}, we add an Eikonal term on the sampled points to regularize the close-range SDF network by $\mathcal{L}_\text{eikonal} = (\lVert \nabla \widehat{\mathcal{S}}(\mathbf{x}) \rVert_2 - 1)^2$.


\textbf{Sparsity regularization loss}. 
Under the settings of non-object-centric camera trajectories in street views, a large proportion of pre-allocated space are always occluded in any of the input views. 
To penalize free geometry in these unobserved regions, we take inspiration from \citep{yu2021plenoctrees,zhang2022critical} and penalizes the density values that are mapped from the uniformly sampled close-range SDF values $\widehat{\mathcal{S}}(\mathbf{x})$ using the normalized logistic density function $\text{NLD}(x)$, with a constant scaling parameter $s_{\text{reg}}$.

\begin{equation}
	\mathcal{L}_{\text{sparisty}} = \sum_{\mathbf{x}} \text{NLD}(\widehat{\mathcal{S}}(\mathbf{x})), \text{where} \: \text{NLD}(x) =
	\frac{ 1 }{ \cosh^2(\frac{x}{2s_{\text{reg}}}) }
\end{equation}

To summarize: 

\begin{equation}
    \mathcal{L}_\text{all} = \mathcal{L}_\text{photometric}
    + \lambda_\text{1} \mathcal{L}_\text{geometry}
    + \lambda_\text{2} \mathcal{L}_\text{mask}
    + \lambda_\text{3} \mathcal{L}_\text{eikonal}
    + \lambda_\text{4} \mathcal{L}_\text{sparisty}
    + \lambda_\text{5} \mathcal{L}_\text{entropy}
\end{equation}

\section{Experiments}

\subsection{Implementation details}


\begin{figure}[htbp]
\vspace{-20pt}
\begin{subfigure}{0.2\textwidth}
	\centering
	\includegraphics[width=\textwidth]{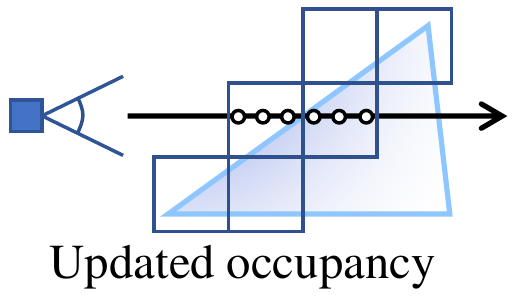}
	\caption{NGP~\citep{muller2022instantngp} samples in occupied area.}
	\label{fig:raymarch:only_occ}
\end{subfigure}
\hfill
\begin{subfigure}{0.75\textwidth}
	\centering
	\includegraphics[width=\textwidth]{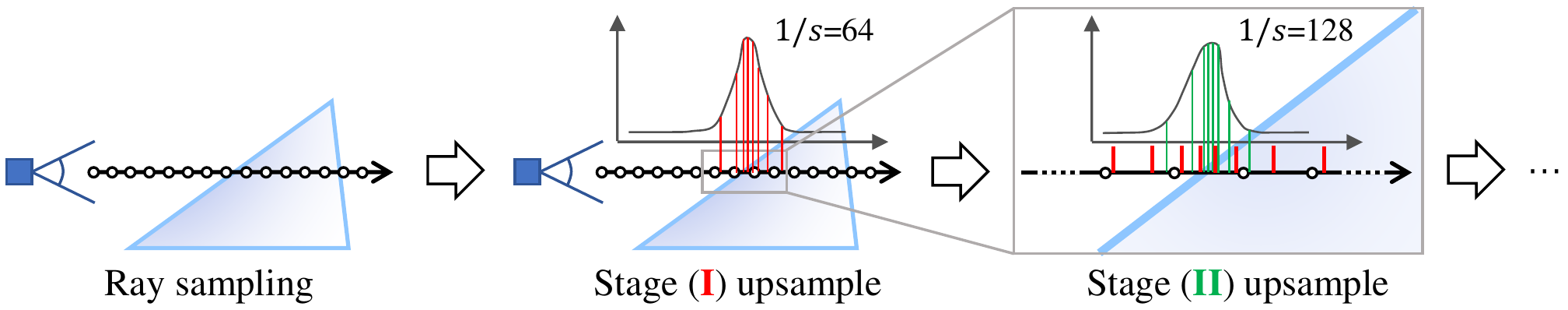}
	\caption{NeuS~\citep{wang2021neus} samples points along the entire ray from the front to the back, and then conduct multi-stage hierarchical sampling.}
	\label{fig:raymarch:only_upsample}
\end{subfigure}
\par
\begin{subfigure}{\textwidth}
	\centering
	\includegraphics[width=0.95\textwidth]{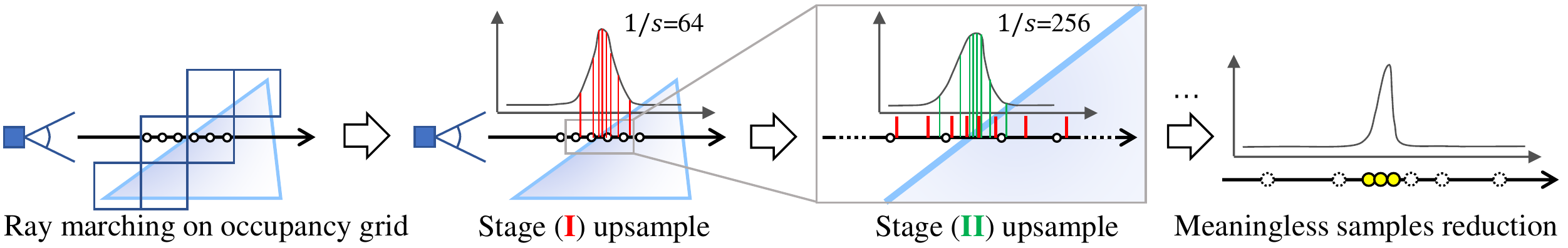}
	\caption{We propose a sampling strategy that begins with sampling in the occupied area, followed by multi-stage hierarchical sampling, and concludes with the reduction of meaningless samples.}
	\label{fig:raymarch:ours}
\end{subfigure}
\caption{Schematic illustration of different ray-marching approach. }
\end{figure}

\textbf{Ray marching with multi-stage hierarchical sampling on occupancy grids}. 
Street views occupy long and narrow spaces with complex occlusions, posing challenges to ray marching, especially if we need to balance with sampling efficiency. 
As shown in Fig.~\ref{fig:raymarch:only_occ}, NGP~\citep{muller2022instantngp} and others propose to boost ray marching efficiency with occupancy information, but its sampling granularity is not enough for long and narrow spaces due to no up-sampling. 
Shown in Fig.~\ref{fig:raymarch:only_upsample}, NeRF~\citep{mildenhall2021nerf} utilizes hierarchical sampling and NeuS~\citep{wang2021neus} introduces multi-stage hierarchical sampling in their implementation that progressively increases up-sampling granularity from one stage to the next, while sacrificing efficiency.

Standing on their shoulders, as illustrated in Fig.~\ref{fig:raymarch:ours}, we propose a novel ray marching strategy that apply multi-stage hierarchical sampling to the marched ray samples of the occupancy grid for our close-range model. 
This occupancy information is periodically updated from the close-range SDF network in a bootstrap manner that does not require reconstructing the scene in advance with COLMAP~\citep{schoenberger2016sfm} as \citep{sun2022neural-recon-w} or feeding LiDAR pointclouds to pre-compute occupied voxels as VoxSurf~\citep{li2022voxsurf}. 
To further ensure efficiency, we reduce meaningless samples for the close-range  NeuS~\citep{wang2021neus} model, similar to NGP~\citep{muller2022instantngp} and \texttt{nerf\_acc}~\citep{li2022nerfacc}. 
We implement a Pytorch~\citep{pytorch} extension library under CUDA
\ifthenelse{\boolean{@submission}}{}{named \texttt{nr3d\_lib}} 
for inference and gradient calculations of cuboid hash-grids and pack-like operations involved in the whole rendering process.
See supplementary for more details.

\textbf{Reconstruction in the wild}. 
Data captured from street views is often inaccurate and subject to complex and fluctuating illumination or other environmental conditions.
We follow the approach of \citep{tancik2022blocknerf,xie2023street-nerf,wang2021nerfmm} and use learnable pose refinement deltas to correct sensor pose errors in an unsupervised manner, which only requires photometric supervisions.
Inspired by \citep{martin2021nerfw,sun2022neural-recon-w}, we adopt per-frame image appearance embeddings in all of our close-range, distant-view and sky models to address the differences and fluctuation in exposure and color domains across various frames and cameras.




\begin{table}
\small
\setlength\tabcolsep{1pt}
\caption{Quantitative comparison on selected Waymo Open Dataset~\citep{waymo} sequences. We employ PSNR to assess the reconstructed appearance and RMSE~(See Sec.~\ref{sec:exp:lidar_simulation}) for geometry.}
\label{tab:waymo_all_comp}
\centering
\resizebox{\columnwidth}{!}{
\begin{tabular}{c||>{\columncolor{VeryLightPink}}c  @{\hskip 0.075cm}  >{\columncolor{VeryLightPink}}c  @{\hskip 0.075cm}  >{\columncolor{VeryLightPink}}c    |    >{\columncolor{VeryLightBlue}}c    >{\columncolor{VeryLightBlue}}c    >{\columncolor{VeryLightBlue}}c||    >{\columncolor{VeryLightPink}}c    >{\columncolor{VeryLightPink}}c|    >{\columncolor{VeryLightBlue}}c    >{\columncolor{VeryLightBlue}}c||    >{\columncolor{VeryLightPink}}c    >{\columncolor{VeryLightPink}}c|    >{\columncolor{VeryLightBlue}}c    >{\columncolor{VeryLightBlue}}c    >{\columncolor{VeryLightBlue}}c}

\toprule[1.5pt]

& \multicolumn{6}{c||}{\scalefont{1}With 1 dense + 4 sparse LiDARs} 
& \multicolumn{4}{c||}{\scalefont{1}With 4 sparse LiDARs} 
& \multicolumn{5}{c}{Without LiDAR data}
\\

\midrule[1pt]

\multirow{2}{*}{Sequence} 
& \multicolumn{3}{c|}{PSNR$\uparrow$}   
& \multicolumn{3}{c||}{RMSE$\downarrow$} 
& \multicolumn{2}{c|}{PSNR$\uparrow$} 
& \multicolumn{2}{c||}{RMSE$\downarrow$} 
& \multicolumn{2}{c|}{PSNR$\uparrow$} 
& \multicolumn{3}{c}{RMSE$\downarrow$} 
\\

& NeuS & {\scalefont{0.75}NGP+L} & Ours 
& NeuS & {\scalefont{0.75}NGP+L} & Ours 
& {\scalefont{0.75}NGP+L} & Ours 
& {\scalefont{0.75}NGP+L} & Ours 

& \tiny{$\text{F}^2$-NeRF} & Ours

& {\tiny{COLMAP}} & {\tiny{$\text{F}^2$-NeRF}} & Ours 
\\

\midrule[1pt]

seg1006130\dots & 14.57 & 25.15 & \textbf{27.38} & 3.14 & 1.67 & \textbf{1.22} & 23.86 & \textbf{27.26} & 3.22 & \textbf{2.58} & 24.41 & \textbf{27.39} & 7.10 & 8.87 &   \textbf{2.99} \\
seg1027514\dots & 9.10  & 23.92 & \textbf{27.19} & 3.30 & 1.94 & \textbf{1.40} & 23.48 & \textbf{27.11} & 3.25 & \textbf{2.82} & 22.14 & \textbf{27.58} & 7.47 & 16.52 &  \textbf{2.91} \\
seg1067626\dots & 12.90 & 25.94 & \textbf{27.68} & 4.40 & 2.61 & \textbf{1.94} & 24.87 & \textbf{28.68} & 7.61 & \textbf{4.12} & 26.43 & \textbf{29.01} & 9.06 & 35.59 &  \textbf{4.34} \\
seg1137922\dots & 13.75 & 25.93 & \textbf{27.64} & 5.77 & 2.56 & \textbf{2.14} & 25.15 & \textbf{27.80} & 9.62 & \textbf{5.91} & 25.42 & \textbf{28.33} & 12.39 & 20.10 & \textbf{5.70} \\
seg1172406\dots & 21.28 & 26.16 & \textbf{27.86} & 2.18 & 1.51 & \textbf{1.03} & 24.89 & \textbf{28.21} & 2.80 & \textbf{1.83} & 27.16 & \textbf{28.50} & 13.62 & 9.00 &  \textbf{2.57} \\
seg1287964\dots & 10.64 & 26.31 & \textbf{29.92} & 3.34 & 1.64 & \textbf{1.32} & 25.98 & \textbf{29.45} & 6.79 & \textbf{3.50} & 25.40 & \textbf{29.89} & 10.34 & 6.73 &  \textbf{3.19} \\
seg1308545\dots & 16.57 & 24.22 & \textbf{25.59} & 4.09 & 2.03 & \textbf{1.59} & 22.99 & \textbf{25.61} & 7.85 & \textbf{4.40} & 23.55 & \textbf{25.95} & 8.64 & 15.50 &  \textbf{4.12} \\
seg1314219\dots & 10.78 & 22.85 & \textbf{25.77} & 3.52 & 2.14 & \textbf{1.53} & 22.18 & \textbf{25.66} & 6.13 & \textbf{3.94} & 21.76 & \textbf{26.07} & 6.75 & 19.30 &  \textbf{3.48} \\
seg1319679\dots & 10.71 & 24.69 & \textbf{24.79} & 4.97 & 2.15 & \textbf{1.68} & 23.78 & \textbf{25.30} & 7.94 & \textbf{5.21} & 24.54 & \textbf{25.66} & 7.63 & 23.50 &  \textbf{4.76} \\
seg1323841\dots & 8.59  & 24.13 & \textbf{28.16} & 3.64 & 1.75 & \textbf{1.37} & 23.37 & \textbf{28.16} & 3.70 & \textbf{2.66} & 24.03 & \textbf{28.49} & 7.32 & 20.19 &  \textbf{3.13} \\
seg1347637\dots & 14.19 & 26.18 & \textbf{28.20} & 2.08 & 1.03 & \textbf{0.84} & 25.30 & \textbf{28.18} & 1.84 & \textbf{1.41} & 26.12 & \textbf{28.26} & 5.93 & 21.72 &  \textbf{1.84} \\
seg1400454\dots & 11.43 & 24.73 & \textbf{24.79} & 3.22 & 1.81 & \textbf{1.38} & 23.96 & \textbf{25.40} & 4.35 & \textbf{2.91} & 24.58 & \textbf{25.49} & 8.08 & 39.85 &  \textbf{3.29} \\
seg1434813\dots & 10.71 & 26.31 & \textbf{26.98} & 4.59 & 2.12 & \textbf{1.81} & 25.43 & \textbf{27.75} & 8.71 & \textbf{4.30} & 26.15 & \textbf{28.23} & 8.48 & 35.96 &  \textbf{4.74} \\
seg1442480\dots & 13.33 & 25.16 & \textbf{27.22} & 3.30 & 1.70 & \textbf{1.42} & 24.28 & \textbf{27.28} & 3.69 & \textbf{2.77} & 25.14 & \textbf{27.56} & 7.85 & 36.35 &  \textbf{2.97} \\
seg1486973\dots & 16.23 & 23.59 & \textbf{24.21} & 2.90 & 1.52 & \textbf{1.22} & 21.99 & \textbf{24.20} & 2.89 & \textbf{2.17} & 22.87 & \textbf{24.42} & 5.52 & 3.53 &   \textbf{2.82} \\
seg1506235\dots & 12.01 & 25.07 & \textbf{27.54} & 2.67 & 1.43 & \textbf{1.04} & 24.12 & \textbf{27.08} & 3.22 & \textbf{1.89} & 25.07 & \textbf{27.41} & 7.84 & 27.61 &  \textbf{2.40} \\
seg1522170\dots & 12.67 & 25.26 & \textbf{25.66} & 4.79 & 2.14 & \textbf{1.68} & 24.77 & \textbf{25.67} & 6.87 & \textbf{4.87} & 25.71 & \textbf{26.29} & 11.28 & 16.66 & \textbf{4.87} \\
seg1527063\dots & 17.46 & 27.04 & \textbf{28.95} & 1.92 & 1.04 & \textbf{0.76} & 26.03 & \textbf{28.76} & 2.46 & \textbf{1.55} & 26.81 & \textbf{29.06} & 2.62 & 7.82 &   \textbf{1.98} \\
seg1534950\dots & 12.35 & 25.42 & \textbf{27.26} & 2.30 & 1.34 & \textbf{1.06} & 24.58 & \textbf{27.20} & 2.77 & \textbf{1.94} & 25.30 & \textbf{27.47} & 4.31 & 7.80 &   \textbf{2.56} \\
seg1536582\dots & 17.03 & 26.35 & \textbf{27.56} & 2.42 & 1.52 & \textbf{1.07} & 24.96 & \textbf{27.43} & 2.72 & \textbf{1.96} & 26.21 & \textbf{27.69} & 6.57 & 10.41 &  \textbf{2.47} \\
seg1586862\dots & 13.35 & 24.26 & \textbf{26.71} & 3.28 & 1.63 & \textbf{1.21} & 23.20 & \textbf{26.68} & 3.37 & \textbf{2.44} & 24.29 & \textbf{26.83} & 5.94 & 18.78 &  \textbf{2.60} \\
seg1634531\dots & 17.38 & 25.68 & \textbf{26.57} & 2.38 & 1.14 & \textbf{0.99} & 24.32 & \textbf{26.50} & 2.21 & \textbf{1.85} & 26.27 & \textbf{26.82} & 5.31 & 11.85 &  \textbf{2.23} \\
seg1647019\dots & 11.35 & \textbf{22.67} & 21.26 & 5.32 & \textbf{2.42} & 3.18 & 21.98 & \textbf{23.88} & 6.08 & \textbf{4.38} & 21.90 & \textbf{24.18} & 10.36 & 12.25 & \textbf{4.31} \\
seg1660852\dots & 15.75 & 23.14 & \textbf{24.58} & 3.68 & 2.13 & \textbf{1.82} & 22.23 & \textbf{24.47} & 3.51 & \textbf{3.49} & 21.37 & \textbf{24.82} & 5.11 & 4.72 &   \textbf{3.91} \\
seg1664636\dots & 16.34 & 24.87 & \textbf{26.86} & 2.65 & 1.36 & \textbf{1.03} & 23.99 & \textbf{26.71} & 2.36 & \textbf{1.79} & 25.23 & \textbf{26.83} & 6.54 & 13.86 &  \textbf{2.26} \\
seg1776195\dots & 8.55  & 24.70 & \textbf{25.35} & 4.34 & 2.14 & \textbf{1.89} & 24.30 & \textbf{26.37} & 5.94 & \textbf{3.87} & 24.95 & \textbf{26.71} & 14.52 & 25.24 & \textbf{3.90} \\
seg3224923\dots & 17.36 & 25.97 & \textbf{27.20} & 3.27 & 1.89 & \textbf{1.43} & 24.91 & \textbf{27.20} & 3.23 & \textbf{2.86} & 26.31 & \textbf{27.26} & 5.42 & 7.16 &   \textbf{3.53} \\
seg3425716\dots & 8.48  & 27.03 & \textbf{29.42} & 3.05 & 2.01 & \textbf{1.38} & 26.18 & \textbf{29.60} & 6.85 & \textbf{2.93} & 26.33 & \textbf{29.79} & 18.81 & 30.68 & \textbf{3.00} \\
seg3988957\dots & 11.70 & 22.62 & \textbf{24.19} & 3.43 & 1.87 & \textbf{1.52} & 21.75 & \textbf{24.25} & 4.18 & \textbf{3.00} & 22.02 & \textbf{24.65} & 6.07 & 5.66 &   \textbf{3.30} \\
seg4058410\dots & 12.14 & 26.40 & \textbf{28.08} & 2.93 & 1.31 & \textbf{1.04} & 25.03 & \textbf{28.09} & 3.13 & \textbf{2.18} & 25.83 & \textbf{28.33} & 5.46 & 7.02 &   \textbf{2.62} \\
seg8811210\dots & 11.18 & 24.87 & \textbf{26.60} & 3.07 & 1.44 & \textbf{1.31} & 23.69 & \textbf{26.49} & 3.92 & \textbf{2.64} & 24.28 & \textbf{26.47} & 7.16 & 27.30 &  \textbf{3.83} \\
seg9385013\dots & 13.80 & 24.16 & \textbf{26.07} & 4.75 & 2.39 & \textbf{1.67} & 23.23 & \textbf{25.92} & 7.47 & \textbf{6.28} & 22.85 & \textbf{26.55} & 9.10 & 49.34 &  \textbf{4.52} \\

\midrule[1pt]

Average & 13.24 & 25.02 & \textbf{26.66} & 3.46 & 1.79 & \textbf{1.44} & 24.09 & \textbf{26.85} & 4.71 & \textbf{3.04} & 24.70 & \textbf{27.12} & 8.08 & 18.65 & \textbf{3.35} \\

\bottomrule[1.5pt]
\end{tabular}
}
\end{table}

\subsection{Street-view reconstruction}

\begin{figure}[htbp]
\centering
\setlength\tabcolsep{1pt}
\begin{tabular}{c@{\hskip 0.1cm}c@{\hskip 0.1cm}c}

\raisebox{0.09\textwidth}[0pt][0pt]{\rotatebox[origin=c]{90}{GT}}
& \multicolumn{2}{c}{\includegraphics[width=0.9\textwidth]{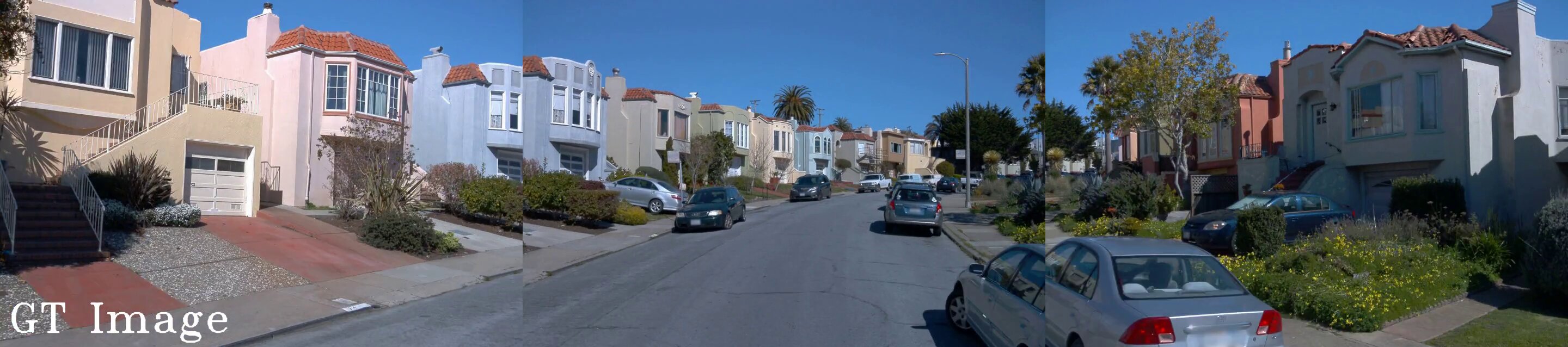}} \\[-2pt]

\raisebox{0.09\textwidth}[0pt][0pt]{\rotatebox[origin=c]{90}{Ours}} 
& \multicolumn{2}{c}{\includegraphics[width=0.9\textwidth]{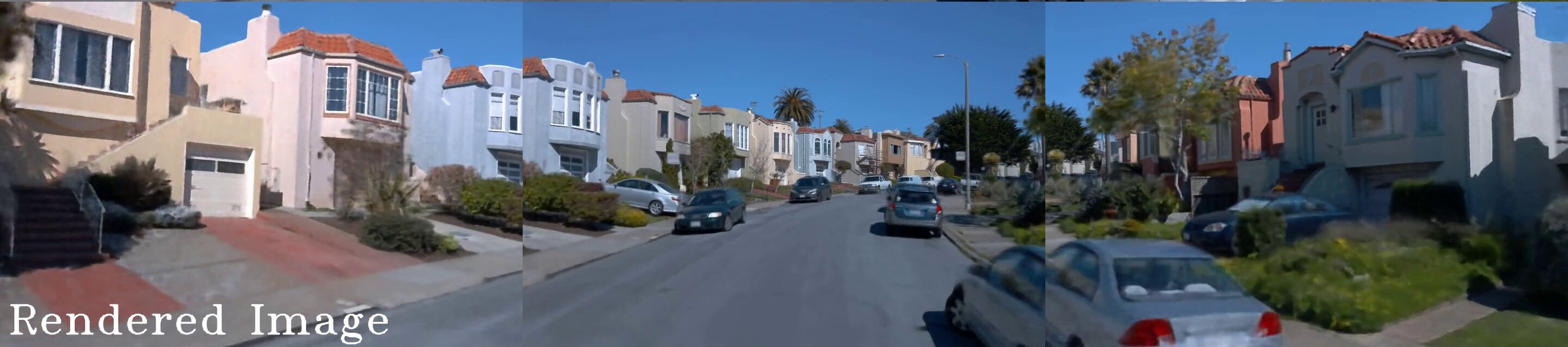}} \\[-2pt]

\raisebox{0.09\textwidth}[0pt][0pt]{\rotatebox[origin=c]{90}{$\text{F}^2$-NeRF~\citep{wang2023f2nerf}}} 
& \multicolumn{2}{c}{\includegraphics[width=0.9\textwidth]{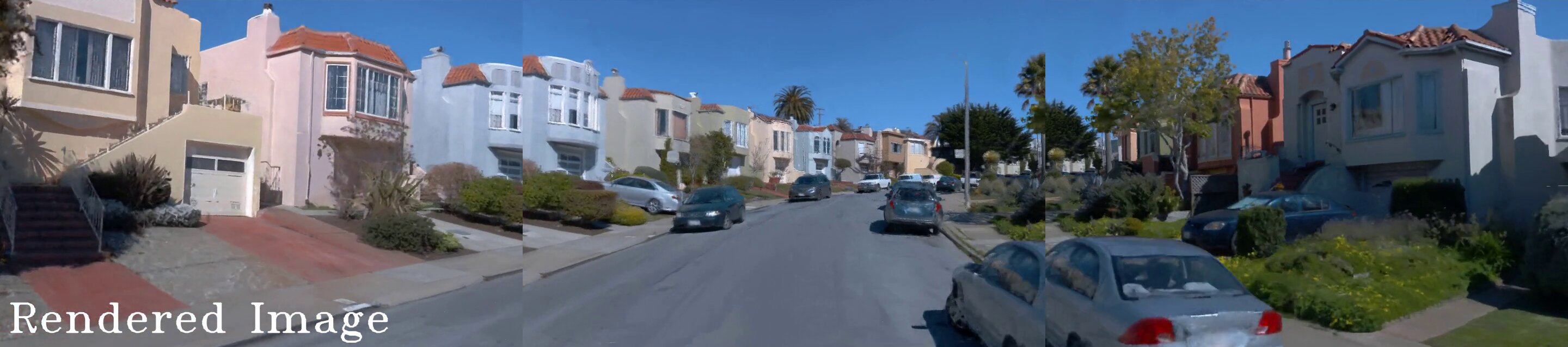}} \\[-2pt]

\raisebox{0.09\textwidth}[0pt][0pt]{\rotatebox[origin=c]{90}{Ours}} 
& \multicolumn{2}{c}{\includegraphics[width=0.9\textwidth]{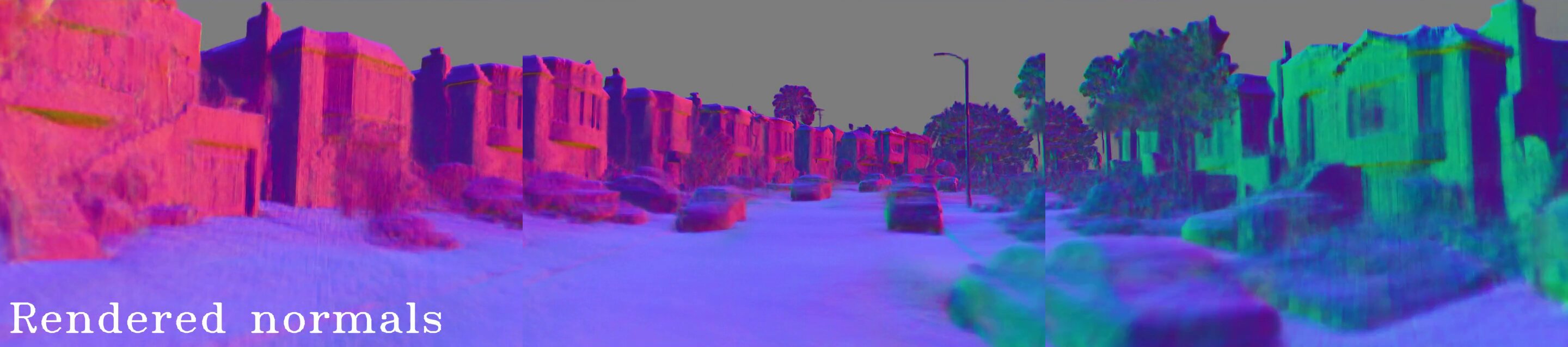}} \\[-2pt]

\raisebox{0.09\textwidth}[0pt][0pt]{\rotatebox[origin=c]{90}{Ours}} 
& \multicolumn{2}{c}{\includegraphics[width=0.9\textwidth]{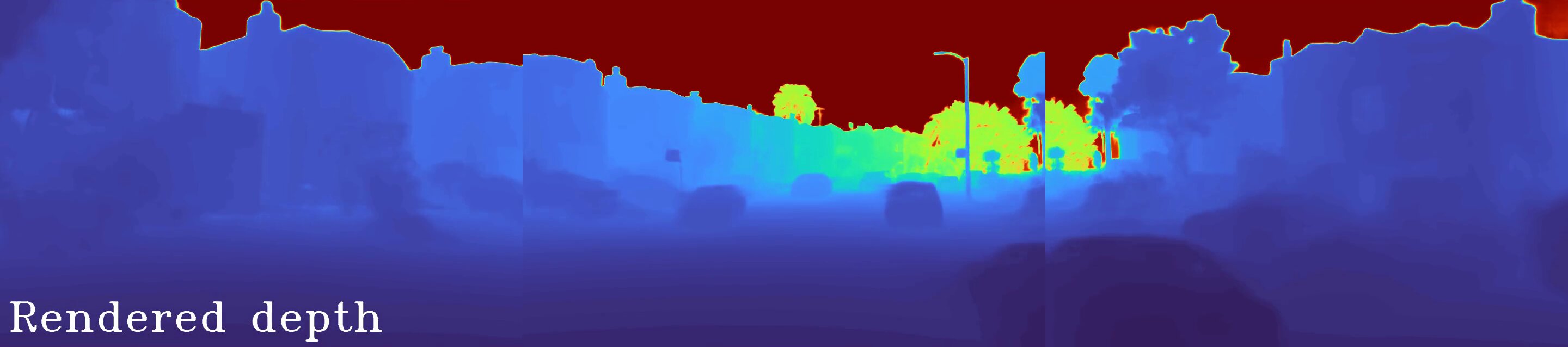}} \\[-2pt]

\raisebox{0.09\textwidth}[0pt][0pt]{\rotatebox[origin=c]{90}{$\text{F}^2$-NeRF~\citep{wang2023f2nerf}}} 
& \multicolumn{2}{c}{\includegraphics[width=0.9\textwidth]{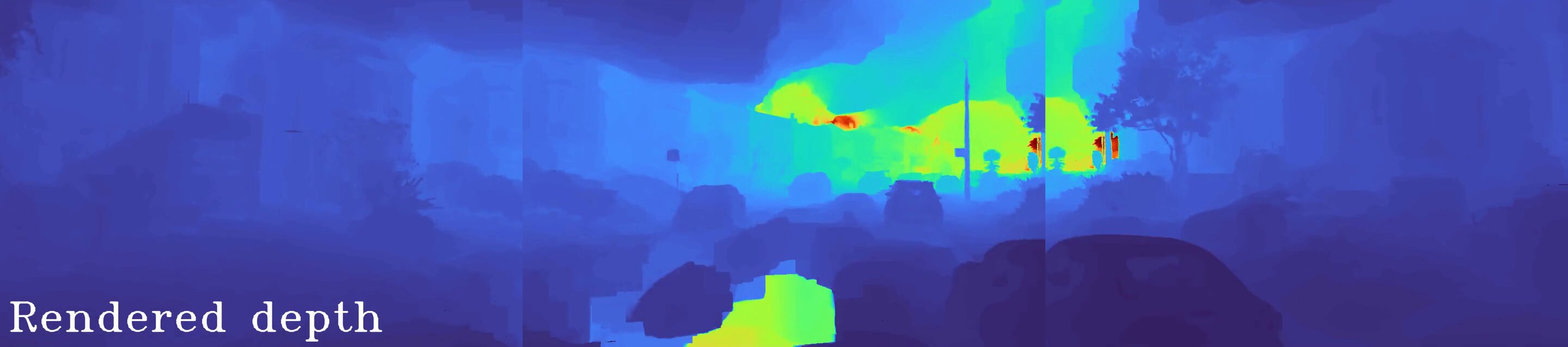}} \\

\raisebox{0.15\textwidth}[0pt][0pt]{\rotatebox[origin=c]{90}{Reconstructed surfaces}} 
& \includegraphics[width=0.45\textwidth]{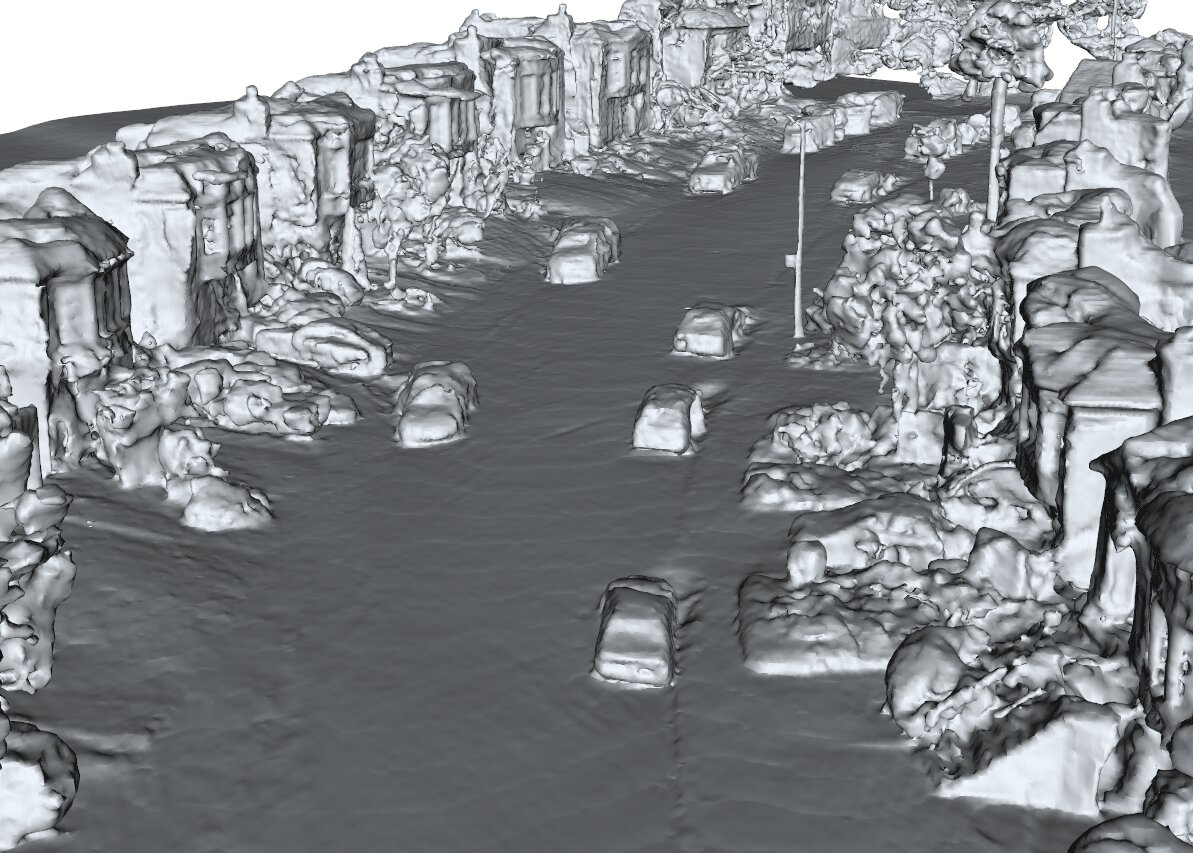}
& \includegraphics[width=0.45\textwidth]{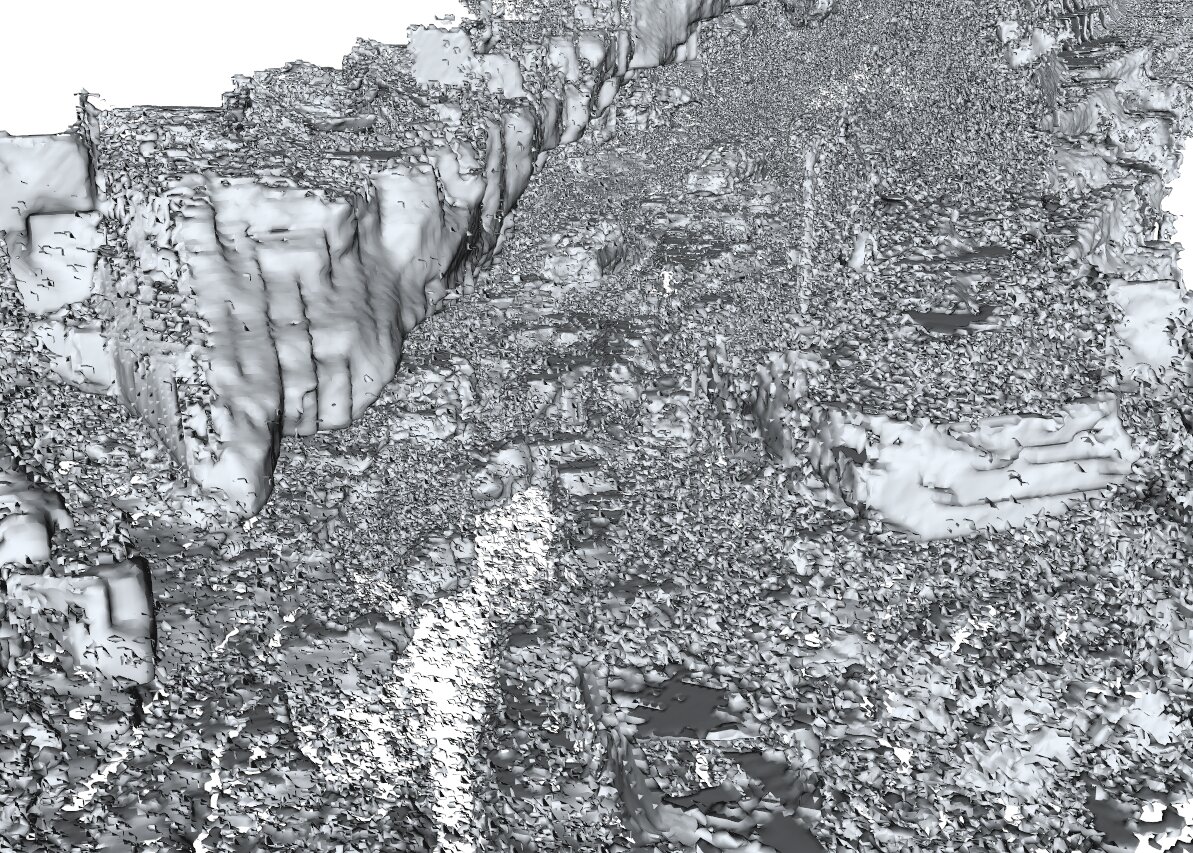} 
\\

& Ours & $\text{F}^2$-NeRF~\citep{wang2023f2nerf}

\end{tabular}
\caption{Qualitative comparison of reconstruction \textbf{without LiDAR}, on Waymo Open Dataset~\citep{waymo} seg1586862\dots.}
\label{fig:demo_withoutlidar}
\end{figure}

As shown in Tab.~\ref{tab:waymo_all_comp}, we manually select 32 sequences from the Waymo Open Dataset~\citep{waymo} to serve as a benchmark for evaluating street-view reconstruction. The criteria for choosing these sequences are to minimize the presence of dynamic vehicles and ensure an unobstructed field of view, while avoiding extremely adverse weather conditions. All of the results are conducted using street-view images captured from three frontal cameras for fair comparison. We also demonstrate example qualitative comparisons of reconstruction without LiDAR in Fig.~\ref{fig:demo_withoutlidar} and with LiDAR in Fig.~\ref{fig:demo_withlidar_all}. See more details and results in the supplementary.

%

\begin{figure}[htbp]
\setlength\tabcolsep{1pt}
\centering
\begin{tabular}{c@{\hskip 0.1cm}c}

\raisebox{0.09\textwidth}[0pt][0pt]{\rotatebox[origin=c]{90}{GT}}
& \includegraphics[width=0.9\textwidth]{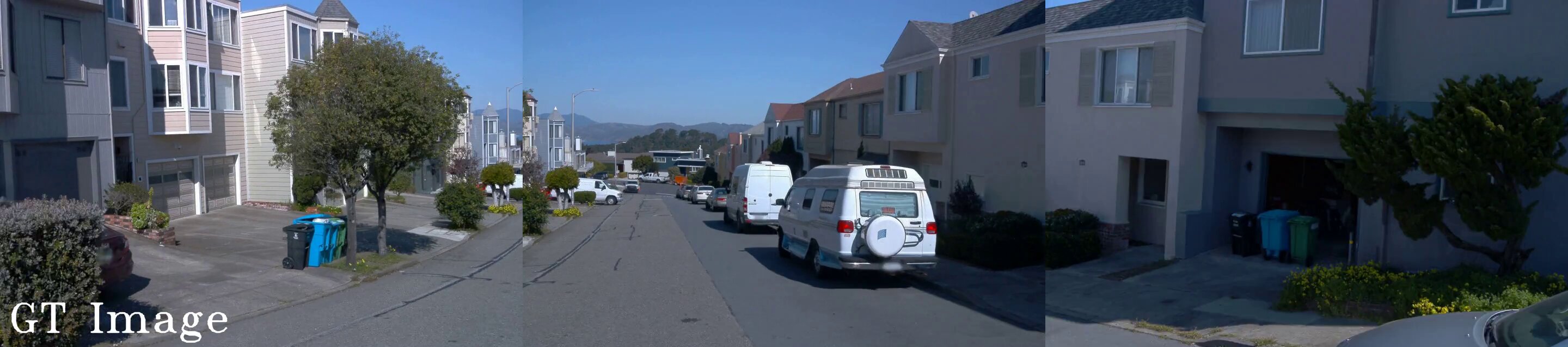} \\[-2pt]

\raisebox{0.09\textwidth}[0pt][0pt]{\rotatebox[origin=c]{90}{Ours}}
& \includegraphics[width=0.9\textwidth]{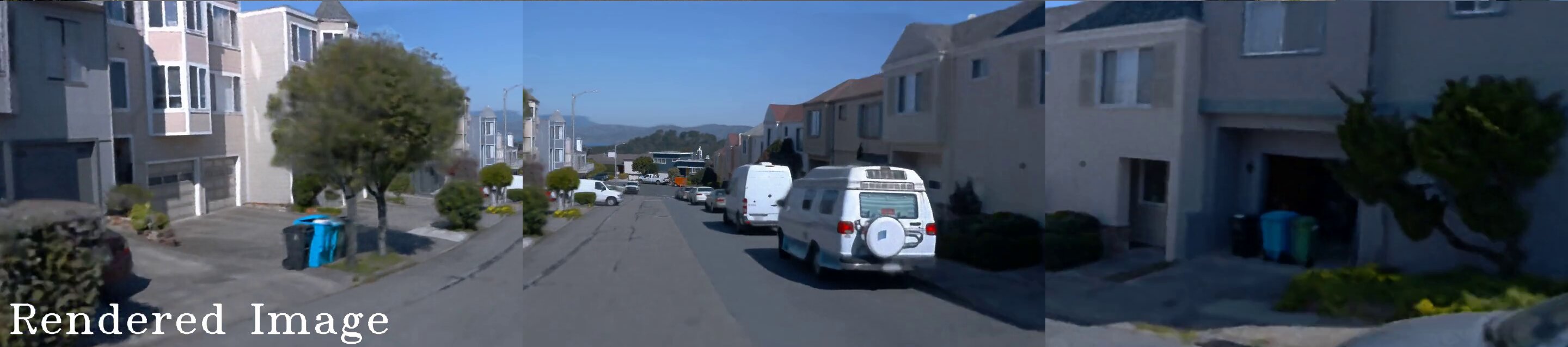} \\[-2pt]

\raisebox{0.09\textwidth}[0pt][0pt]{\rotatebox[origin=c]{90}{NGP~\citep{muller2022instantngp}+L~\citep{rematas2022urban}}}
& \includegraphics[width=0.9\textwidth]{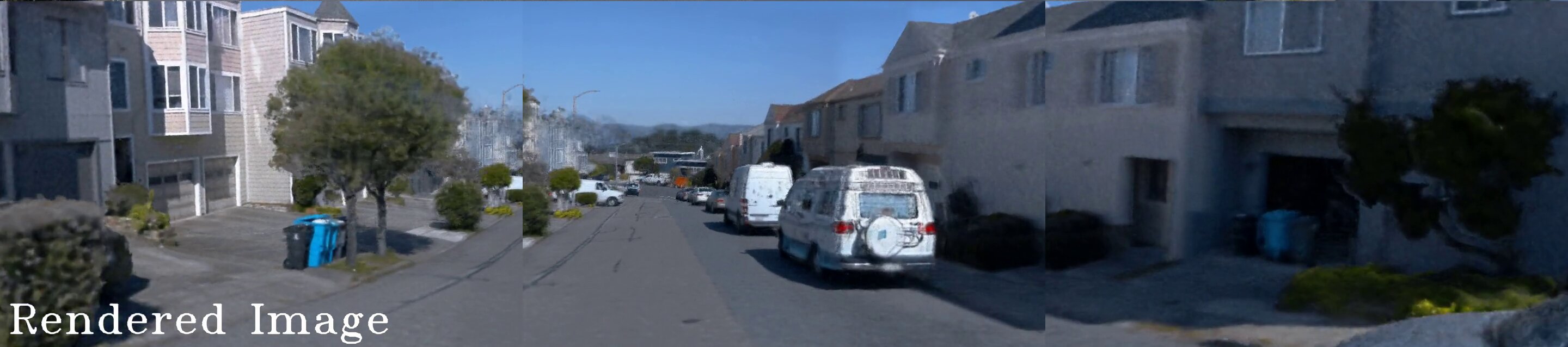} \\[-2pt]

\raisebox{0.09\textwidth}[0pt][0pt]{\rotatebox[origin=c]{90}{Ours}}
& \includegraphics[width=0.9\textwidth]{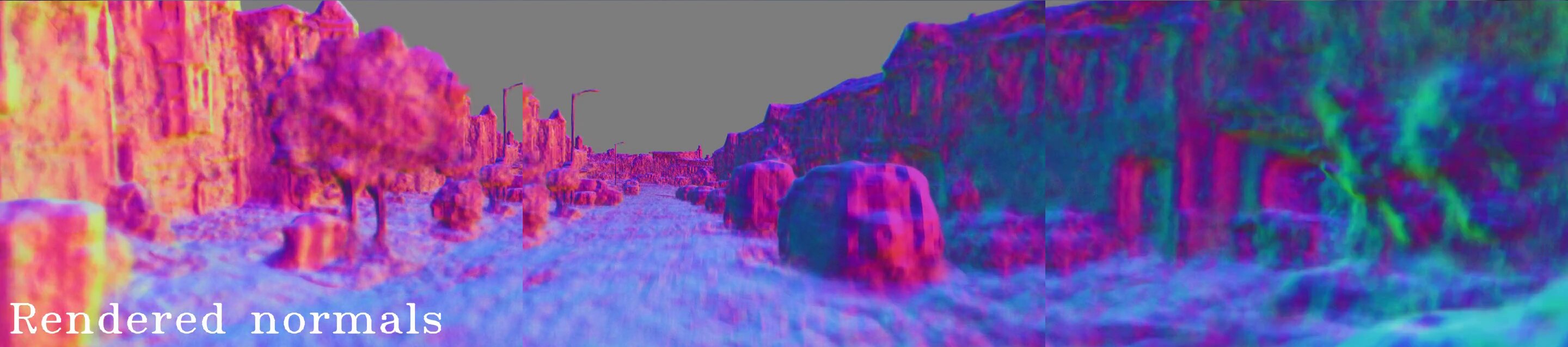} \\[-2pt]

\raisebox{0.09\textwidth}[0pt][0pt]{\rotatebox[origin=c]{90}{Ours + Normal cues}}
& \includegraphics[width=0.9\textwidth]{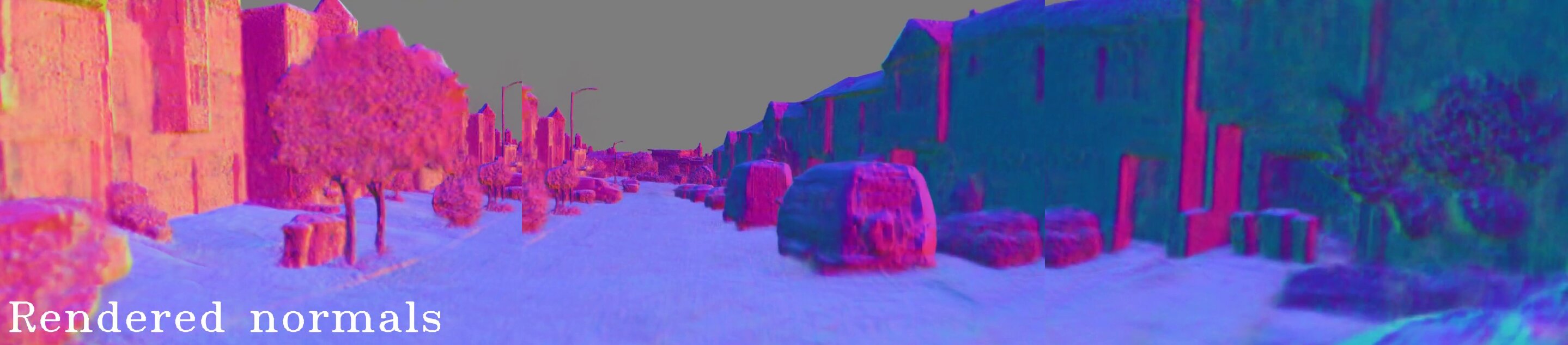} \\[-2pt]

\raisebox{0.09\textwidth}[0pt][0pt]{\rotatebox[origin=c]{90}{NeuS~\citep{wang2021neus}}}
& \includegraphics[width=0.9\textwidth]{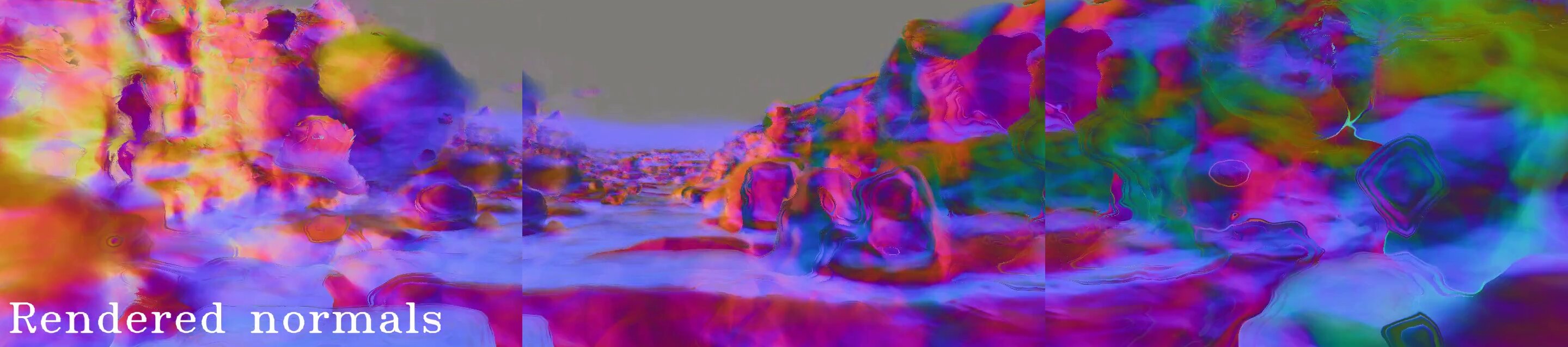} \\[-2pt]

\raisebox{0.09\textwidth}[0pt][0pt]{\rotatebox[origin=c]{90}{Ours}}
& \includegraphics[width=0.9\textwidth]{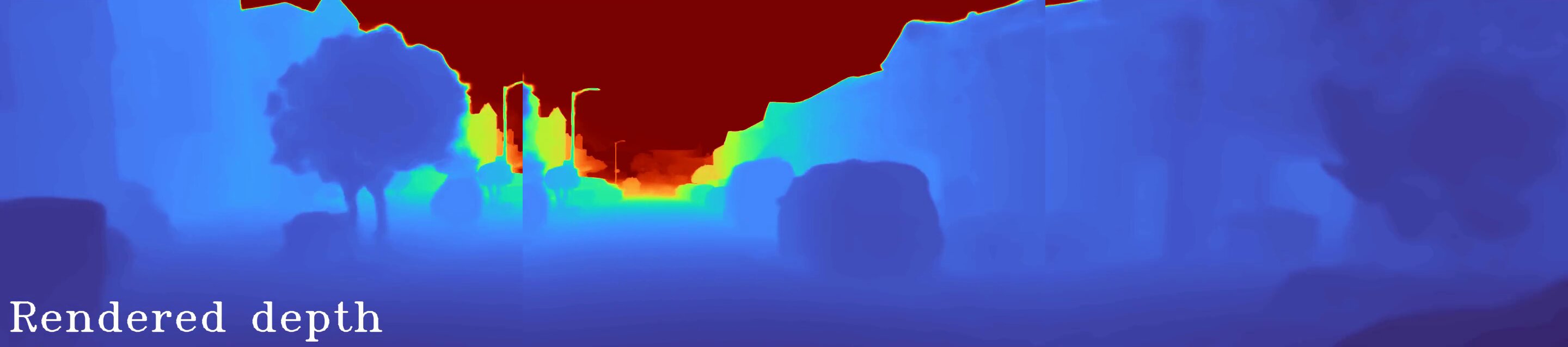} \\[-2pt]

\raisebox{0.09\textwidth}[0pt][0pt]{\rotatebox[origin=c]{90}{NGP~\citep{muller2022instantngp}+L~\citep{rematas2022urban}}}
& \includegraphics[width=0.9\textwidth]{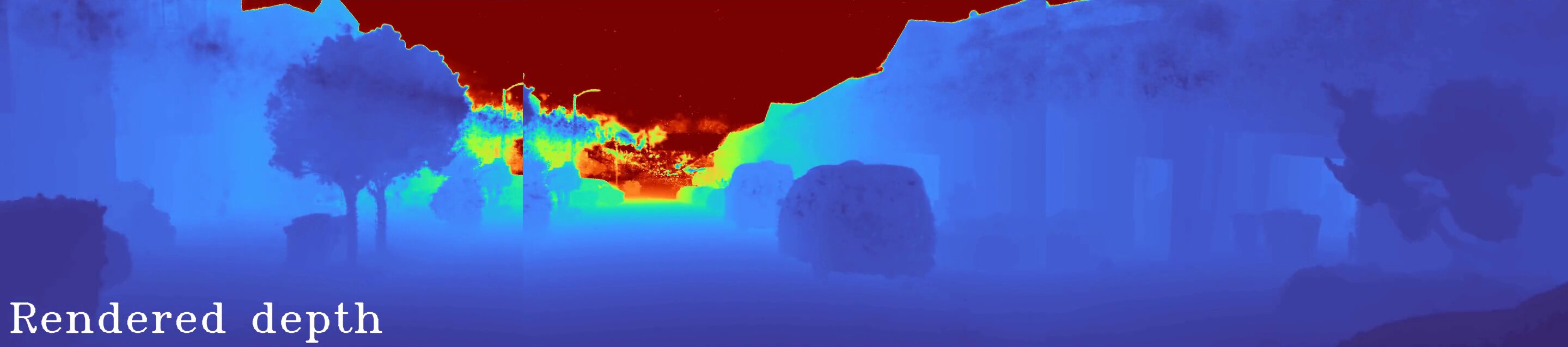}

\end{tabular}
\vspace{-5pt}
\caption{Qualitative comparison of reconstruction \textbf{with LiDAR}, using all five LiDARs, on Waymo Open Dataset~\citep{waymo} seg1006130\dots.}
\label{fig:demo_withlidar_all}
\end{figure}

\subsection{Geometric metrics}
\label{sec:exp:lidar_simulation}
Ground truth 3D scene geometry is often unavailable or difficult to obtain for street views.
However, LiDAR sensors in autonomous driving datasets can provide indirect 3D geometry information.
Despite the potential for noise and inaccuracy, having a metric is still preferable to having none at all.
Therefore, we propose to use the the Chamfer Distance (C-D)\footnote{For chamfer-distance, we utilize the implementation of Pytorch3D~\citep{pytorch3d}.} between the rendered and original LiDAR point clouds and the RMSE~(Root mean square error)~\citep{godard2019digging} of the rendered depth. See supplementary for more details on these metrics.

\subsection{Ablation}

\begin{figure}[htbp]
	\small
	\setlength\tabcolsep{1pt}
	\centering
	\begin{tabular}{c@{\hskip 0.1cm}c@{\hskip 0.1cm}c@{\hskip 0.1cm}c@{\hskip 0.15cm}c}
		\raisebox{0.07\textwidth}[0pt][0pt]{\rotatebox[origin=c]{90}{Disentangled}}
		& \includegraphics[width=0.235\textwidth]{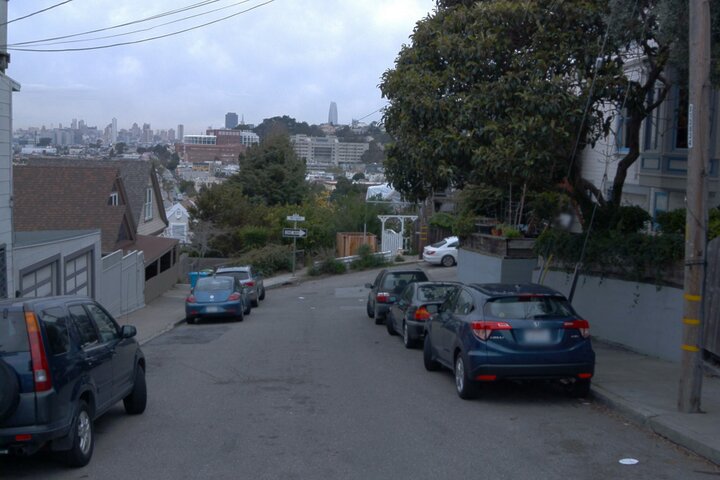}
		& \includegraphics[width=0.235\textwidth]{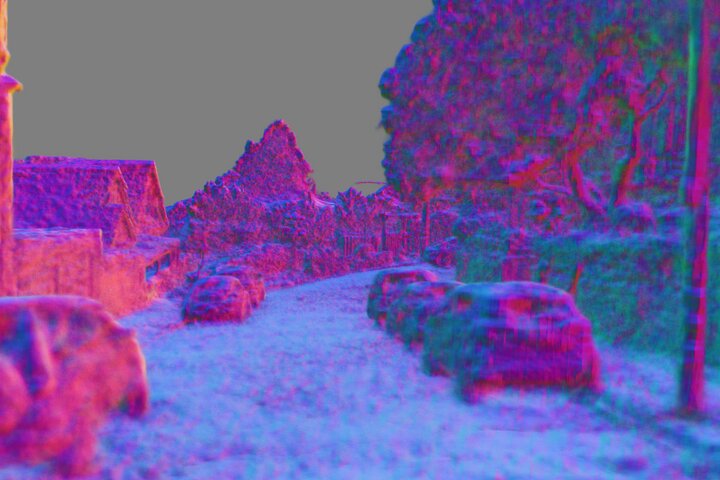}
		& \includegraphics[width=0.235\textwidth]{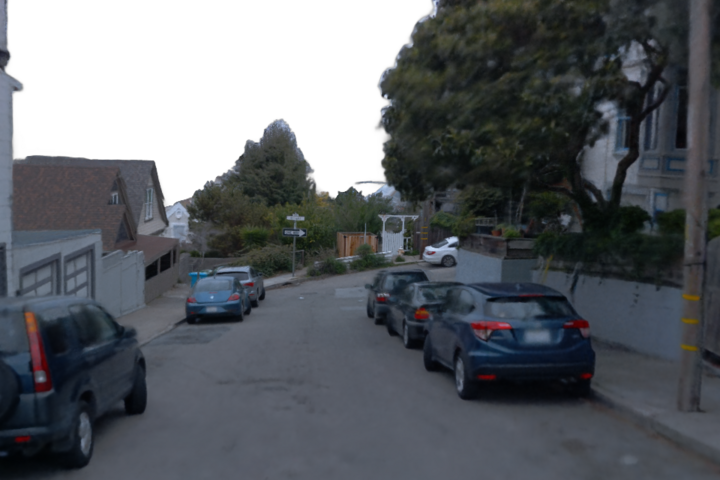}
		& \includegraphics[width=0.235\textwidth]{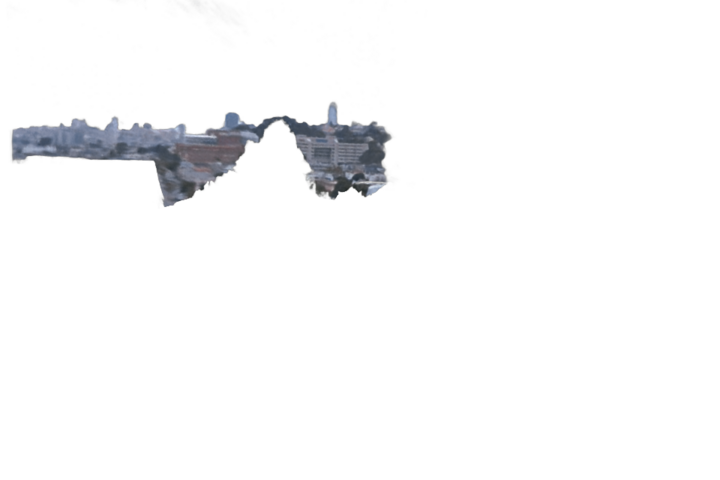} \\
		
		\raisebox{0.07\textwidth}[0pt][0pt]{\rotatebox[origin=c]{90}{Only close-range}}
		& \includegraphics[width=0.235\textwidth]{figs/crdv/crdv_134763_withdistant_9/rgb_gt.jpg}
		& \includegraphics[width=0.235\textwidth]{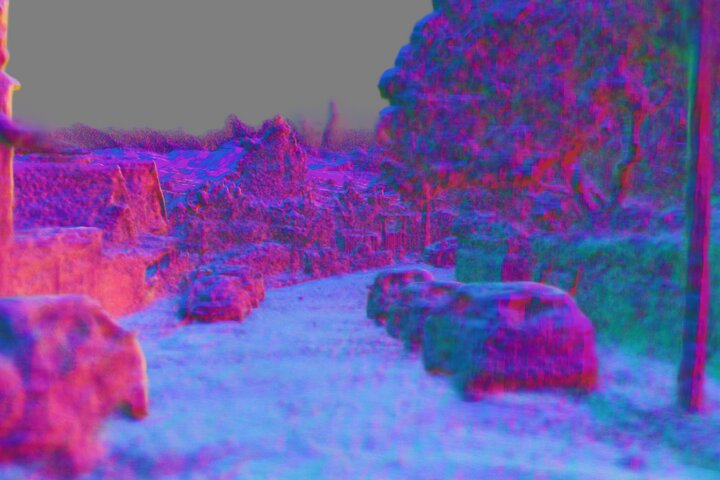}
		& \includegraphics[width=0.235\textwidth]{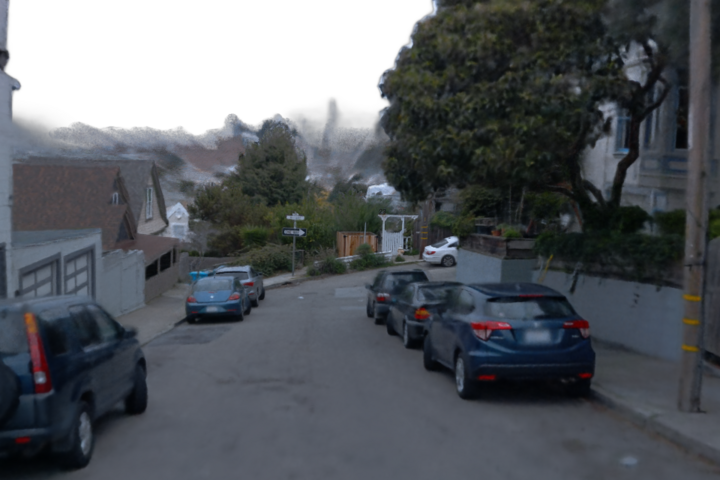}
		&  \\
		
		\multicolumn{5}{c}{Waymo Open Dataset~\citep{waymo} seg1347637\dots} \\[0.1cm]
		
		\raisebox{0.07\textwidth}[0pt][0pt]{\rotatebox[origin=c]{90}{Disentangled}}
		& \includegraphics[width=0.235\textwidth]{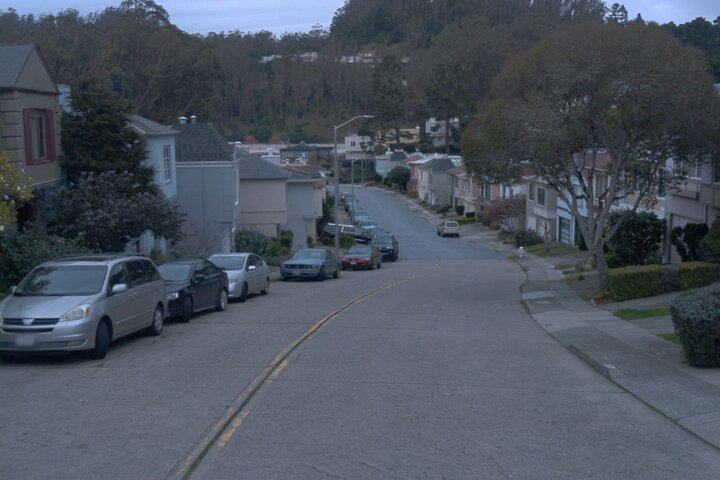}
		& \includegraphics[width=0.235\textwidth]{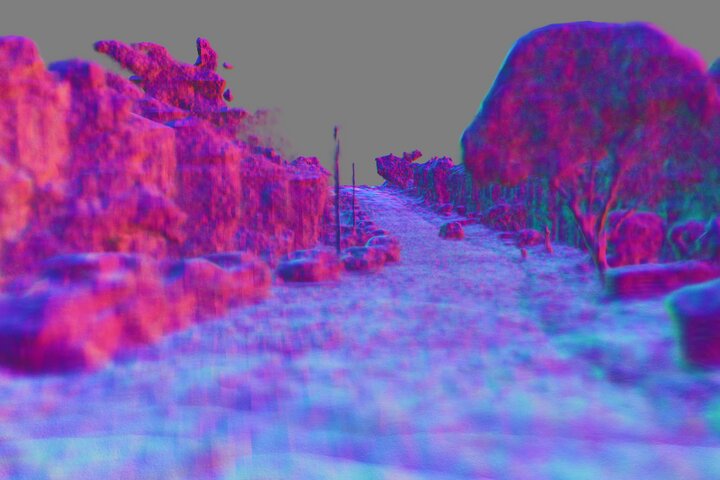}
		& \includegraphics[width=0.235\textwidth]{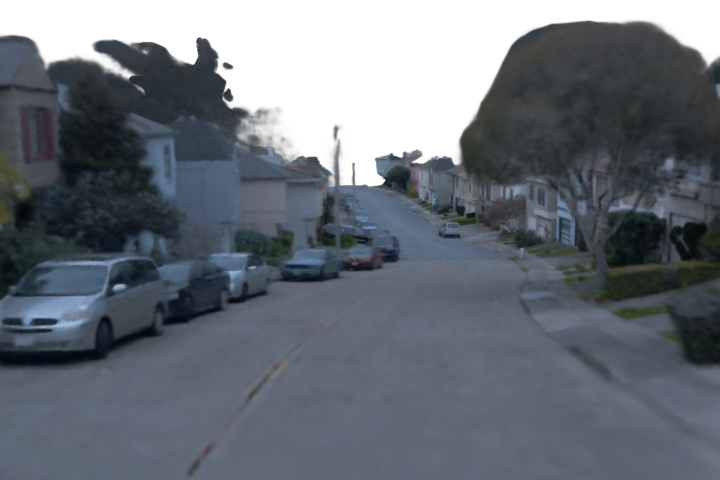}
		& \includegraphics[width=0.235\textwidth]{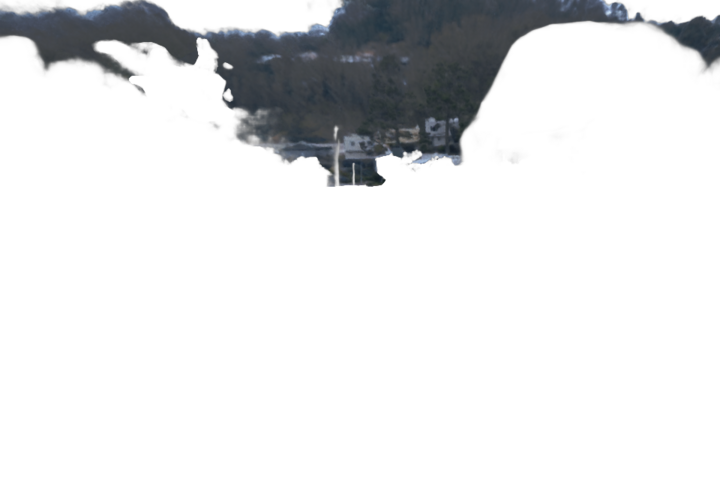} \\
		
		\raisebox{0.07\textwidth}[0pt][0pt]{\rotatebox[origin=c]{90}{Only close-range}}
		& \includegraphics[width=0.235\textwidth]{figs/crdv/crdv_117240_withdistant_75/rgb_gt.jpg}
		& \includegraphics[width=0.235\textwidth]{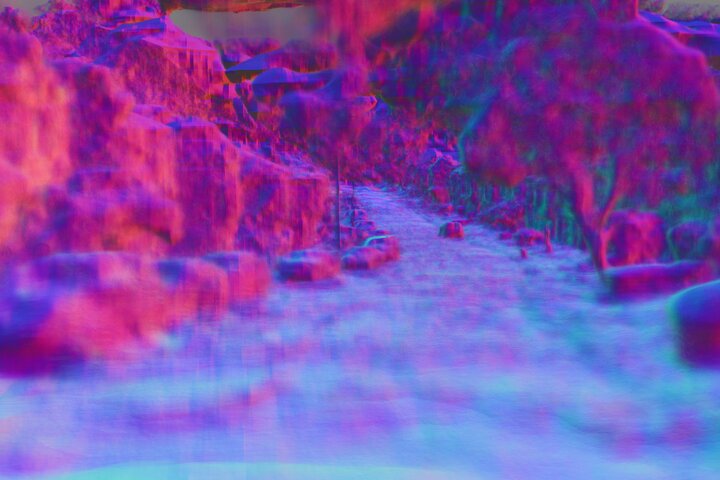}
		& \includegraphics[width=0.235\textwidth]{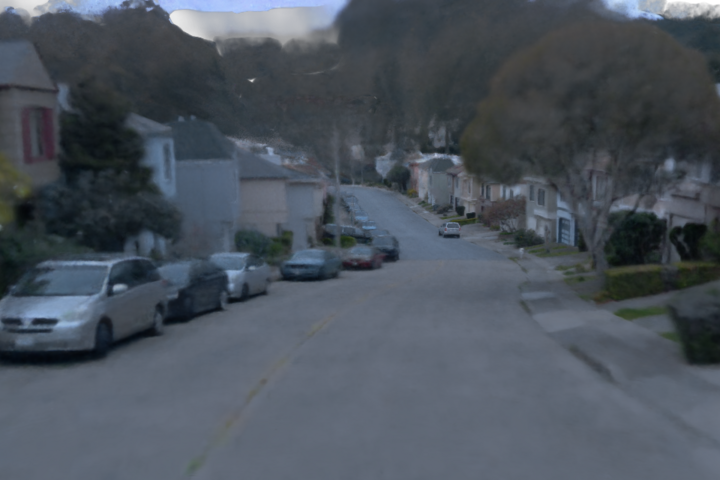}
		&  \\
		
		\multicolumn{5}{c}{Waymo Open Dataset~\citep{waymo} seg1172406\dots} \\[0.1cm]
		
		\raisebox{0.07\textwidth}[0pt][0pt]{\rotatebox[origin=c]{90}{Disentangled}}
		& \includegraphics[width=0.235\textwidth]{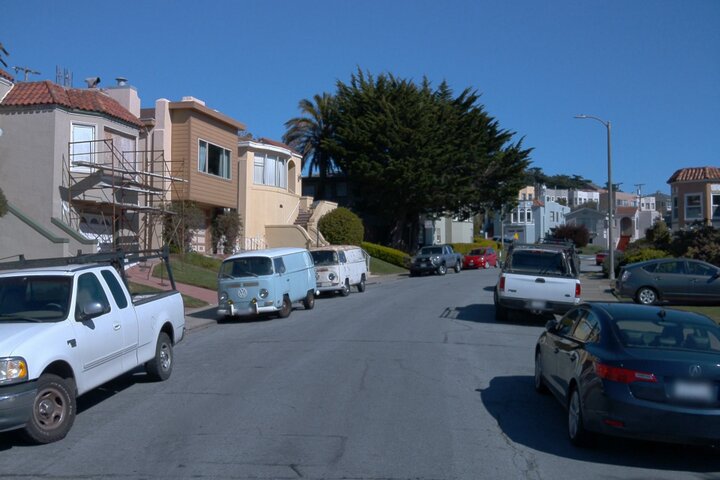}
		& \includegraphics[width=0.235\textwidth]{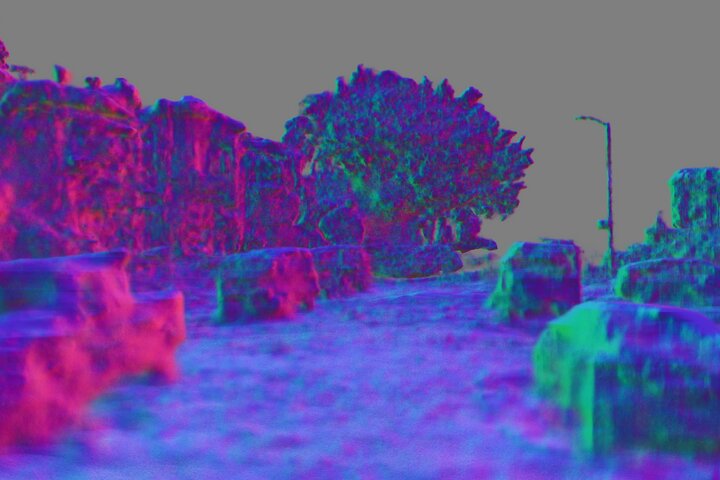}
		& \includegraphics[width=0.235\textwidth]{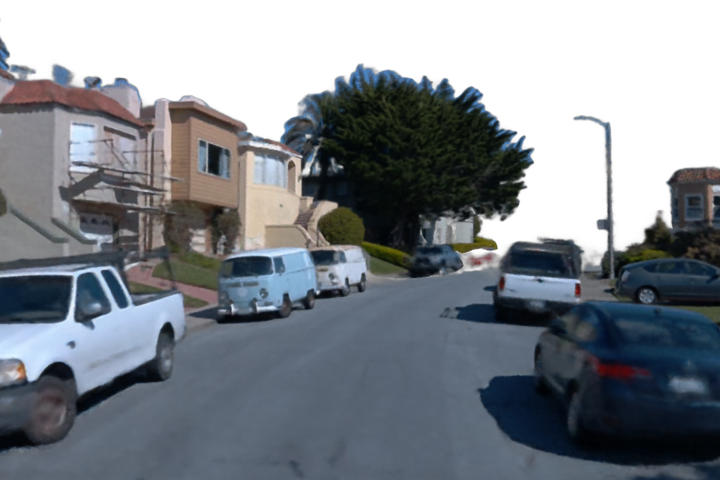}
		& \includegraphics[width=0.235\textwidth]{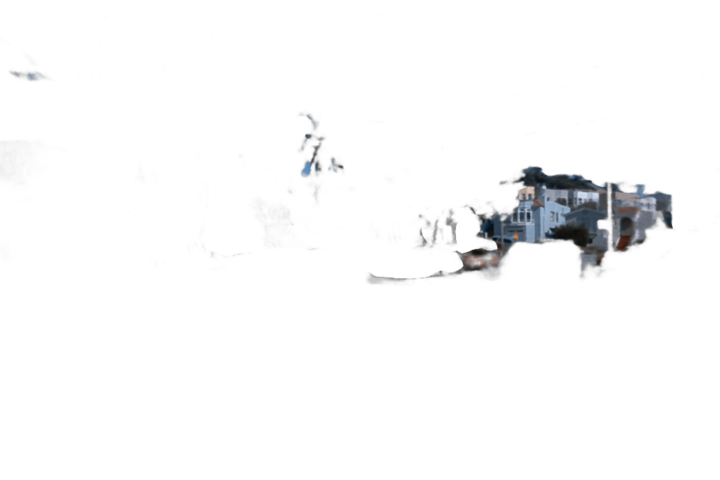} \\
		
		\raisebox{0.07\textwidth}[0pt][0pt]{\rotatebox[origin=c]{90}{Only close-range}}
		& \includegraphics[width=0.235\textwidth]{figs/crdv/crdv_158686_withdistant_120/rgb_gt.jpg}
		& \includegraphics[width=0.235\textwidth]{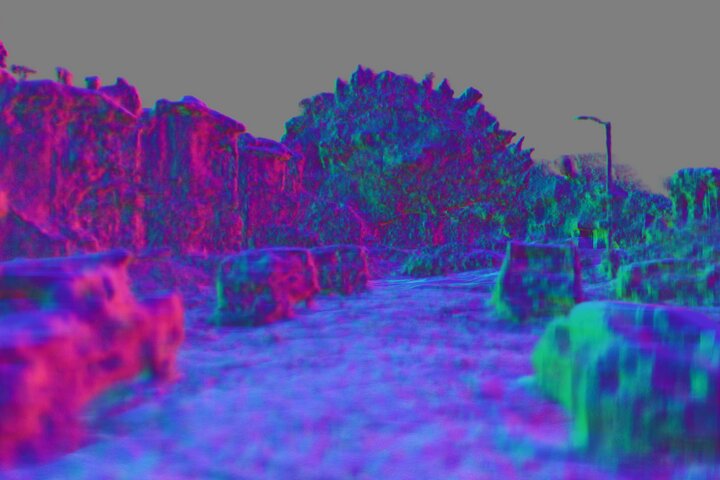}
		& \includegraphics[width=0.235\textwidth]{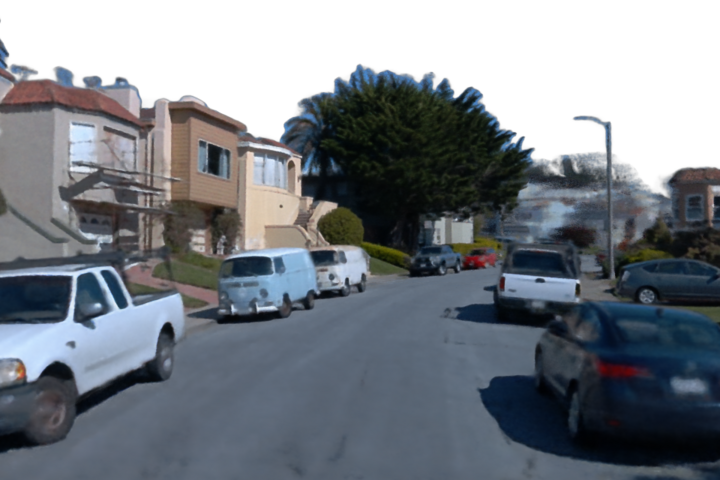}
		&  \\
		
		\multicolumn{5}{c}{Waymo Open Dataset~\citep{waymo} seg1586862\dots.} \\
		
		& Ground truth image & Predicted normals & Predicted close-range & Predicted distant-view \\
		
	\end{tabular}
	\caption{Ablation study of reconstruction using disentangled close-range and distant-view models against that using the close-range model alone. In Waymo seg1347637\dots and seg1172406\dots, distant-view space includes distant landscapes, whereas in seg1586862\dots and seg1172406\dots the distant-view also represents scenes placed at the end of capture journey.}
	\label{fig:ablation:crdv}
\end{figure}

\textbf{Disentangled vs. using only close-range model}. 
The distant-view model mainly accounts for frontal or lateral distant landscapes, as well as objects that extend beyond the end of the capture trajectory, both of which are not distinguishable even using semantic segmentation. 
But with our approach, they can be disentangled with no supervisions, as shown by the "Disentangled" section in Fig.~\ref{fig:ablation:crdv}.

\begin{figure}[htbp]
	\setlength\tabcolsep{1pt}
	\centering
	\begin{tabular}{c@{\hskip 0.1cm}c@{\hskip 0.3cm}c}
		\raisebox{0.135\textwidth}[0pt][0pt]{\rotatebox[origin=c]{90}{seg1006130\dots}}
		& 
		\includegraphics[width=0.35\textwidth]{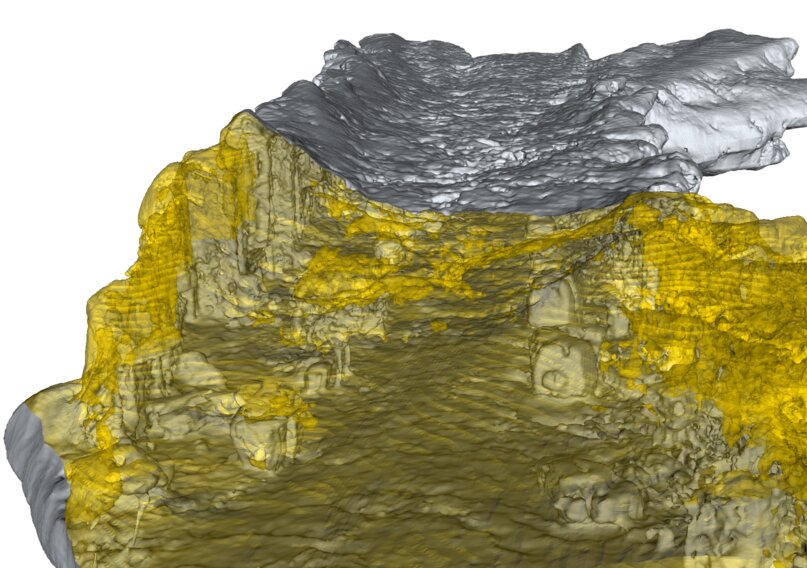}
		&
		\includegraphics[width=0.35\textwidth]{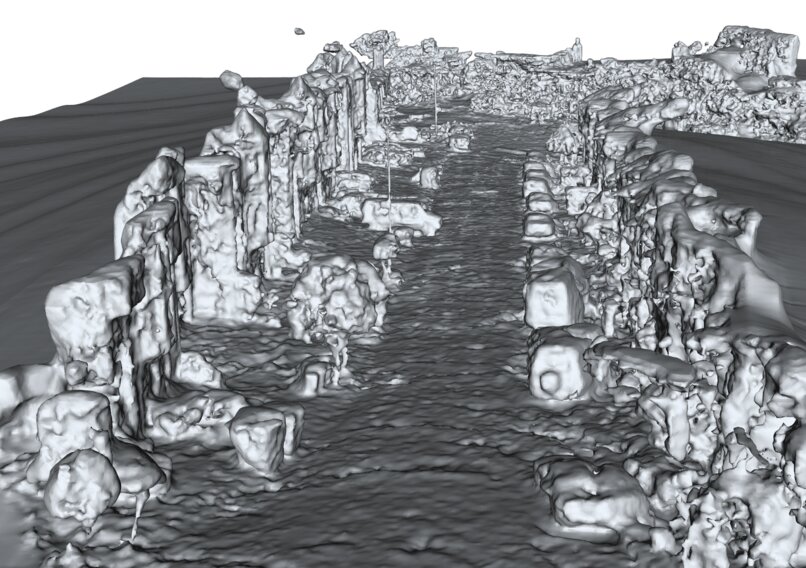}
		\\
		\raisebox{0.135\textwidth}[0pt][0pt]{\rotatebox[origin=c]{90}{seg1347637\dots}}
		& 
		\includegraphics[width=0.35\textwidth]{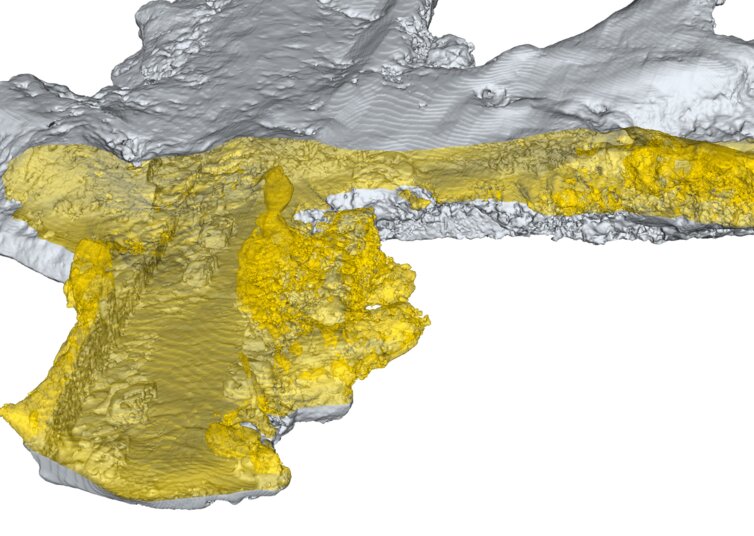}
		&
		\includegraphics[width=0.35\textwidth]{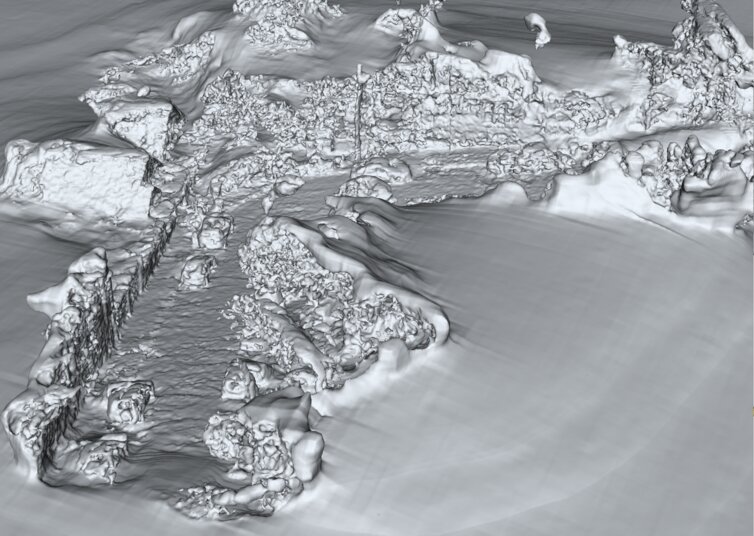}
		\\
		& 
		\begin{subfigure}{0.35\textwidth}
			\caption{Reconstructed surfaces using capsule initialization.}
		\end{subfigure}
		&
		\begin{subfigure}{0.35\textwidth}
			\caption{Reconstructed surfaces using our proposed road-surface initialization.}
		\end{subfigure}
	\end{tabular}
	\caption{
		Qualitative comparison of the reconstructed surfaces using capsule initialization (left) and using our proposed road-surface initialization (right). 
		We cut and paint a transparent yellow region on the reconstructed surface using capsule initialization for better illustration.
	}
	\label{fig:ablation:init}
\end{figure}

\textbf{Road-surface vs. enclosed initialization schemes}. As shown in Fig.~\ref{fig:ablation:init}, compared to our road-surface initialization scheme, enclosed initial shapes may lead to failed disentanglement when training, in which the reconstructed close-range geometry encases all ego trajectories and is completely enclosed.

\newpage

\begin{figure}[htbp]
	\centering
	\includegraphics[width=0.4\textwidth]{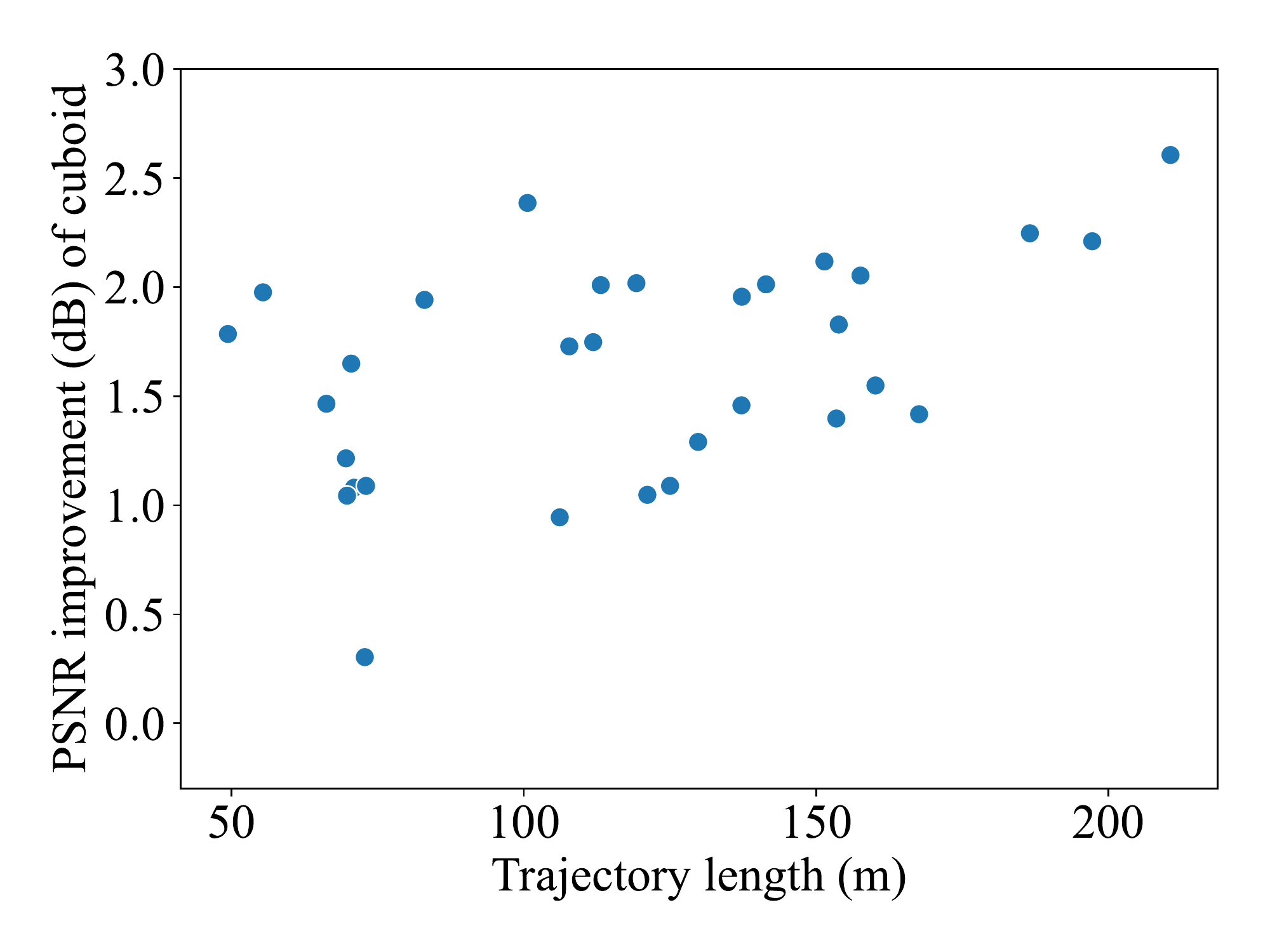}
	\caption{Scatter plot showcasing the roughly positive correlation of PSNR improvement with respect to trajectory length when we change from cubic to the proposed cuboid spaces and hash-grids.}
	\label{fig:ablation:cuboid}
\end{figure}

\begin{wraptable}{r}{0.5\textwidth}
\setlength\tabcolsep{1pt}
\caption{Average results of the selected Waymo sequences: a comparison between utilizing cuboid space and models versus cubic ones.}
\label{tab:ablation:cuboid}
\centering
\begin{tabular}{c||c|c}
\toprule[1pt]
\multirow{2}{110pt}{Hash-grid~parameter~size (cr+dv,~number~of~cr~levels)}
& Cubic & Cuboid \\

& PSNR$\uparrow$ & PSNR$\uparrow$ \\
\midrule[0.5pt]
16+8 Mi, 9-10 levels & 24.49 & 26.14 \\
32+16 Mi, 17-18 levels & 26.73 & 26.86 \\
\bottomrule[1pt]
\end{tabular}
\vspace{-5pt}
\end{wraptable}

\textbf{Cuboid vs. cubic}. 

In Sec.~\ref{sec:cuboid}, we delimit the close-range space using automatically computed cuboid boundaries that align with the long and narrow camera trajectories. 
We report the results of comparing this cuboid boundary and model with cubic ones in Tab.~\ref{tab:ablation:cuboid}. It can be observed that cuboid boundaries and hash-grids prove to be more fine-grained and achieve comparable performance with that of cubic ones while using only half of the parameter size and hash-grid levels.

As illustrated in Fig.~\ref{fig:ablation:cuboid}, we also provide a detailed scatter plot demonstrating the improvement of PSNR performance with respect to trajectory length when switching from the original cubic boundaries and hash-grids to our proposed cuboid ones. The results indicate a greater performance improvement can be gained for longer trajectories or streets, thus validating the benefits of our proposed cuboid space delimitation and models for street-view scenarios.


\section{Downstream applications of the reconstructed surface}
\label{sec:downtream}

\subsection{LiDAR simulation}

\begin{figure}[htbp]
\begin{subfigure}{\textwidth}
\setlength\tabcolsep{1pt}
\centering
\begin{tabular}{c@{\hskip 0.1cm}c}
\includegraphics[width=0.48\textwidth]{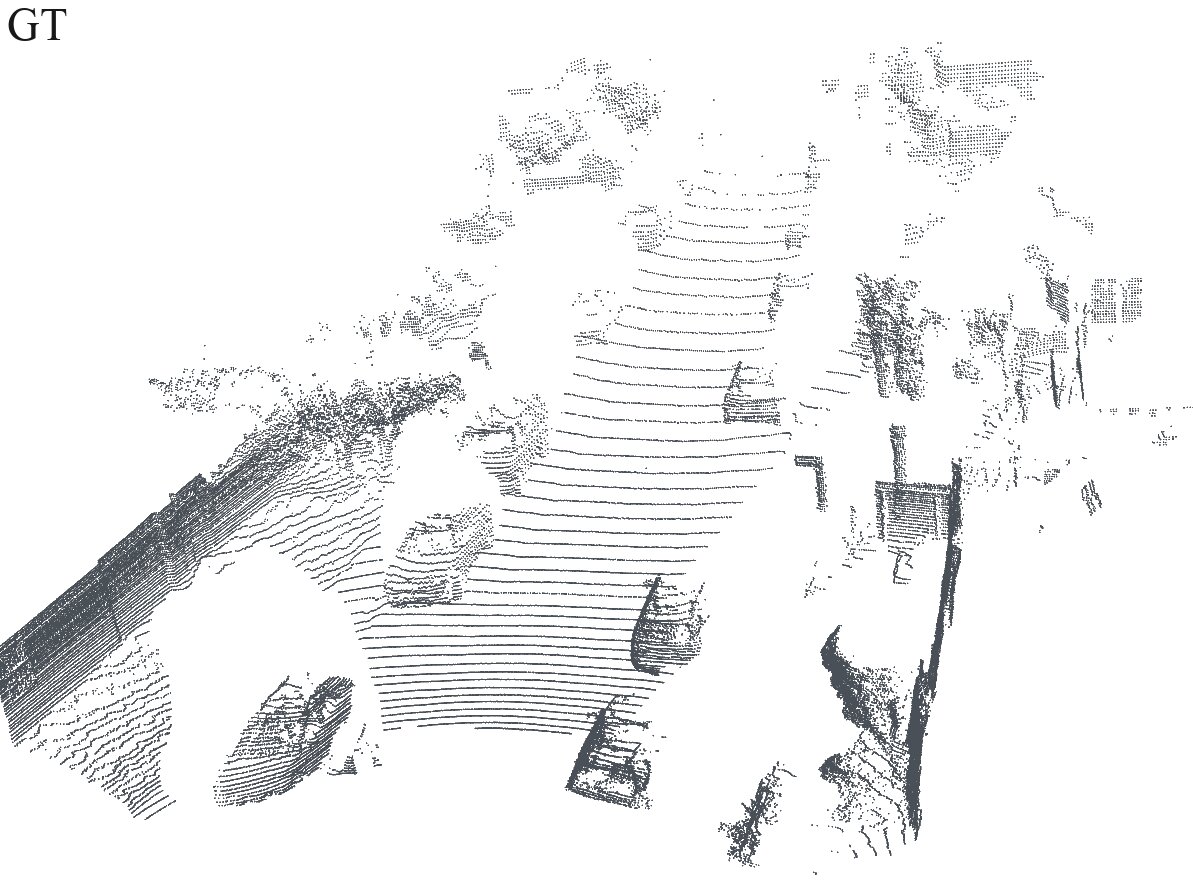}
& \includegraphics[width=0.48\textwidth]{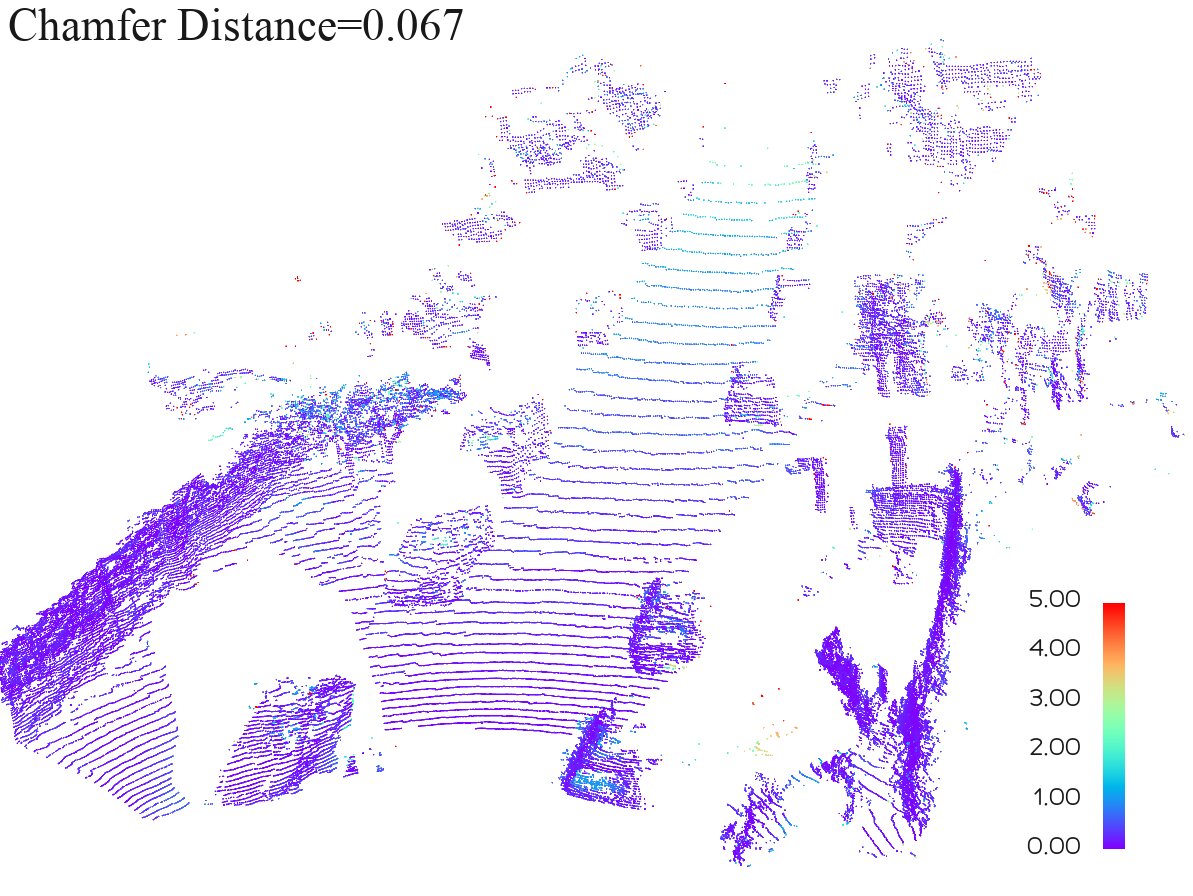}
\end{tabular}
\caption{Qualitative comparison of the ground truth LiDAR pointcloud (left) and the volume-rendered pointcloud (right) of our close-range NeuS model. The colors of the predicted pointcloud (right) depict L1 error of the rendered depth.}
\label{fig:lidar_simulation_gtpred}
\end{subfigure}
\par
\begin{subfigure}{\textwidth}
\centering
\includegraphics[width=0.5\textwidth]{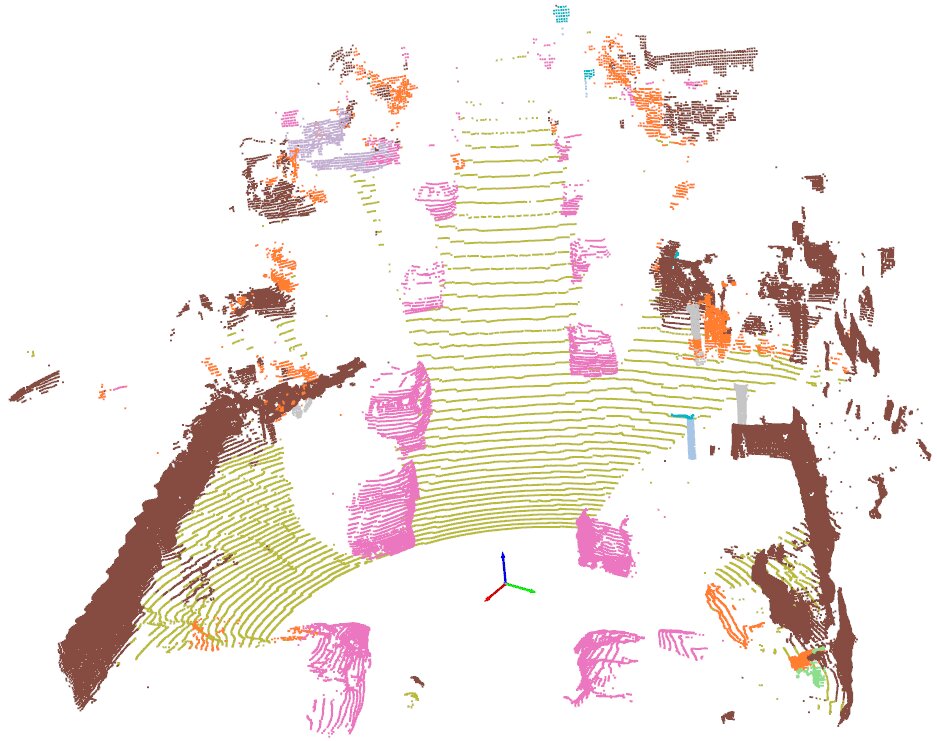}
\caption{Directly feeding our volume rendered pointcloud into a segmentation model based on PCSeg~\citep{pcseg2023}. Different colors indicate different inferred semantics using this coordinate-feature-only pointcloud segmentation model.}
\label{fig:lidar_simulation_seg}
\end{subfigure}
\caption{Qualitative demonstration of the volume-rendered LiDAR pointclouds using our reconstructed geometry of Waymo Open Dataset~\citep{waymo} seg4058410\dots.}
\end{figure}

One of the major fallback of volume rendering is that its depth rendering is often indefinite and inaccurate. 
Even when the training concludes with high average opacity, there will still be a significant interval of non-negligible width that contributes to the rendering. 
This is contrary to common real-world geometry where each ray intersects the surfaces of objects at one definite depth, which can be viewed as an infinitesimal interval.

Luckily under the NeuS~\citep{wang2021neus} setting, the geometry of the scene is explicitly modeled by implicit surfaces, which naturally provides guidance to the volume rendering process. 
In addition, we can also directly apply implicit surface tracing techniques, such as sphere tracing, which guarantees a definite intersection depth.

In this paper, we manually set the mapping parameter $s$ in Supp.~\ref{supp:impl:neus} to a large value, e.g. 64000.
Fig.~\ref{fig:lidar_simulation_gtpred} demonstrates a qualitative example of the simulated pointcloud. 
As shown in Fig.~\ref{fig:lidar_simulation_seg}, in order to further demonstrate the accuracy of our LiDAR simulation, we directly feed the volume-rendered pointclouds into a pointcloud segmentation model obtained using the PCSeg~\citep{pcseg2023} framework.
This pointcloud segmentation model to evaluate our LiDAR simulation is trained on original Waymo Open Dataset~\citep{waymo} and only uses coordinate features as input since we can not produce intensity channel for now.

\subsection{High-resolution mesh and occupancy grid extraction}

\begin{figure}[htbp]
	\centering
	\includegraphics[width=0.8\textwidth]{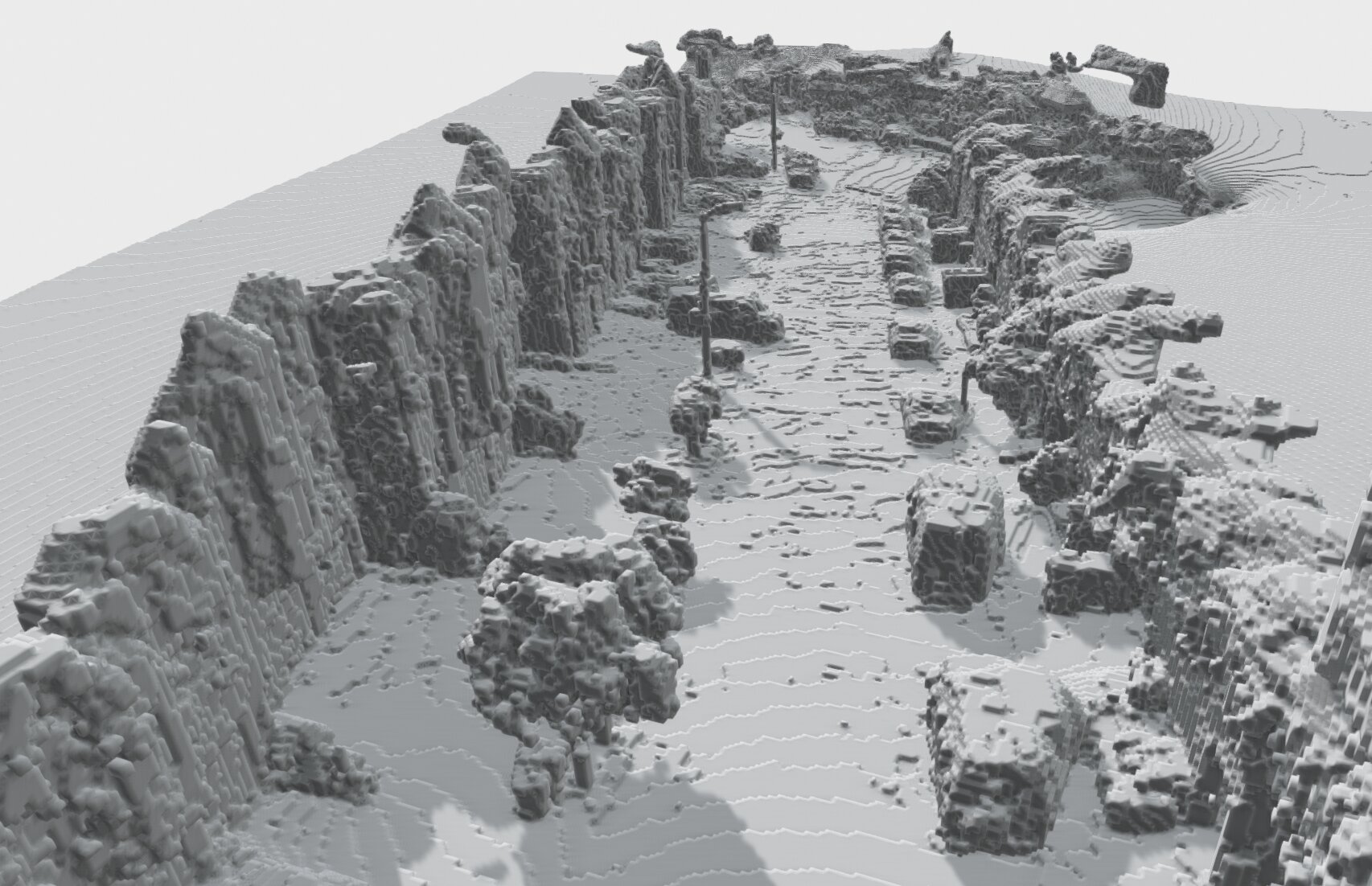}
	\caption{Qualitative example of the extracted occupancy grid @0.1(m) resolution on Waymo Open Dataset~\citep{waymo} seg1006130\dots. The geometry is reconstructed without using LiDAR data.}
	\label{fig:extract_occgrid}
\end{figure}

With the implicit surface being reconstructed, we obtain an continuous representation of scene geometry that has infinitesimal granularity. 
Subsequently, we can extract high-resolution meshes or occupancy grids out of the reconstructed implicit surface, as shown in Fig.~\ref{fig:overview},~\ref{fig:extract_occgrid}.

\subsection{Implicit-surface-guided ray tracing}

\begin{figure}[htbp]
\centering
\begin{tabular}{c@{\hskip 0.1cm}c@{\hskip 0.1cm}c@{\hskip 0.1cm}c}
\includegraphics[width=0.24\textwidth]{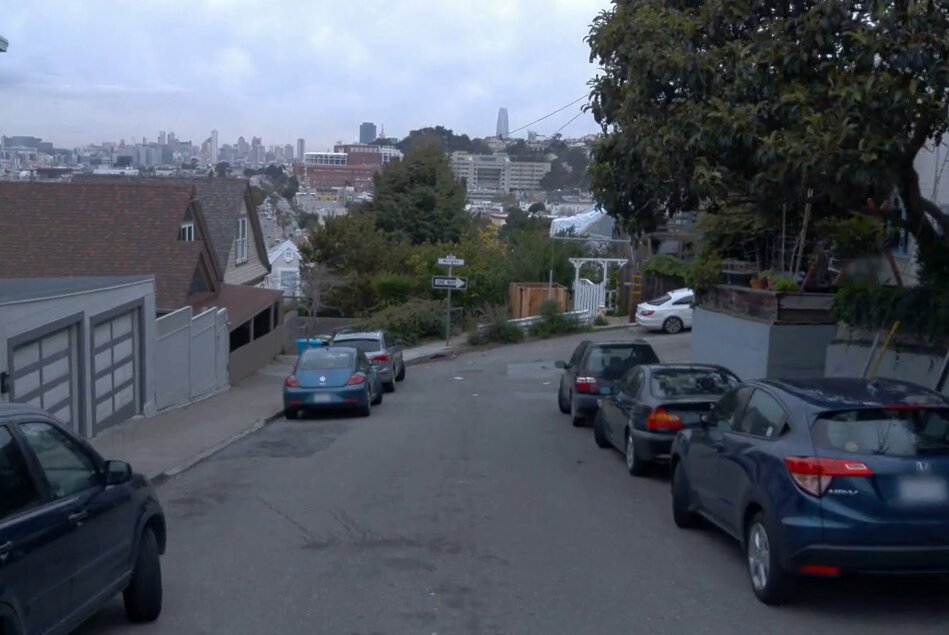}
& \includegraphics[width=0.24\textwidth]{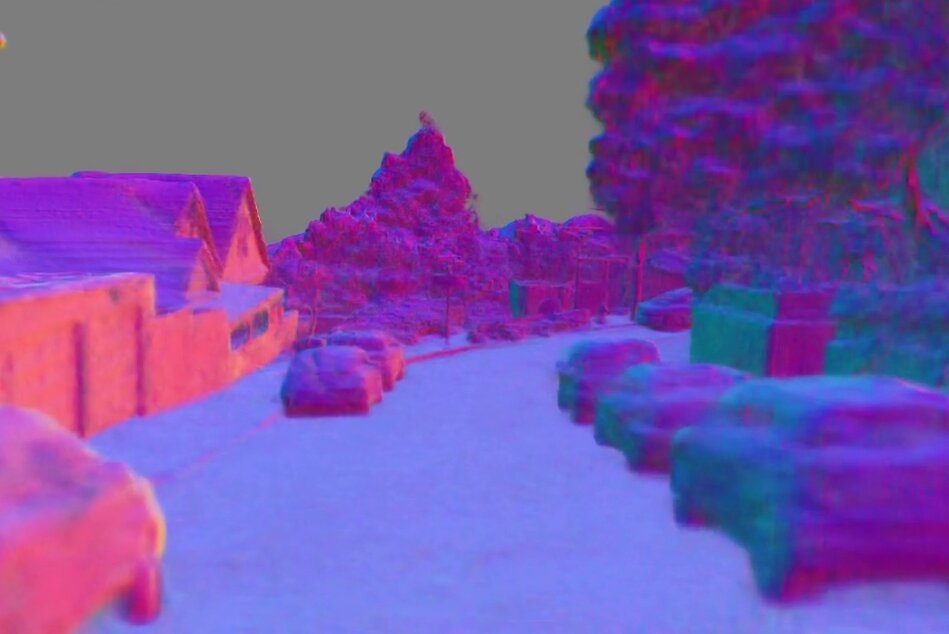}
& \includegraphics[width=0.24\textwidth]{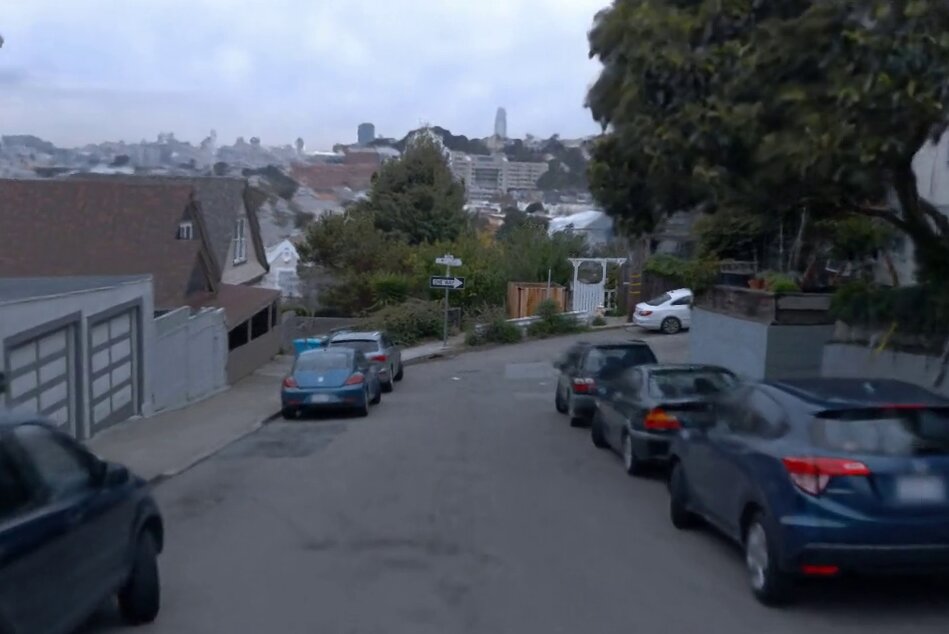}
& \includegraphics[width=0.24\textwidth]{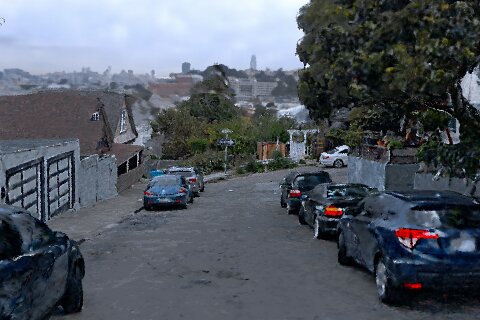} \\
\small{GT image} & \small{Volume-rendered normals} & \small{Volume-rendered image} & \small{Sphere-traced image} \\

\multicolumn{4}{c}{
\begin{subfigure}{\textwidth}
\caption{Example comparison of volume-rendering vs. sphere tracing.}
\label{fig:raytrace:comp}
\end{subfigure}
} \\

& & & \\

\includegraphics[width=0.24\textwidth]{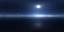}
& \includegraphics[width=0.24\textwidth]{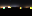}
& \includegraphics[width=0.24\textwidth]{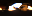}
& \includegraphics[width=0.24\textwidth]{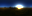} \\

\includegraphics[width=0.24\textwidth]{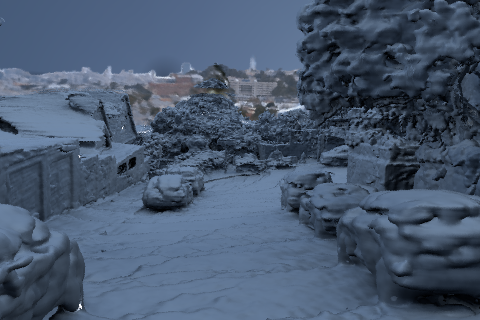}
& \includegraphics[width=0.24\textwidth]{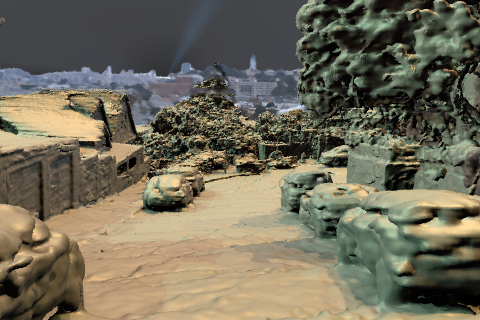}
& \includegraphics[width=0.24\textwidth]{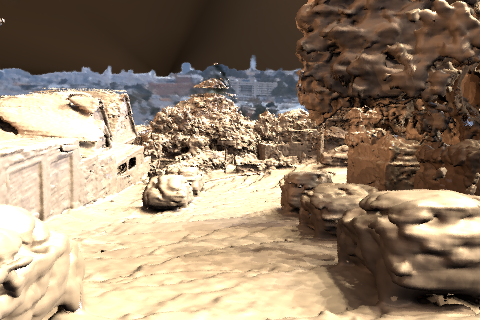}
& \includegraphics[width=0.24\textwidth]{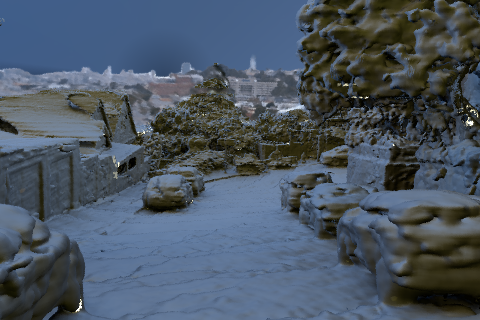} \\

\small{Puresky} & \small{Night} & \small{Courtyard} & \small{Sunrise} \\

\multicolumn{4}{c}{
\begin{subfigure}{\textwidth}
\caption{Shading the implicit surface (bottom) under different envmaps (top) based on sphere-tracing.}
\label{fig:raytrace:shading}
\end{subfigure}
} \\
\end{tabular}
\caption{Preliminary examples of sphere-tracing and shading the implicit surface based on sphere-tracing. Experiments are conducted on Waymo Open Dataset~\citep{waymo} seg1347637\dots. 
}
\end{figure}

Apart from indefinite ray intersection depth, another major fallback of volume rendering is that it can not obtain definite surface geometry and surface normals for shading, re-lighting, change of environment map, or other graphics applications requiring further factorization of the radiance representation.
Using our approach, however, a definite implicit surface with high-fidelity surface normals can be obtained for street views even without requiring 3D sensors like LiDAR, as shown in Fig.~\ref{fig:head},~\ref{fig:demo_withoutlidar}. 
Subsequently, we can conduct shading or re-lighting on the reconstructed implicit surface. 
We present a preliminary example of the rendered image using sphere-tracing in Fig.~\ref{fig:raytrace:comp}, as well as applying different environment maps to shade the implicit surfaces in Fig.~\ref{fig:raytrace:shading}.

\section{Conclusion, limitations and future work}

In this paper, we present StreetSurf, a novel multi-view implicit surface reconstruction method designed for street views. We mainly tackle the problem of close-range-vs-distant-view disentanglement and geometric errors caused by non-object-centric views and unbounded scenes.

However, scenes in autonomous driving datasets are still much challenging and StreetSurf only handles a small proportion of scenarios in these datasets.  
Many other challenges are remained, including dark or dawn illuminations, numerous dynamic objects, long-tail environmental conditions and so on.
In addition, StreetSurf ignores dynamic foregrounds in street views, but they are also important to reconstruct and render for various downstream tasks.
In our future work, we will study the task of multi-object rendering and reconstruction, with StreetSurf as the background model.

\begin{ack}
The research was supported by the Science and Technology Commission of Shanghai Municipality (grant No. 22DZ1100102).
\end{ack}

{\small
	\bibliographystyle{abbrvnat}
	\bibliography{egbib}
}

%



\newcommand{\customtitle}[1]{
	\begingroup 
	\centering
	\Large\bfseries
	#1\\ 
	\bigskip 
	\endgroup 
}

\clearpage
\customtitle{Supplemental Material for StreetSurf: Extending Multi-view Implicit Surface Reconstruction to Street Views}
\setcounter{section}{0} 
\renewcommand{\thesection}{\AlphAlph{\value{section}}}

In this \textbf{supplementary document}, we first demonstrate more results on street-view reconstruction in Sec.~\ref{supp:additional_results}.
Then, we go through more implementation details over the whole optimization process in Sec.~\ref{supp:implementation_details}.

\section{Additional results}
\label{supp:additional_results}

For all Waymo Open Dataset~\citep{waymo} experiments, we use images captured from three frontal cameras for fair comparison.
Some of sequences are trimmed due to limited ego-motion or a high density of dynamic objects. The details of selection are shown in Tab.~\ref{tab:supp:waymo}.

\begin{table}[!htbp]
\caption{Information on the selected and trimmed Waymo Open Dataset~\citep{waymo} sequences. "--" indicates using the original start or end frames (typically 198 frames) without trimming.}
\label{tab:supp:waymo}
\begin{minipage}{0.48\textwidth}
\centering
\small
\begin{tabular}{c||c|c|r}
	
\toprule[1pt]
\multirow{2}{*}{Sequence} & Start & End & \multicolumn{1}{c}{Trajectory} \\
& frame & frame & \multicolumn{1}{c}{length (m)} \\

\midrule[0.5pt]
	
seg1006130\dots & -- & 163 & 100.6 \\
seg1027514\dots & -- & -- & 153.9 \\
seg1067626\dots & -- & -- & 210.6 \\
seg1137922\dots & -- & -- & 167.6 \\
seg1172406\dots & -- & -- & 141.4 \\
seg1287964\dots & -- & -- & 71.0 \\
seg1308545\dots & -- & -- & 106.1 \\
seg1314219\dots & 17 & -- & 121.1 \\
seg1319679\dots & -- & -- & 73.0 \\
seg1323841\dots & -- & -- & 186.6 \\
seg1347637\dots & -- & 140 & 49.4 \\
seg1400454\dots & 24 & -- & 129.8 \\
seg1434813\dots & -- & -- & 69.6 \\
seg1442480\dots & -- & -- & 157.6 \\
seg1486973\dots & -- & -- & 119.3 \\
seg1506235\dots & -- & -- & 197.2 \\
\bottomrule[1pt]
\end{tabular}
\end{minipage}
\hfill
\begin{minipage}{0.48\textwidth}
\centering
\small
\begin{tabular}{c||c|c|r}
\toprule[1pt]
\multirow{2}{*}{Sequence} & Start & End & \multicolumn{1}{c}{Trajectory} \\
& frame & frame & \multicolumn{1}{c}{length (m)} \\
\midrule[0.5pt]
seg1522170\dots & -- & -- & 72.8 \\
seg1527063\dots & 30 & -- & 113.1 \\
seg1534950\dots & 80 & -- & 70.5 \\
seg1536582\dots & -- & 170 & 111.9 \\
seg1586862\dots & 70 & -- & 83.0 \\
seg1634531\dots & -- & -- & 160.1 \\
seg1647019\dots & -- & -- & 125.0 \\
seg1660852\dots & -- & 120 & 66.2 \\
seg1664636\dots & -- & -- & 107.8 \\
seg1776195\dots & -- & -- & 153.5 \\
seg3224923\dots & -- & -- & 151.4 \\
seg3425716\dots & -- & -- & 69.7 \\
seg3988957\dots & -- & -- & 137.2 \\
seg4058410\dots & -- & 90 & 55.4 \\
seg8811210\dots & -- & -- & 137.3 \\
seg9385013\dots & -- & -- & 263.9 \\
\bottomrule[1pt]
\end{tabular}
\end{minipage}
\end{table}

\subsection{Street-view reconstruction without LiDAR}

\begin{figure}[htbp]
\centering
\setlength\tabcolsep{1pt}
\begin{tabular}{c@{\hskip 0.03cm}c@{\hskip 0.1cm}c}
\raisebox{0.09\textwidth}[0pt][0pt]{\rotatebox[origin=c]{90}{GT}}
& \multicolumn{2}{c}{\includegraphics[width=0.9\textwidth]{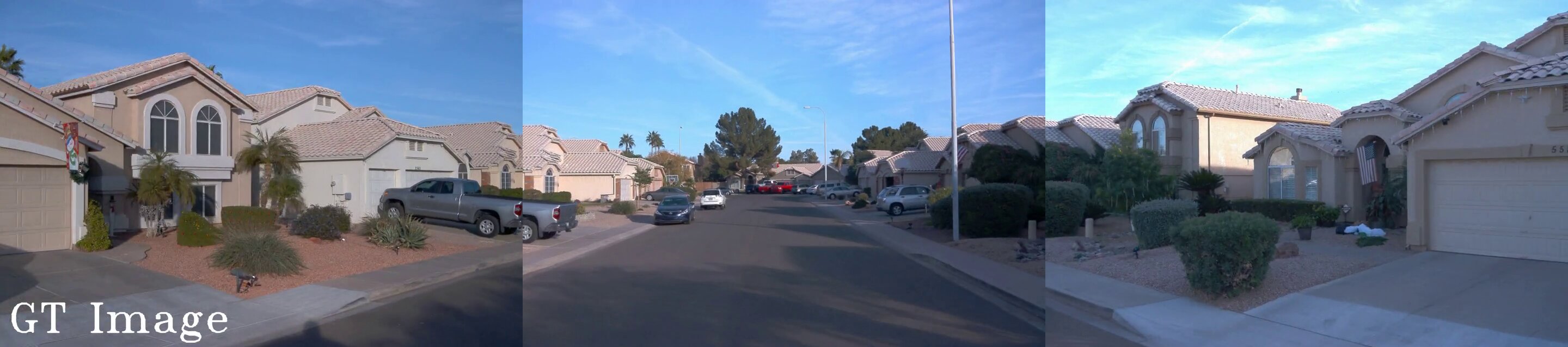}} \\[-2pt]

\raisebox{0.09\textwidth}[0pt][0pt]{\rotatebox[origin=c]{90}{Ours}} 
& \multicolumn{2}{c}{\includegraphics[width=0.9\textwidth]{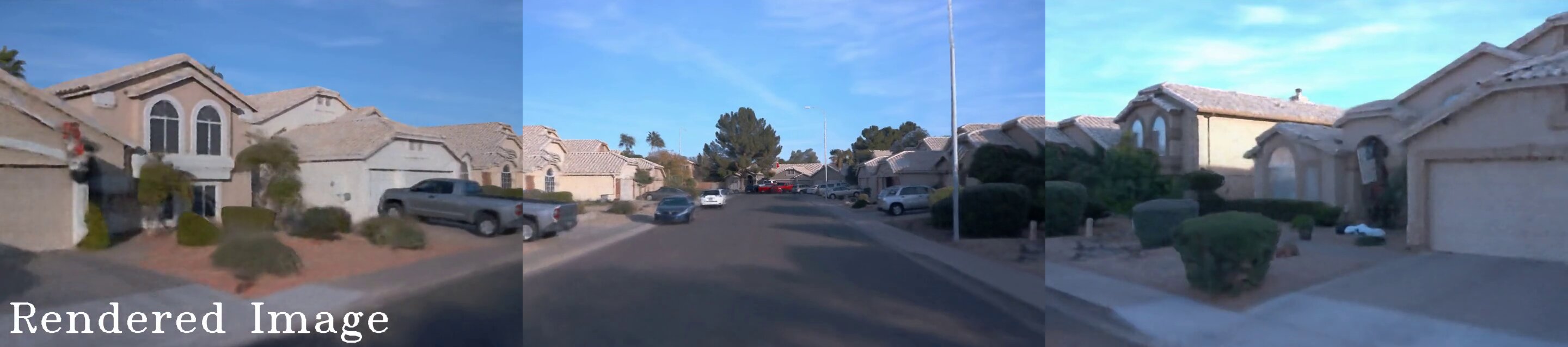}} \\[-2pt]

\raisebox{0.09\textwidth}[0pt][0pt]{\rotatebox[origin=c]{90}{$\text{F}^2$-NeRF~\citep{wang2023f2nerf}}} 
& \multicolumn{2}{c}{\includegraphics[width=0.9\textwidth]{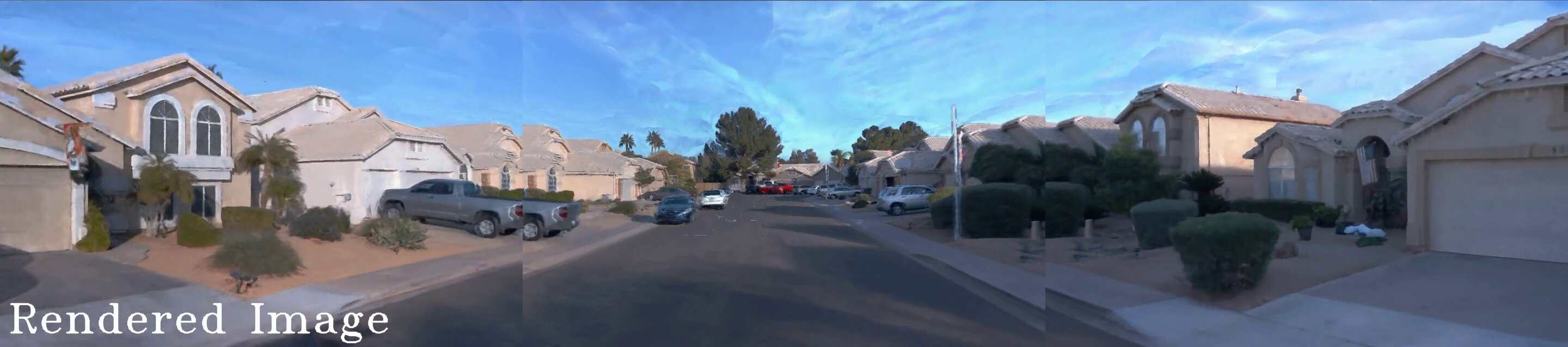}} \\[-2pt]

\raisebox{0.09\textwidth}[0pt][0pt]{\rotatebox[origin=c]{90}{Ours}} 
& \multicolumn{2}{c}{\includegraphics[width=0.9\textwidth]{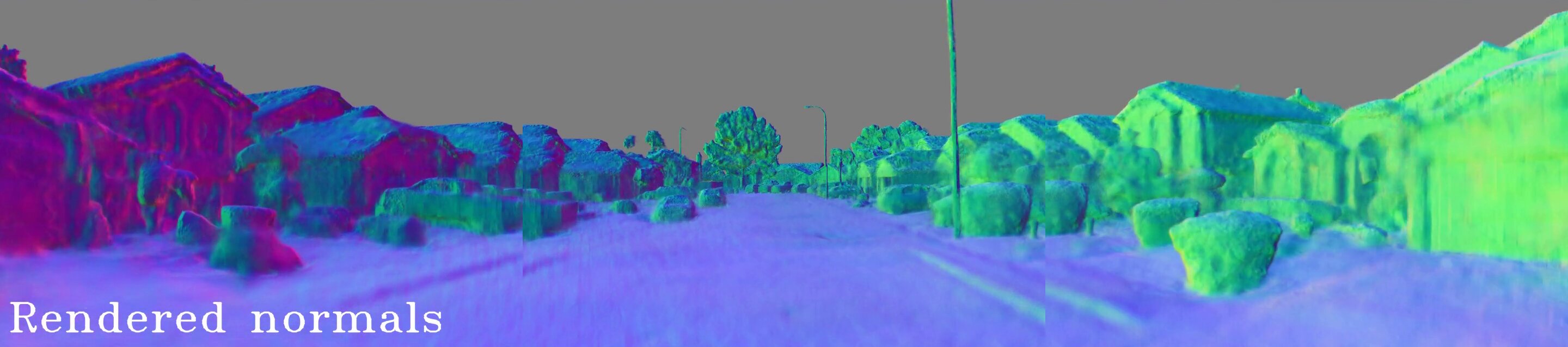}} \\[-2pt]

\raisebox{0.09\textwidth}[0pt][0pt]{\rotatebox[origin=c]{90}{Ours}} 
& \multicolumn{2}{c}{\includegraphics[width=0.9\textwidth]{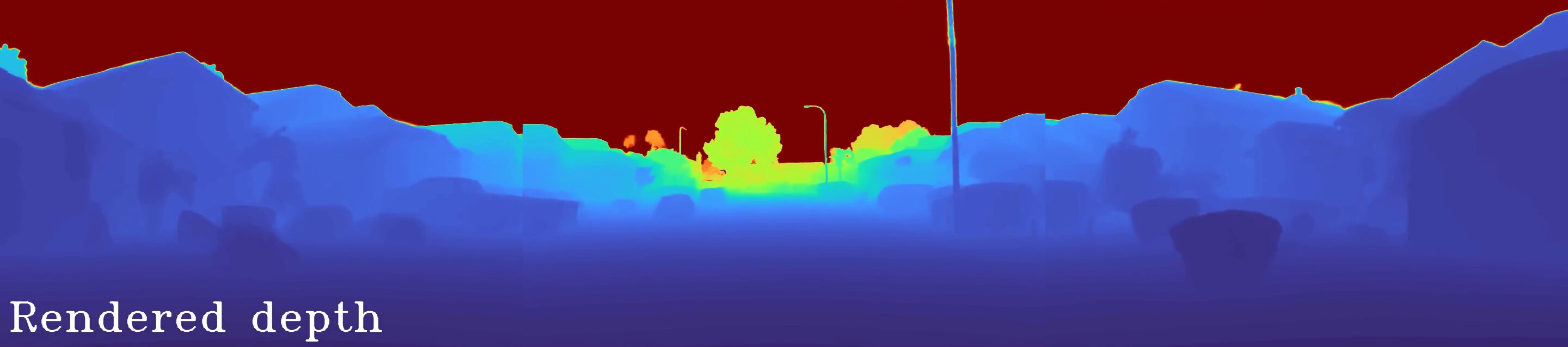}} \\[-2pt]

\raisebox{0.09\textwidth}[0pt][0pt]{\rotatebox[origin=c]{90}{$\text{F}^2$-NeRF~\citep{wang2023f2nerf}}} 
& \multicolumn{2}{c}{\includegraphics[width=0.9\textwidth]{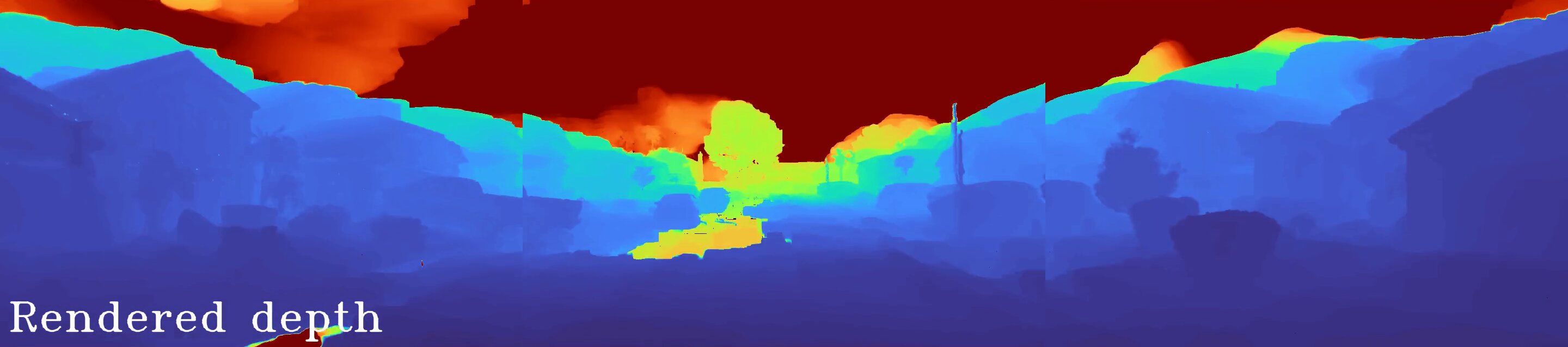}} \\

\raisebox{0.15\textwidth}[0pt][0pt]{\rotatebox[origin=c]{90}{Reconstructed surfaces}} 
& \includegraphics[width=0.45\textwidth]{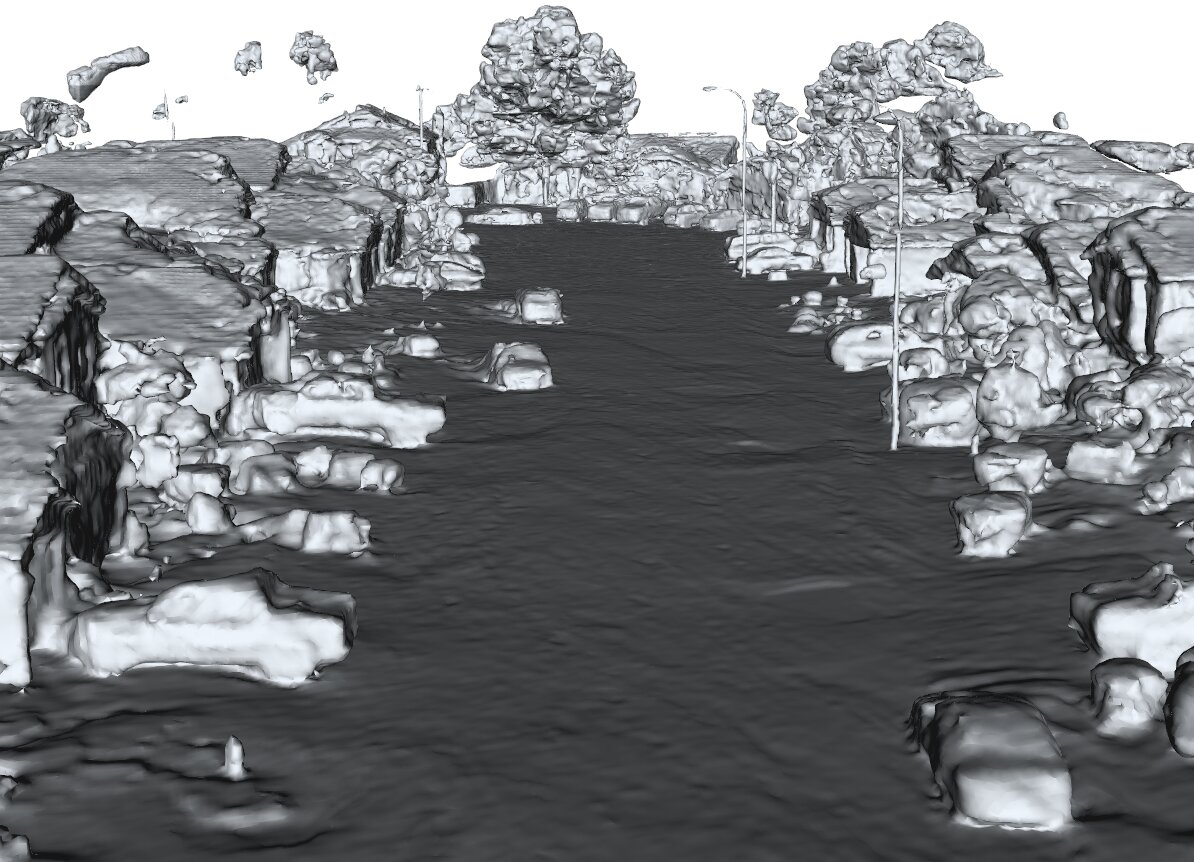}
& \includegraphics[width=0.45\textwidth]{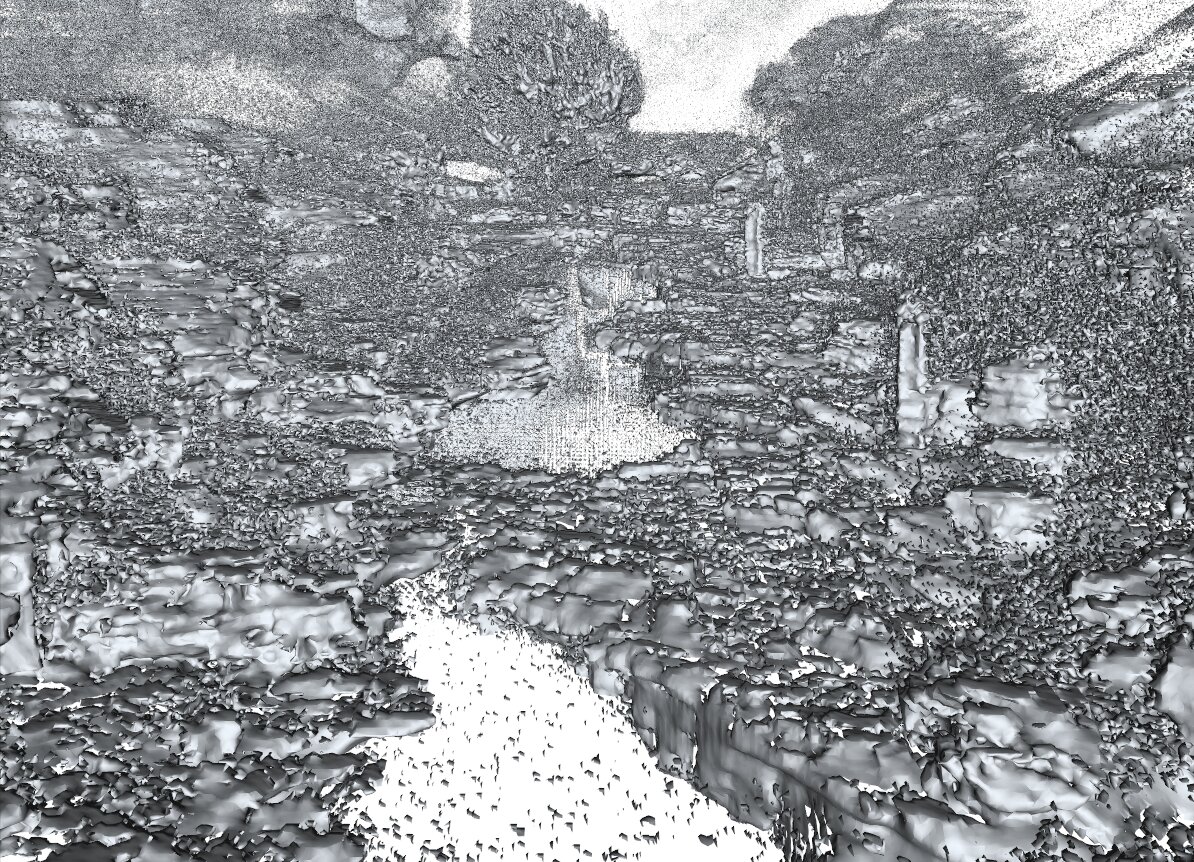} 
\\

& Ours & $\text{F}^2$-NeRF~\citep{wang2023f2nerf}
\end{tabular}
\caption{Qualitative comparison of reconstruction \textbf{without LiDAR}, on Waymo Open Dataset~\citep{waymo} seg1027514\dots.}
\label{fig:supp:demo_withoutlidar2}
\end{figure}

\begin{figure}[htbp]
\centering
\setlength\tabcolsep{1pt}
\begin{tabular}{c@{\hskip 0.1cm}c@{\hskip 0.1cm}c}
\raisebox{0.09\textwidth}[0pt][0pt]{\rotatebox[origin=c]{90}{GT}}
& \multicolumn{2}{c}{\includegraphics[width=0.9\textwidth]{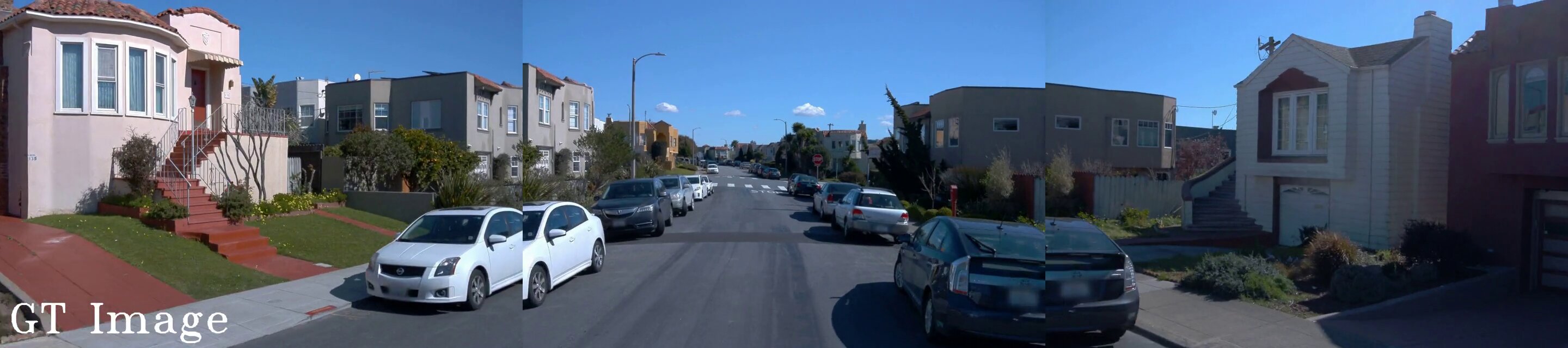}} \\[-2pt]

\raisebox{0.09\textwidth}[0pt][0pt]{\rotatebox[origin=c]{90}{Ours}} 
& \multicolumn{2}{c}{\includegraphics[width=0.9\textwidth]{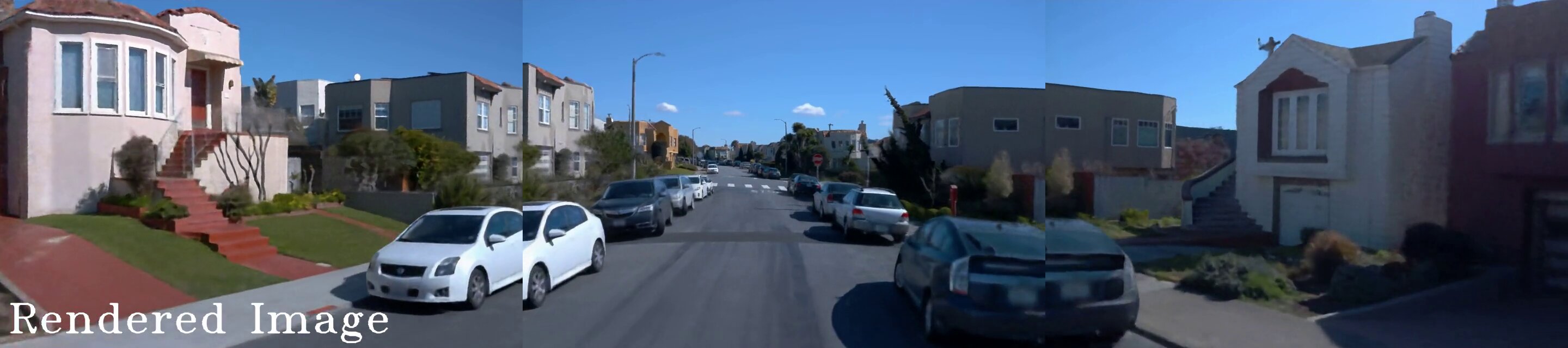}} \\[-2pt]

\raisebox{0.09\textwidth}[0pt][0pt]{\rotatebox[origin=c]{90}{$\text{F}^2$-NeRF~\citep{wang2023f2nerf}}} 
& \multicolumn{2}{c}{\includegraphics[width=0.9\textwidth]{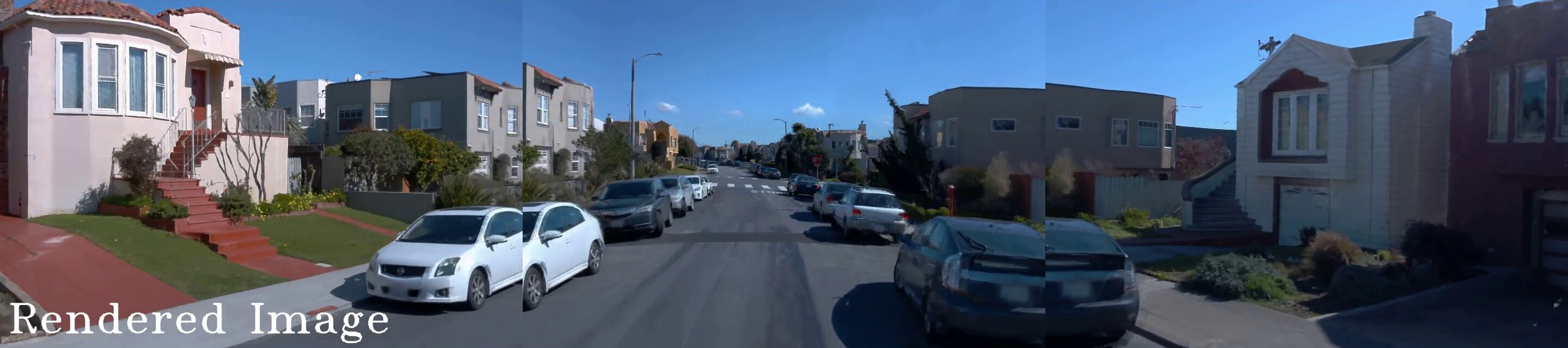}} \\[-2pt]

\raisebox{0.09\textwidth}[0pt][0pt]{\rotatebox[origin=c]{90}{Ours}} 
& \multicolumn{2}{c}{\includegraphics[width=0.9\textwidth]{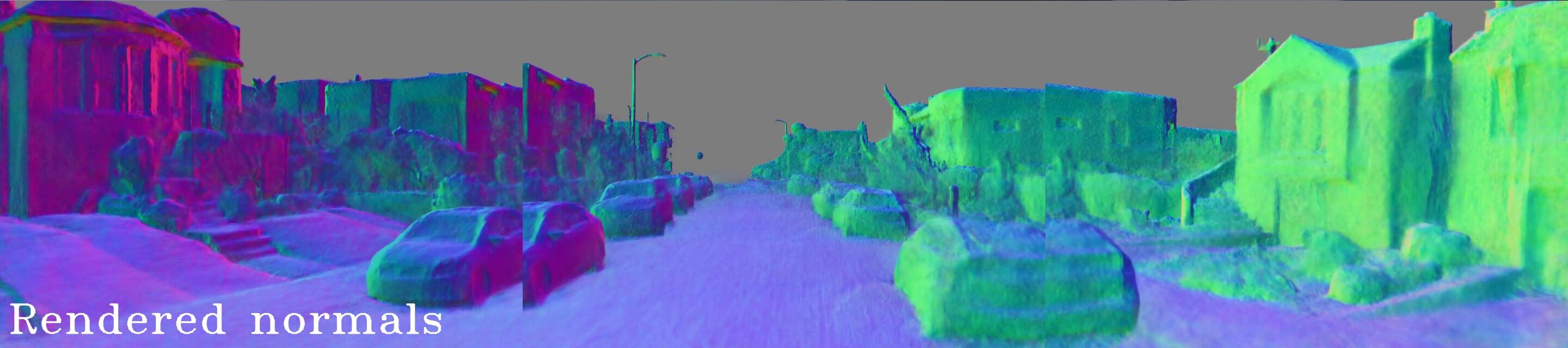}} \\[-2pt]

\raisebox{0.09\textwidth}[0pt][0pt]{\rotatebox[origin=c]{90}{Ours}} 
& \multicolumn{2}{c}{\includegraphics[width=0.9\textwidth]{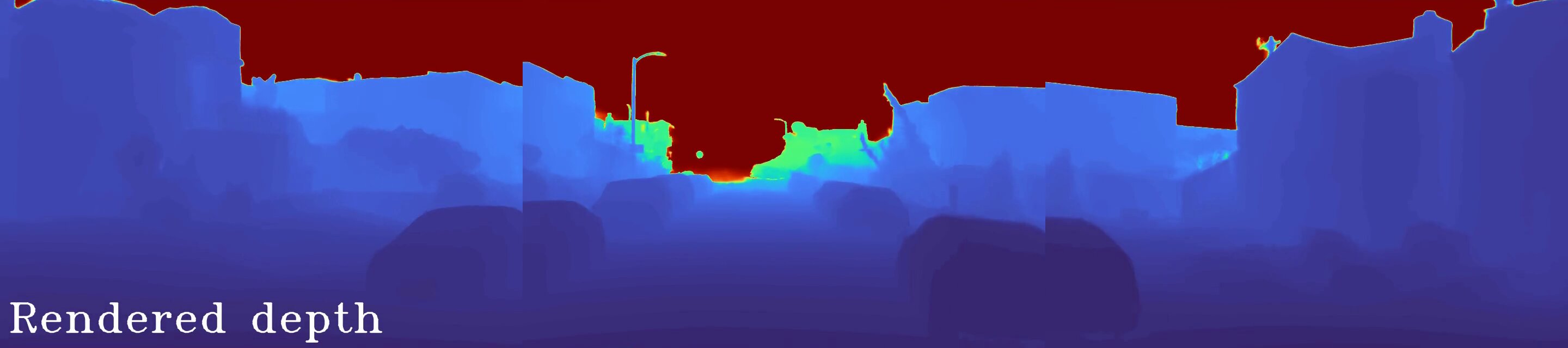}} \\[-2pt]

\raisebox{0.09\textwidth}[0pt][0pt]{\rotatebox[origin=c]{90}{$\text{F}^2$-NeRF~\citep{wang2023f2nerf}}} 
& \multicolumn{2}{c}{\includegraphics[width=0.9\textwidth]{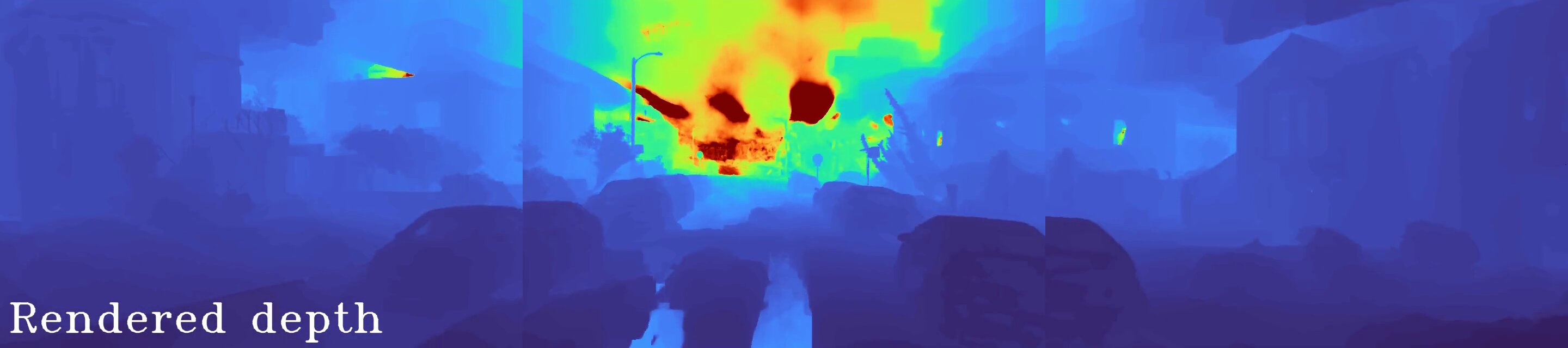}} \\

\raisebox{0.15\textwidth}[0pt][0pt]{\rotatebox[origin=c]{90}{Reconstructed surfaces}} 
& \includegraphics[width=0.45\textwidth]{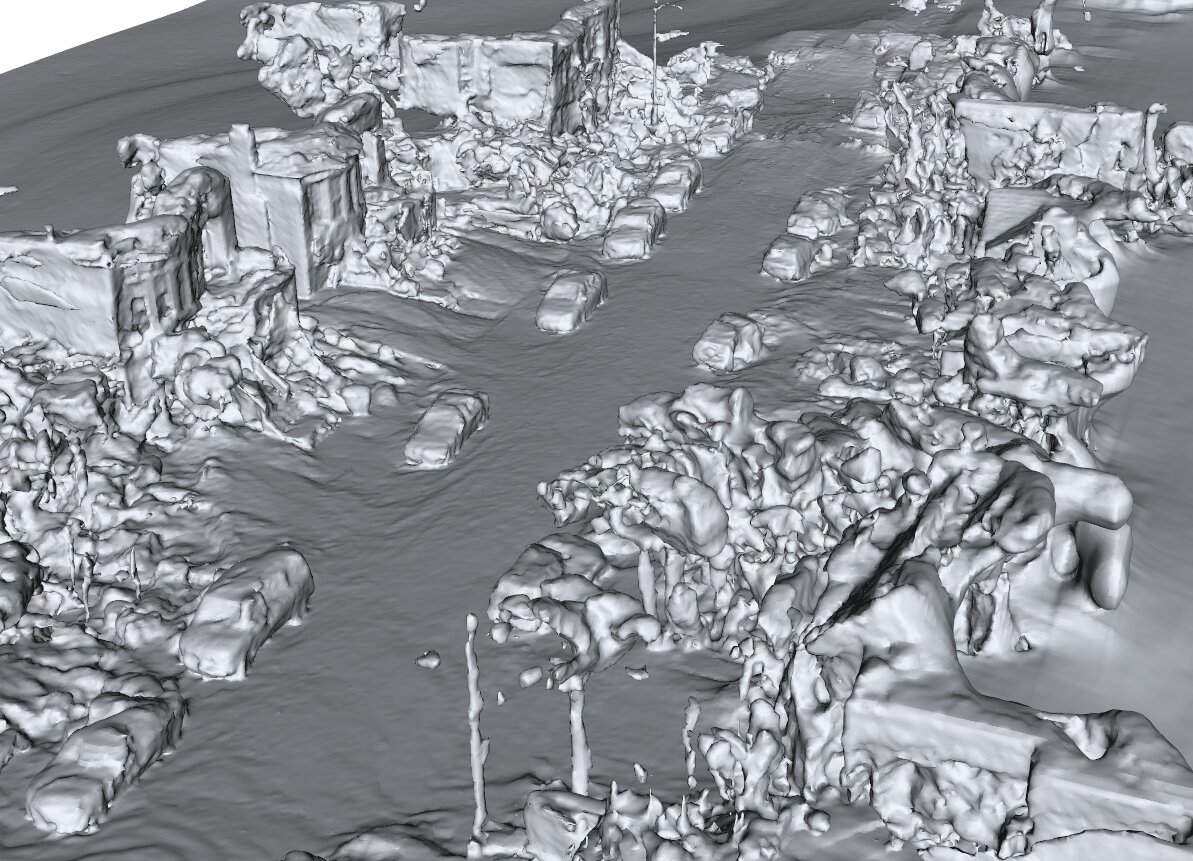}
& \includegraphics[width=0.45\textwidth]{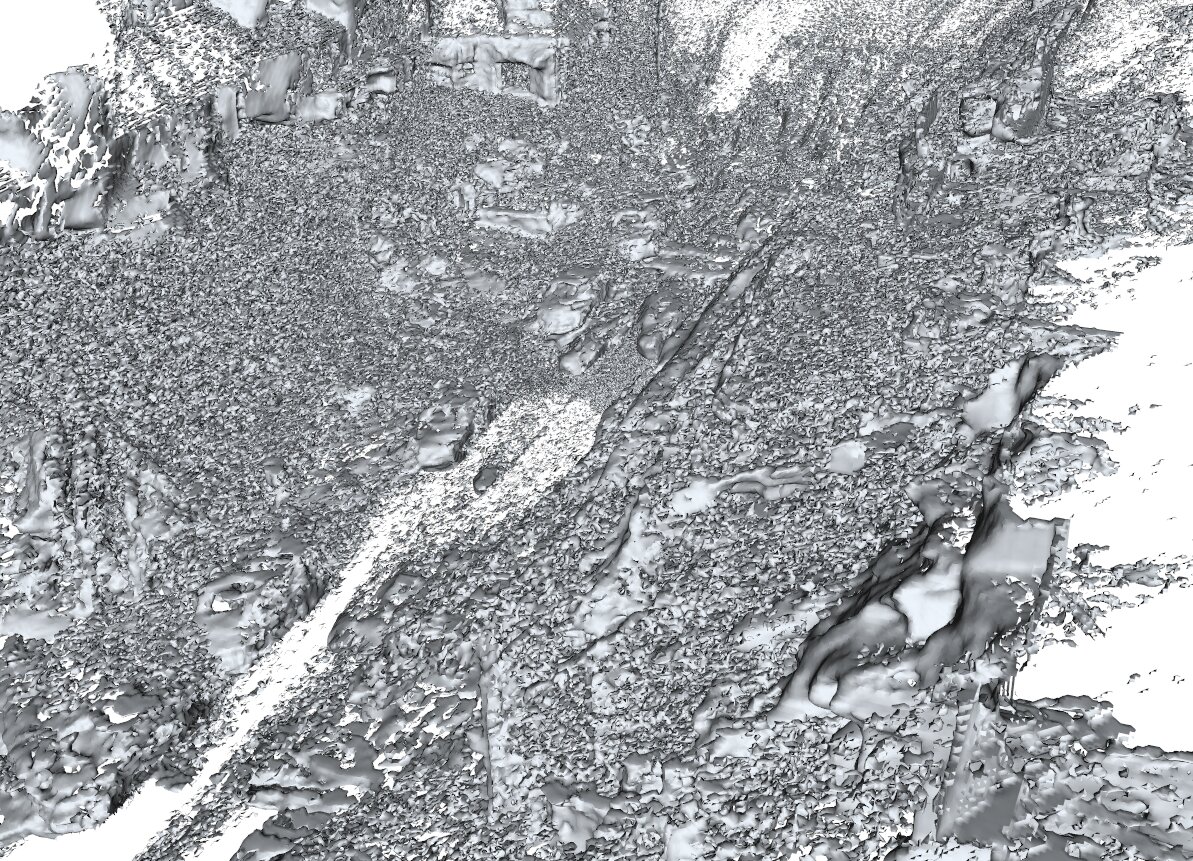} 
\\

& Ours & $\text{F}^2$-NeRF~\citep{wang2023f2nerf}
\end{tabular}
\caption{Qualitative comparison of reconstruction \textbf{without LiDAR}, on Waymo Open Dataset~\citep{waymo} seg4058410\dots.}
\label{fig:supp:demo_withoutlidar3}
\end{figure}

For reconstruction without LiDAR data, we compare our approach with two baselines.
The first is the traditional multi-view reconstruction pipeline COLMAP~\citep{schoenberger2016sfm}. The second is the state of the art novel view synthesis framework aiming at non-object-centric camera trajectories, $\text{F}^2$-NeRF~\citep{wang2022neus2}. Example qualitative comparisons are shown in Fig.~\ref{fig:supp:demo_withoutlidar2},~\ref{fig:supp:demo_withoutlidar3}.
Quantitative comparison is reported in Tab.~\ref{tab:supp:recon_withoutlidar}.

%
%

\begin{figure}[htbp]
	\setlength\tabcolsep{1pt}
	\centering
	\begin{tabular}{c@{\hskip 0.1cm}c}
		
		\raisebox{0.09\textwidth}[0pt][0pt]{\rotatebox[origin=c]{90}{GT}}
		& \includegraphics[width=0.9\textwidth]{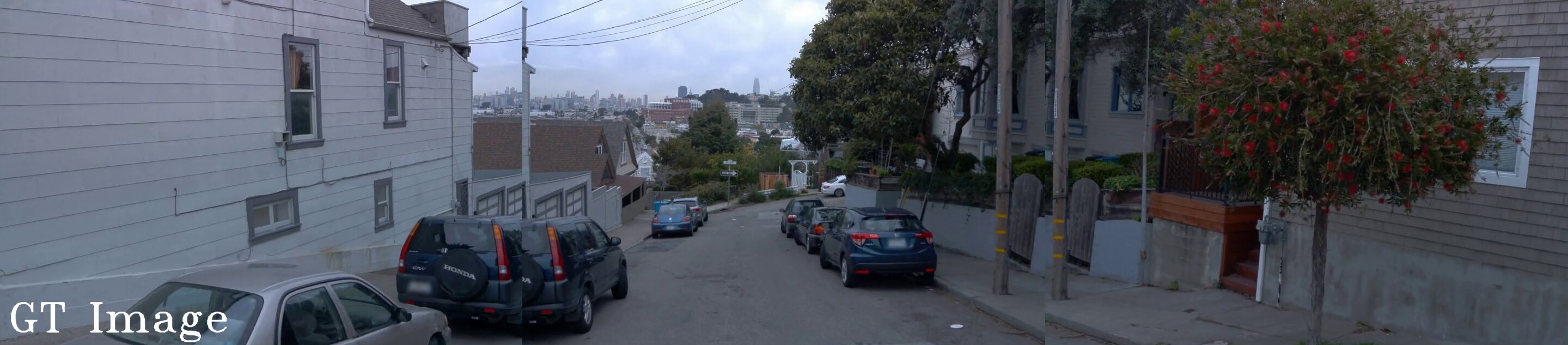} \\[-2pt]
		
		\raisebox{0.09\textwidth}[0pt][0pt]{\rotatebox[origin=c]{90}{Ours}}
		& \includegraphics[width=0.9\textwidth]{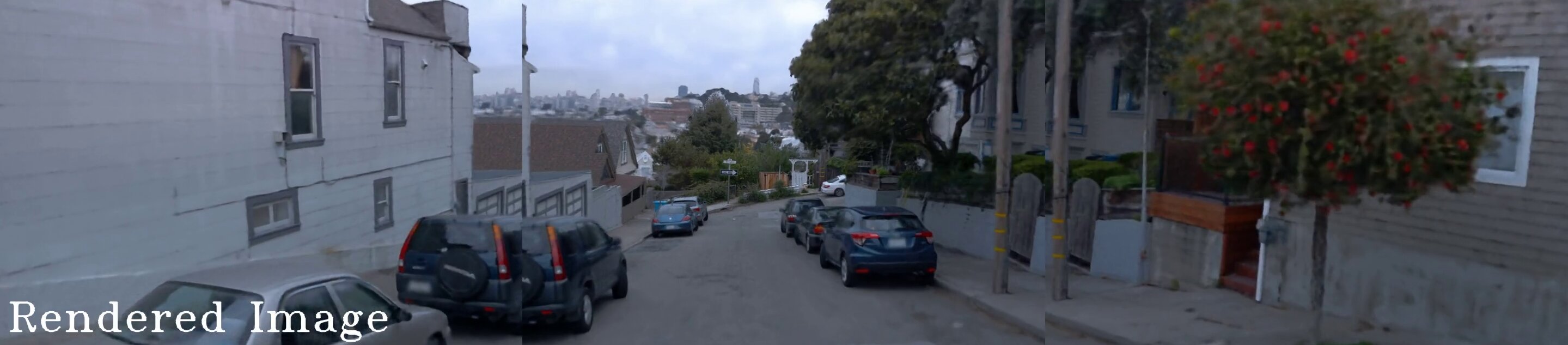} \\[-2pt]
		
		\raisebox{0.09\textwidth}[0pt][0pt]{\rotatebox[origin=c]{90}{NGP~\citep{muller2022instantngp}+L~\citep{rematas2022urban}}}
		& \includegraphics[width=0.9\textwidth]{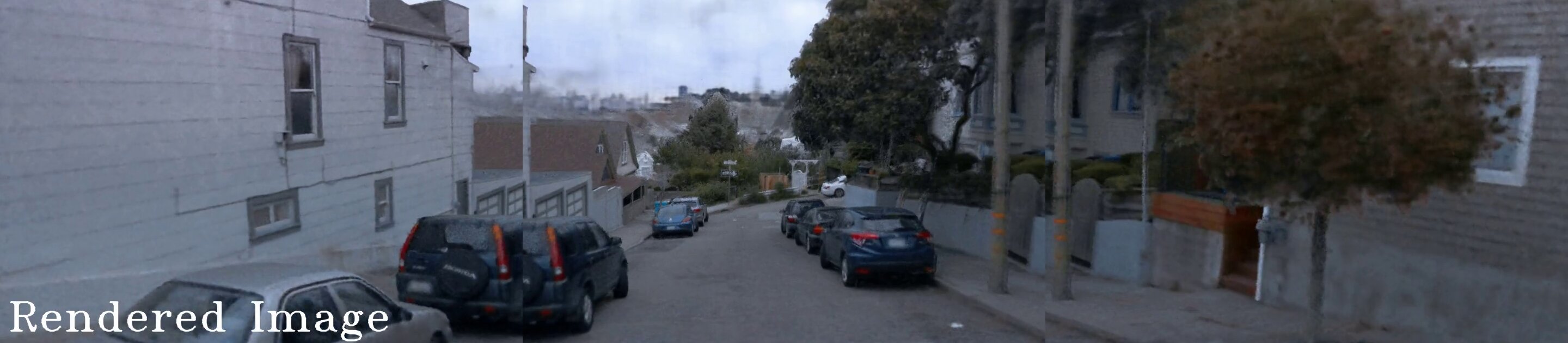} \\[-2pt]
		
		\raisebox{0.09\textwidth}[0pt][0pt]{\rotatebox[origin=c]{90}{Ours}}
		& \includegraphics[width=0.9\textwidth]{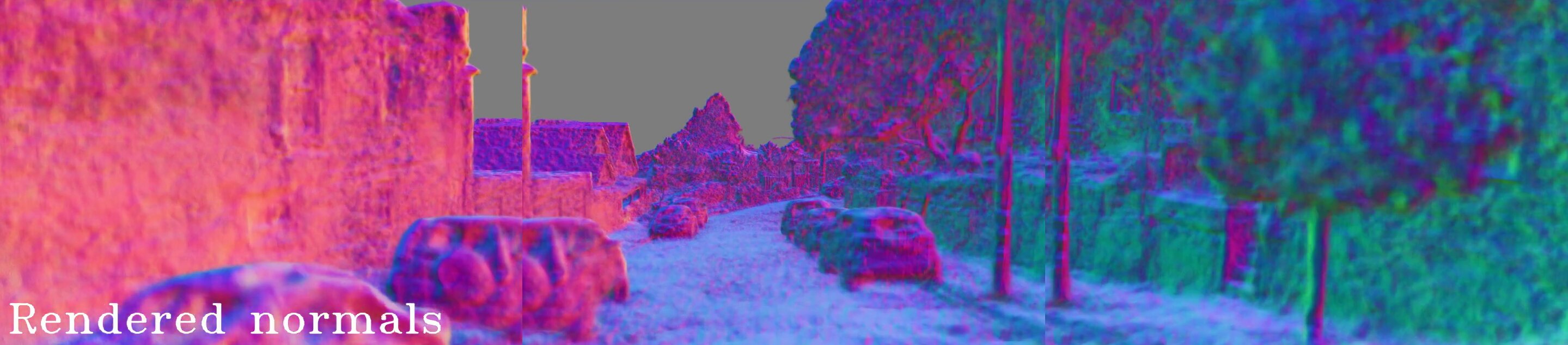} \\[-2pt]

		\raisebox{0.09\textwidth}[0pt][0pt]{\rotatebox[origin=c]{90}{Ours + Normal cues}}
		& \includegraphics[width=0.9\textwidth]{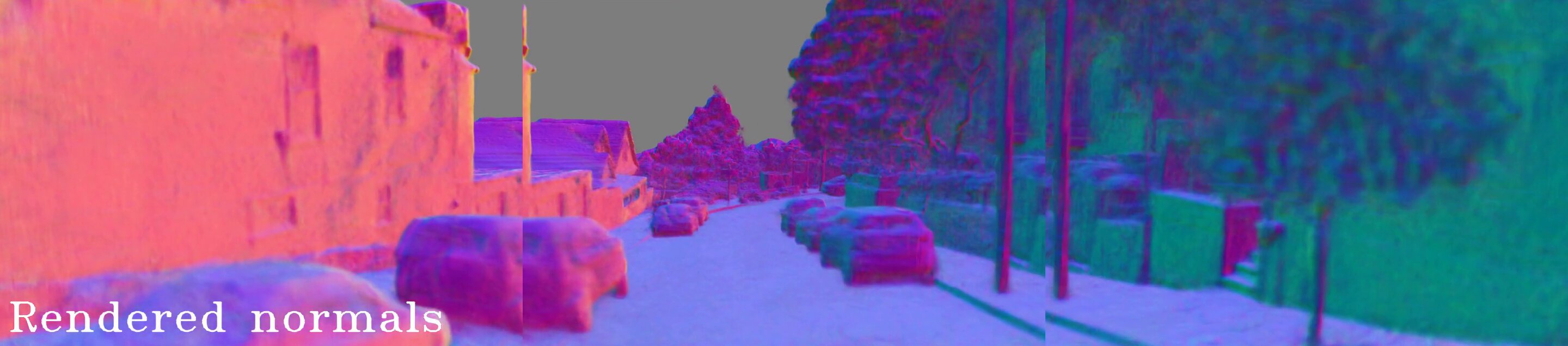} \\[-2pt]

		\raisebox{0.09\textwidth}[0pt][0pt]{\rotatebox[origin=c]{90}{NeuS~\citep{wang2021neus}}}
		& \includegraphics[width=0.9\textwidth]{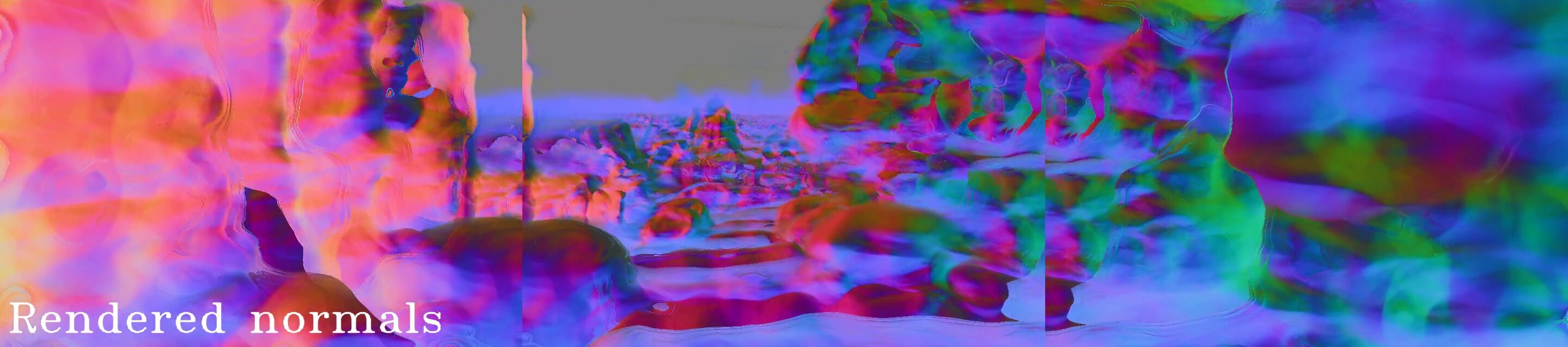} \\[-2pt]

		\raisebox{0.09\textwidth}[0pt][0pt]{\rotatebox[origin=c]{90}{Ours}}
		& \includegraphics[width=0.9\textwidth]{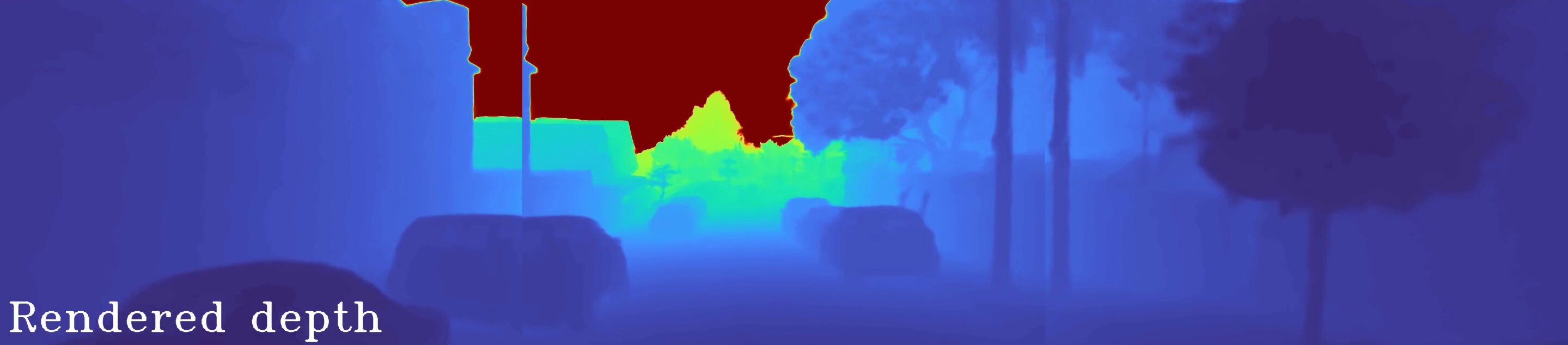} \\[-2pt]

		\raisebox{0.09\textwidth}[0pt][0pt]{\rotatebox[origin=c]{90}{NGP~\citep{muller2022instantngp}+L~\citep{rematas2022urban}}}
		& \includegraphics[width=0.9\textwidth]{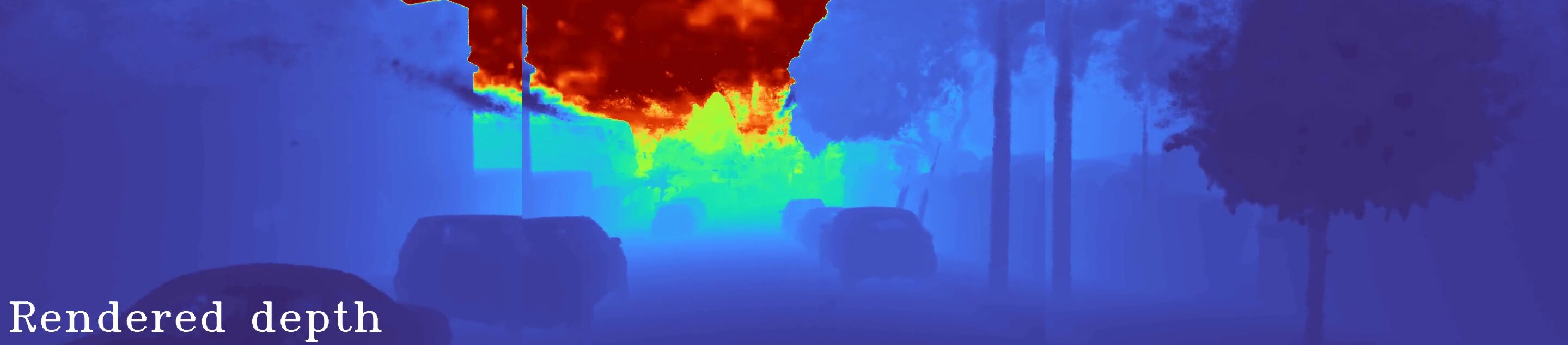}
		
	\end{tabular}
	\caption{Qualitative comparison of reconstruction with LiDAR, using all five LiDARs, on Waymo Open Dataset~\citep{waymo} seg1347637\dots.}
	\label{fig:supp:demo_withlidar_all2}
\end{figure}

\begin{figure}[htbp]
	\setlength\tabcolsep{1pt}
	\centering
	\begin{tabular}{c@{\hskip 0.1cm}c}
		\raisebox{0.09\textwidth}[0pt][0pt]{\rotatebox[origin=c]{90}{GT}}
		& \includegraphics[width=0.9\textwidth]{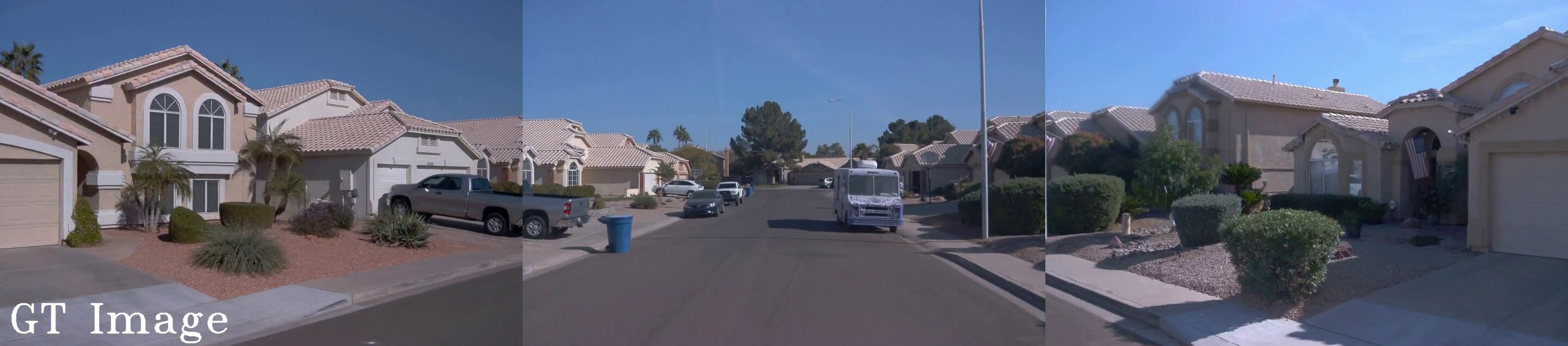} \\[-2pt]
		
		\raisebox{0.09\textwidth}[0pt][0pt]{\rotatebox[origin=c]{90}{Ours}}
		& \includegraphics[width=0.9\textwidth]{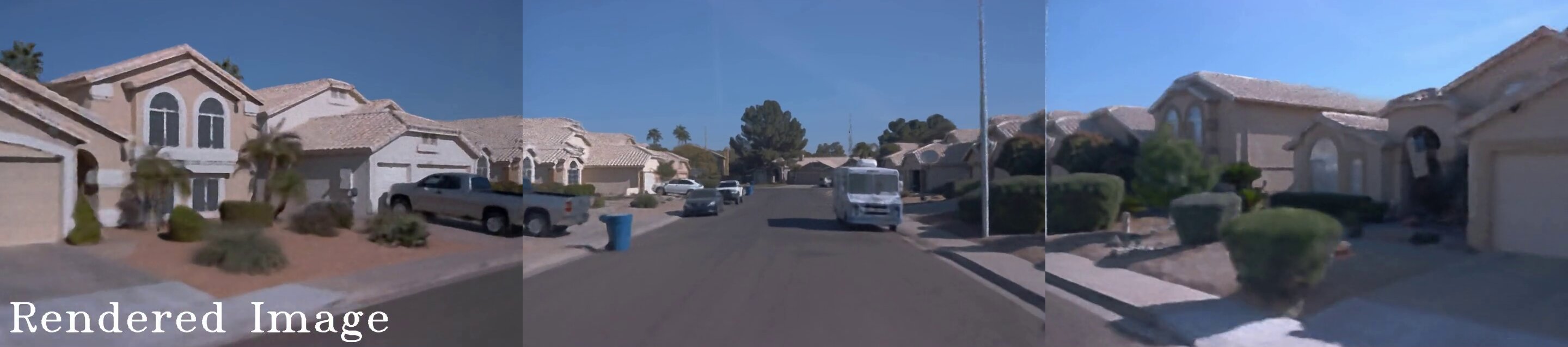} \\[-2pt]
		
		\raisebox{0.09\textwidth}[0pt][0pt]{\rotatebox[origin=c]{90}{NGP~\citep{muller2022instantngp}+L~\citep{rematas2022urban}}}
		& \includegraphics[width=0.9\textwidth]{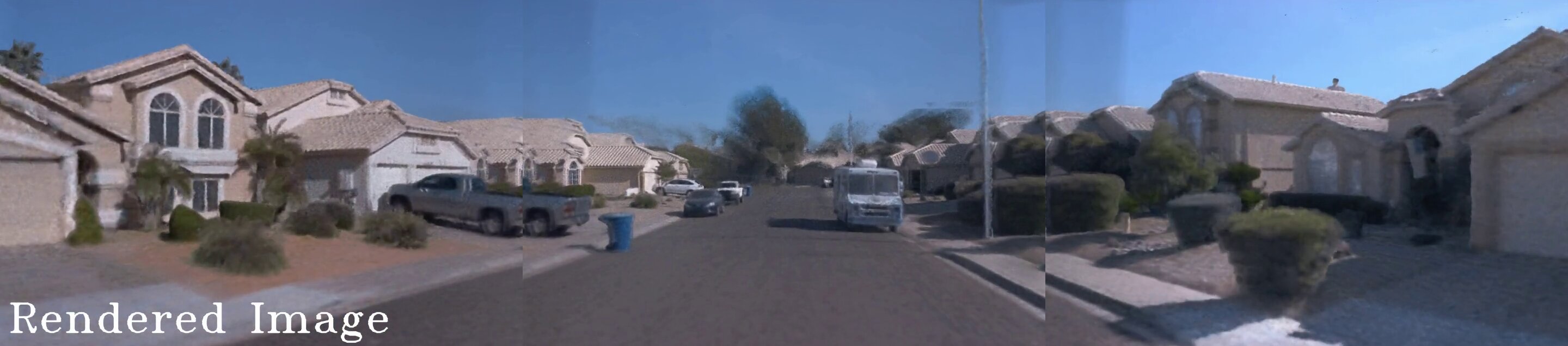} \\[-2pt]
		
		\raisebox{0.09\textwidth}[0pt][0pt]{\rotatebox[origin=c]{90}{Ours}}
		& \includegraphics[width=0.9\textwidth]{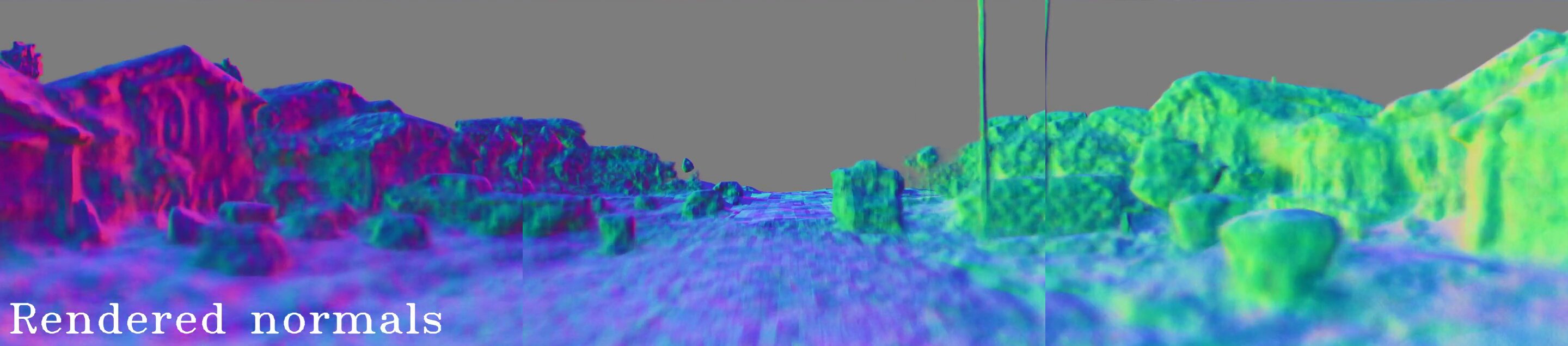} \\[-2pt]

		\raisebox{0.09\textwidth}[0pt][0pt]{\rotatebox[origin=c]{90}{Ours + Normal cues}}
		& \includegraphics[width=0.9\textwidth]{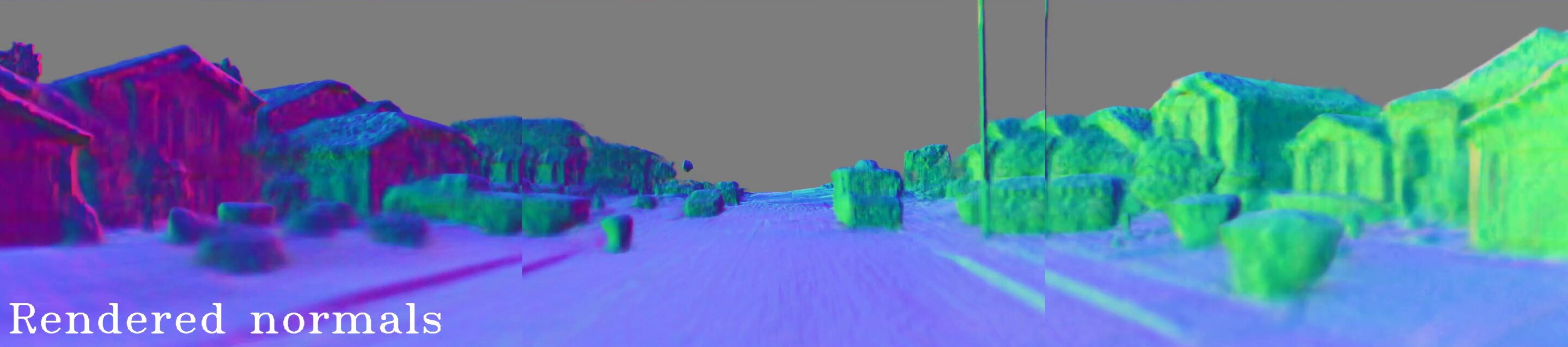} \\[-2pt]

		\raisebox{0.09\textwidth}[0pt][0pt]{\rotatebox[origin=c]{90}{NeuS~\citep{wang2021neus}}}
		& \includegraphics[width=0.9\textwidth]{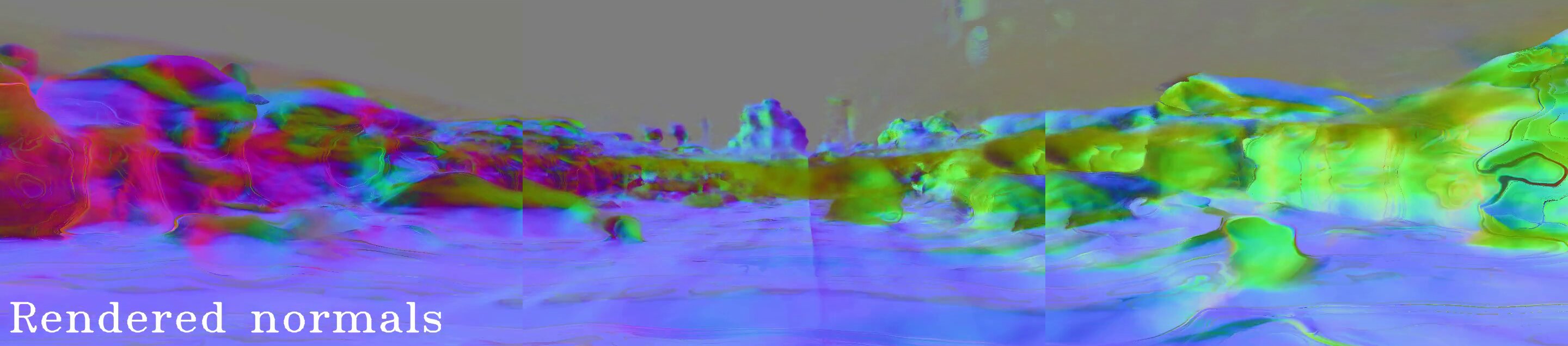} \\[-2pt]
		
		\raisebox{0.09\textwidth}[0pt][0pt]{\rotatebox[origin=c]{90}{Ours}}
		& \includegraphics[width=0.9\textwidth]{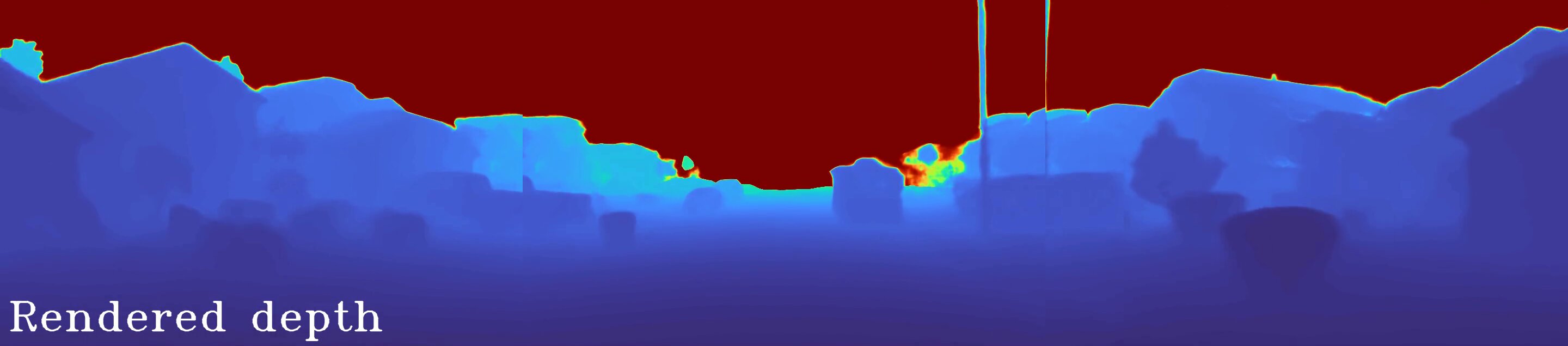} \\[-2pt]
		
		\raisebox{0.09\textwidth}[0pt][0pt]{\rotatebox[origin=c]{90}{NGP~\citep{muller2022instantngp}+L~\citep{rematas2022urban}}}
		& \includegraphics[width=0.9\textwidth]{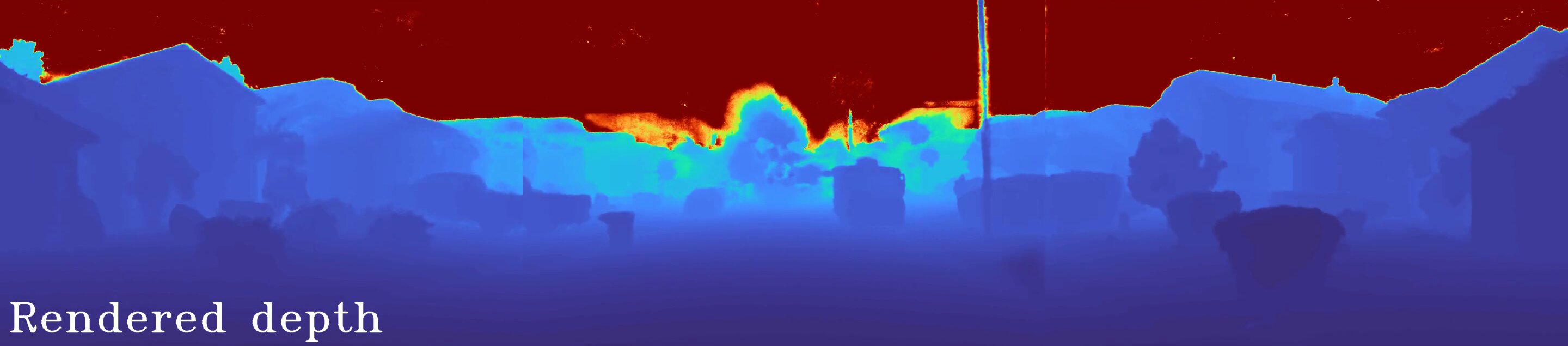}
		
	\end{tabular}
	\vspace{-5pt}
	\caption{Qualitative comparison of reconstruction with LiDAR, using all five LiDARs, on Waymo Open Dataset~\citep{waymo} seg1506235\dots.}
	\label{fig:supp:demo_withlidar_all3}
\end{figure}

\begin{table}
\centering
\setlength\tabcolsep{1pt}
\caption{Reconstruction without LiDAR on selected Waymo Open Dataset~\citep{waymo} sequences.}
\label{tab:supp:recon_withoutlidar}
\begin{tabular}{c||>{\columncolor{VeryLightPink}}c>{\columncolor{VeryLightPink}}c|>{\columncolor{VeryLightGreen}}c>{\columncolor{VeryLightGreen}}c>{\columncolor{VeryLightGreen}}c|>{\columncolor{VeryLightBlue}}c>{\columncolor{VeryLightBlue}}c>{\columncolor{VeryLightBlue}}c}

\toprule[1.5pt]

\multirow{2}{*}{Sequence} 
& \multicolumn{2}{c|}{PSNR$\uparrow$}  
& \multicolumn{3}{c|}{C-D$\downarrow$} 
& \multicolumn{3}{c}{RMSE$\downarrow$} 
\\

& \tiny{$\text{F}^2$-NeRF} & Ours
& {\tiny{COLMAP}} & {\tiny{$\text{F}^2$-NeRF}} & Ours 
& {\tiny{COLMAP}} & {\tiny{$\text{F}^2$-NeRF}} & Ours 
\\

\midrule[1pt]

seg1006130\dots & 24.41 & \textbf{27.39} & 99.01  & 70.90    & \textbf{0.88} & 7.10 & 8.87 &   \textbf{2.99} \\
seg1027514\dots & 22.14 & \textbf{27.58} & 127.19 & 295.85   & \textbf{0.81} & 7.47 & 16.52 &  \textbf{2.91} \\
seg1067626\dots & 26.43 & \textbf{29.01} & 235.26 & 906.56   & \textbf{1.58} & 9.06 & 35.59 &  \textbf{4.34} \\
seg1137922\dots & 25.42 & \textbf{28.33} & 241.81 & 398.55   & \textbf{2.67} & 12.39 & 20.10 & \textbf{5.70} \\
seg1172406\dots & 27.16 & \textbf{28.50} & 175.13 & 480.75   & \textbf{0.56} & 13.62 & 9.00 &  \textbf{2.57} \\
seg1287964\dots & 25.40 & \textbf{29.89} & 152.94 & 21.50    & \textbf{0.89} & 10.34 & 6.73 &  \textbf{3.19} \\
seg1308545\dots & 23.55 & \textbf{25.95} & 150.52 & 560.63   & \textbf{1.43} & 8.64 & 15.50 &  \textbf{4.12} \\
seg1314219\dots & 21.76 & \textbf{26.07} & 139.69 & 588.07   & \textbf{0.88} & 6.75 & 19.30 &  \textbf{3.48} \\
seg1319679\dots & 24.54 & \textbf{25.66} & 115.46 & 517.39   & \textbf{1.48} & 7.63 & 23.50 &  \textbf{4.76} \\
seg1323841\dots & 24.03 & \textbf{28.49} & 144.61 & 473.01   & \textbf{0.80} & 7.32 & 20.19 &  \textbf{3.13} \\
seg1347637\dots & 26.12 & \textbf{28.26} & 69.94  & 932.15   & \textbf{0.41} & 5.93 & 21.72 &  \textbf{1.84} \\
seg1400454\dots & 24.58 & \textbf{25.49} & 148.23 & 11196.87 & \textbf{1.09} & 8.08 & 39.85 &  \textbf{3.29} \\
seg1434813\dots & 26.15 & \textbf{28.23} & 111.42 & 1567.38  & \textbf{1.95} & 8.48 & 35.96 &  \textbf{4.74} \\
seg1442480\dots & 25.14 & \textbf{27.56} & 171.49 & 1111.72  & \textbf{0.84} & 7.85 & 36.35 &  \textbf{2.97} \\
seg1486973\dots & 22.87 & \textbf{24.42} & 89.58  & 9.85     & \textbf{0.57} & 5.52 & 3.53 &   \textbf{2.82} \\
seg1506235\dots & 25.07 & \textbf{27.41} & 132.43 & 638.32   & \textbf{0.67} & 7.84 & 27.61 &  \textbf{2.40} \\
seg1522170\dots & 25.71 & \textbf{26.29} & 145.69 & 388.61   & \textbf{2.11} & 11.28 & 16.66 & \textbf{4.87} \\
seg1527063\dots & 26.81 & \textbf{29.06} & 93.03  & 142.34   & \textbf{0.44} & 2.62 & 7.82 &   \textbf{1.98} \\
seg1534950\dots & 25.30 & \textbf{27.47} & 99.29  & 53.91    & \textbf{0.58} & 4.31 & 7.80 &   \textbf{2.56} \\
seg1536582\dots & 26.21 & \textbf{27.69} & 71.87  & 56.03    & \textbf{0.56} & 6.57 & 10.41 &  \textbf{2.47} \\
seg1586862\dots & 24.29 & \textbf{26.83} & 112.04 & 653.87   & \textbf{0.57} & 5.94 & 18.78 &  \textbf{2.60} \\
seg1634531\dots & 26.27 & \textbf{26.82} & 90.64  & 187.37   & \textbf{0.44} & 5.31 & 11.85 &  \textbf{2.23} \\
seg1647019\dots & 21.90 & \textbf{24.18} & 170.05 & 96.24    & \textbf{1.08} & 10.36 & 12.25 & \textbf{4.31} \\
seg1660852\dots & 21.37 & \textbf{24.82} & 102.14 & 4.21     & \textbf{0.98} & 5.11 & 4.72 &   \textbf{3.91} \\
seg1664636\dots & 25.23 & \textbf{26.83} & 97.62  & 265.26   & \textbf{0.52} & 6.54 & 13.86 &  \textbf{2.26} \\
seg1776195\dots & 24.95 & \textbf{26.71} & 234.98 & 540.94   & \textbf{1.32} & 14.52 & 25.24 & \textbf{3.90} \\
seg3224923\dots & 26.31 & \textbf{27.26} & 126.61 & 26.14    & \textbf{1.14} & 5.42 & 7.16 &   \textbf{3.53} \\
seg3425716\dots & 26.33 & \textbf{29.79} & 196.65 & 2673.64  & \textbf{0.69} & 18.81 & 30.68 & \textbf{3.00} \\
seg3988957\dots & 22.02 & \textbf{24.65} & 126.66 & 23.62    & \textbf{0.77} & 6.07 & 5.66 &   \textbf{3.30} \\
seg4058410\dots & 25.83 & \textbf{28.33} & 88.46  & 34.89    & \textbf{0.62} & 5.46 & 7.02 &   \textbf{2.62} \\
seg8811210\dots & 24.28 & \textbf{26.47} & 154.04 & 680.01   & \textbf{1.35} & 7.16 & 27.30 &  \textbf{3.83} \\
seg9385013\dots & 22.85 & \textbf{26.55} & 179.23 & 2780.03  & \textbf{1.87} & 9.10 & 49.34 &  \textbf{4.52} \\

\midrule[1pt]

Average & 24.70 & \textbf{27.12} & 137.30 & 886.77 & \textbf{1.02} & 8.08 & 18.65 & \textbf{3.35} \\

\bottomrule[1.5pt]

\end{tabular}
\end{table}

\begin{table}
\small
\setlength\tabcolsep{1pt}
\caption{Reconstruction with LiDAR on selected Waymo Open Dataset~\citep{waymo} sequences.}
\label{tab:supp:recon_withlidar}
\resizebox{\columnwidth}{!}{
\begin{tabular}{c||>{\columncolor{VeryLightPink}}c>{\columncolor{VeryLightPink}}c>{\columncolor{VeryLightPink}}c|>{\columncolor{VeryLightGreen}}c>{\columncolor{VeryLightGreen}}c>{\columncolor{VeryLightGreen}}c|>{\columncolor{VeryLightBlue}}c>{\columncolor{VeryLightBlue}}c>{\columncolor{VeryLightBlue}}c||>{\columncolor{VeryLightPink}}c>{\columncolor{VeryLightPink}}c|>{\columncolor{VeryLightGreen}}c>{\columncolor{VeryLightGreen}}c|>{\columncolor{VeryLightBlue}}c>{\columncolor{VeryLightBlue}}c}

\toprule[1.5pt]

& \multicolumn{9}{c||}{Using 1 dense + 4 sparse LiDARs} & \multicolumn{6}{c}{Using 4 sparse LiDARs} 
\\

\midrule[1pt]

\multirow{2}{*}{Sequence} 
& \multicolumn{3}{c|}{PSNR$\uparrow$}  
& \multicolumn{3}{c|}{C-D$\downarrow$} 
& \multicolumn{3}{c||}{RMSE$\downarrow$} 
& \multicolumn{2}{c|}{PSNR$\uparrow$} 
& \multicolumn{2}{c|}{C-D$\downarrow$} 
& \multicolumn{2}{c}{RMSE$\downarrow$} 
\\

& NeuS & \tiny{NGP+L} & Ours 
& NeuS & \tiny{NGP+L} & Ours 
& NeuS & \tiny{NGP+L} & Ours 
& \tiny{NGP+L} & Ours 
& \tiny{NGP+L} & Ours 
& \tiny{NGP+L} & Ours 
\\

\midrule[1pt]
	
seg1006130\dots & 14.57 & 25.15 & \textbf{27.38} & 0.68 & 0.19 & \textbf{0.13} & 3.14 & 1.67 & \textbf{1.22} & 23.86 & \textbf{27.26} & \textbf{0.40} & 0.42 & 3.22 & \textbf{2.58} \\
seg1027514\dots & ~9.10 & 23.92 & \textbf{27.19} & 0.88 & 0.44 & \textbf{0.25} & 3.30 & 1.94 & \textbf{1.40} & 23.48 & \textbf{27.11} & 0.69 & \textbf{0.66} & 3.25 & \textbf{2.82} \\
seg1067626\dots & 12.90 & 25.94 & \textbf{27.68} & 1.11 & 0.58 & \textbf{0.35} & 4.40 & 2.61 & \textbf{1.94} & 24.87 & \textbf{28.68} & 3.03 & \textbf{1.50} & 7.61 & \textbf{4.12} \\
seg1137922\dots & 13.75 & 25.93 & \textbf{27.64} & 1.23 & 0.57 & \textbf{0.33} & 5.77 & 2.56 & \textbf{2.14} & 25.15 & \textbf{27.80} & 6.60 & \textbf{3.27} & 9.62 & \textbf{5.91} \\
seg1172406\dots & 21.28 & 26.16 & \textbf{27.86} & 0.40 & 0.17 & \textbf{0.13} & 2.18 & 1.51 & \textbf{1.03} & 24.89 & \textbf{28.21} & 0.33 & \textbf{0.25} & 2.80 & \textbf{1.83} \\
seg1287964\dots & 10.64 & 26.31 & \textbf{29.92} & 0.63 & 0.20 & \textbf{0.16} & 3.34 & 1.64 & \textbf{1.32} & 25.98 & \textbf{29.45} & 2.29 & \textbf{1.31} & 6.79 & \textbf{3.50} \\
seg1308545\dots & 16.57 & 24.22 & \textbf{25.59} & 0.95 & 0.25 & \textbf{0.20} & 4.09 & 2.03 & \textbf{1.59} & 22.99 & \textbf{25.61} & 4.43 & \textbf{2.98} & 7.85 & \textbf{4.40} \\
seg1314219\dots & 10.78 & 22.85 & \textbf{25.77} & 0.77 & 0.37 & \textbf{0.22} & 3.52 & 2.14 & \textbf{1.53} & 22.18 & \textbf{25.66} & \textbf{1.26} & 1.59 & 6.13 & \textbf{3.94} \\
seg1319679\dots & 10.71 & 24.69 & \textbf{24.79} & 1.10 & 0.25 & \textbf{0.17} & 4.97 & 2.15 & \textbf{1.68} & 23.78 & \textbf{25.30} & 2.45 & \textbf{2.36} & 7.94 & \textbf{5.21} \\
seg1323841\dots & ~8.59 & 24.13 & \textbf{28.16} & 0.83 & 0.29 & \textbf{0.18} & 3.64 & 1.75 & \textbf{1.37} & 23.37 & \textbf{28.16} & 0.69 & \textbf{0.50} & 3.70 & \textbf{2.66} \\
seg1347637\dots & 14.19 & 26.18 & \textbf{28.20} & 0.39 & 0.09 & \textbf{0.08} & 2.08 & 1.03 & \textbf{0.84} & 25.30 & \textbf{28.18} & 0.17 & \textbf{0.15} & 1.84 & \textbf{1.41} \\
seg1400454\dots & 11.43 & 24.73 & \textbf{24.79} & 0.80 & 0.29 & \textbf{0.19} & 3.22 & 1.81 & \textbf{1.38} & 23.96 & \textbf{25.40} & 0.87 & \textbf{0.76} & 4.35 & \textbf{2.91} \\
seg1434813\dots & 10.71 & 26.31 & \textbf{26.98} & 1.67 & \textbf{0.22} & 0.25 & 4.59 & 2.12 & \textbf{1.81} & 25.43 & \textbf{27.75} & 4.40 & \textbf{0.97} & 8.71 & \textbf{4.30} \\
seg1442480\dots & 13.33 & 25.16 & \textbf{27.22} & 0.70 & 0.31 & \textbf{0.20} & 3.30 & 1.70 & \textbf{1.42} & 24.28 & \textbf{27.28} & 0.81 & \textbf{0.73} & 3.69 & \textbf{2.77} \\
seg1486973\dots & 16.23 & 23.59 & \textbf{24.21} & 0.52 & 0.20 & \textbf{0.15} & 2.90 & 1.52 & \textbf{1.22} & 21.99 & \textbf{24.20} & \textbf{0.31} & 0.36 & 2.89 & \textbf{2.17} \\
seg1506235\dots & 12.01 & 25.07 & \textbf{27.54} & 0.66 & 0.24 & \textbf{0.13} & 2.67 & 1.43 & \textbf{1.04} & 24.12 & \textbf{27.08} & 0.64 & \textbf{0.39} & 3.22 & \textbf{1.89} \\
seg1522170\dots & 12.67 & 25.26 & \textbf{25.66} & 0.99 & 0.33 & \textbf{0.18} & 4.79 & 2.14 & \textbf{1.68} & 24.77 & \textbf{25.67} & 3.46 & \textbf{1.84} & 6.87 & \textbf{4.87} \\
seg1527063\dots & 17.46 & 27.04 & \textbf{28.95} & 0.28 & 0.11 & \textbf{0.07} & 1.92 & 1.04 & \textbf{0.76} & 26.03 & \textbf{28.76} & 0.31 & \textbf{0.24} & 2.46 & \textbf{1.55} \\
seg1534950\dots & 12.35 & 25.42 & \textbf{27.26} & 0.36 & \textbf{0.09} & 0.11 & 2.30 & 1.34 & \textbf{1.06} & 24.58 & \textbf{27.20} & 0.37 & \textbf{0.18} & 2.77 & \textbf{1.94} \\
seg1536582\dots & 17.03 & 26.35 & \textbf{27.56} & 0.34 & 0.20 & \textbf{0.11} & 2.42 & 1.52 & \textbf{1.07} & 24.96 & \textbf{27.43} & \textbf{0.24} & 0.27 & 2.72 & \textbf{1.96} \\
seg1586862\dots & 13.35 & 24.26 & \textbf{26.71} & 0.57 & 0.21 & \textbf{0.13} & 3.28 & 1.63 & \textbf{1.21} & 23.20 & \textbf{26.68} & 0.54 & \textbf{0.49} & 3.37 & \textbf{2.44} \\
seg1634531\dots & 17.38 & 25.68 & \textbf{26.57} & 0.37 & 0.10 & \textbf{0.10} & 2.38 & 1.14 & \textbf{0.99} & 24.32 & \textbf{26.50} & \textbf{0.17} & 0.23 & 2.21 & \textbf{1.85} \\
seg1647019\dots & 11.35 & \textbf{22.67} & 21.26 & 1.24 & 0.36 & \textbf{0.29} & 5.32 & \textbf{2.42} & 3.18 & 21.98 & \textbf{23.88} & 1.28 & \textbf{1.23} & 6.08 & \textbf{4.38} \\
seg1660852\dots & 15.75 & 23.14 & \textbf{24.58} & 0.65 & 0.24 & \textbf{0.22} & 3.68 & 2.13 & \textbf{1.82} & 22.23 & \textbf{24.47} & \textbf{0.73} & 0.78 & 3.51 & \textbf{3.49} \\
seg1664636\dots & 16.34 & 24.87 & \textbf{26.86} & 0.63 & 0.22 & \textbf{0.13} & 2.65 & 1.36 & \textbf{1.03} & 23.99 & \textbf{26.71} & 0.62 & \textbf{0.20} & 2.36 & \textbf{1.79} \\
seg1776195\dots & ~8.55 & 24.70 & \textbf{25.35} & 1.13 & 0.47 & \textbf{0.31} & 4.34 & 2.14 & \textbf{1.89} & 24.30 & \textbf{26.37} & 1.70 & \textbf{1.03} & 5.94 & \textbf{3.87} \\
seg3224923\dots & 17.36 & 25.97 & \textbf{27.20} & 0.75 & 0.43 & \textbf{0.20} & 3.27 & 1.89 & \textbf{1.43} & 24.91 & \textbf{27.20} & \textbf{0.68} & 0.80 & 3.23 & \textbf{2.86} \\
seg3425716\dots & ~8.48 & 27.03 & \textbf{29.42} & 0.58 & 0.43 & \textbf{0.19} & 3.05 & 2.01 & \textbf{1.38} & 26.18 & \textbf{29.60} & 1.80 & \textbf{0.56} & 6.85 & \textbf{2.93} \\
seg3988957\dots & 11.70 & 22.62 & \textbf{24.19} & 0.75 & 0.28 & \textbf{0.19} & 3.43 & 1.87 & \textbf{1.52} & 21.75 & \textbf{24.25} & 0.99 & \textbf{0.64} & 4.18 & \textbf{3.00} \\
seg4058410\dots & 12.14 & 26.40 & \textbf{28.08} & 0.48 & 0.15 & \textbf{0.10} & 2.93 & 1.31 & \textbf{1.04} & 25.03 & \textbf{28.09} & \textbf{0.39} & 0.48 & 3.13 & \textbf{2.18} \\
seg8811210\dots & 11.18 & 24.87 & \textbf{26.60} & 0.72 & \textbf{0.19} & 0.21 & 3.07 & 1.44 & \textbf{1.31} & 23.69 & \textbf{26.49} & \textbf{0.68} & 0.70 & 3.92 & \textbf{2.64} \\
seg9385013\dots & 13.80 & 24.16 & \textbf{26.07} & 1.20 & 0.58 & \textbf{0.27} & 4.75 & 2.39 & \textbf{1.67} & 23.23 & \textbf{25.92} & \textbf{3.02} & 4.94 & 7.47 & \textbf{6.28} \\

\midrule[1pt]

Average & 13.24 & 25.02 & \textbf{26.66} & 0.76 & 0.28 & \textbf{0.19} & 3.46 & 1.79 & \textbf{1.44} & 24.09 & \textbf{26.85} & 1.45 & \textbf{0.90} & 4.71 & \textbf{3.04} \\

\bottomrule[1.5pt]

\end{tabular}
}
\vspace{-20pt}
\end{table}

\subsection{Street-view reconstruction with LiDAR}

For reconstruction with LiDAR of Waymo Open Dataset~\citep{waymo}, we compare our reconstruction method using LiDAR with two baselines.
The first is our basis implicit surface reconstruction method, NeuS~\citep{wang2021neus}. We deepen and widen its MLP~(Multi Layer Perceptron) to 10 layers with 512 neurons, and increase its ray batch size from 512 to 2048.
The second one is the combination of NGP~\citep{muller2022instantngp} with the line-of-sight and depth losses from  UrbanNeRF~\citep{rematas2022urban} for leveraging LiDAR supervision, dubbed "NGP+L". 

We conduct experiments in two different settings. 
In the first setting, we only employ the four relatively sparser auxiliary LiDARs excluding the top LiDAR, while using the top LiDAR data for evaluation. 
%
In the second setting, we utilize all five LiDARs, including the top LiDAR data as well as four additional auxiliary LiDARs, and evaluate using the top LiDAR. 
Example qualitative comparisons are shown in Fig.,~\ref{fig:supp:demo_withlidar_all2},~\ref{fig:supp:demo_withlidar_all3}.
Detailed quantitative comparisons are shown in Tab.~\ref{tab:supp:recon_withlidar}. 
\section{Implementation Details}
\label{supp:implementation_details}

\subsection{Volume rendering of close-range implicit surfaces}
\label{supp:impl:neus}

Following NeuS~\citep{wang2021neus}, we use SDF~(Signed Distance Function) fields for the geometry of close-range scene and map the SDF values $\widehat{\mathcal{S}}_i$ of sample points $\{\mathbf{x}^{(\text{cr})}_i=\mathbf{o}+t_i^{(\text{cr})} \mathbf{v}\}$ to opacity $\alpha_i^{(\text{cr})}$ in order to apply volume rendering:

\begin{equation}
\alpha^{(\text{cr})}_{i} = \max
\left(
\frac{
		\Phi_{s}( \widehat{\mathcal{S}}_i) ) - \Phi_{s}( \widehat{\mathcal{S}}_{i+1} )
	}{
		\Phi_{s}( \widehat{\mathcal{S}}_i )
}, 0
\right)
\end{equation}

where $s$ is a learnable scaling parameter of the Sigmoid function $\Phi_{s}(x)=(1+e^{-s\cdot x})^{-1}$. This mapping strategy ensures unbiased calculation of color contribution (i.e. visibility weights) while respecting occlusion. For more details, please refer to NeuS~\citep{wang2021neus}.

In practice, it is difficult for $s$ to converge to a sufficient extent when training without LiDAR data. In this case, we manually adjust $s$ to increase linearly after the initial training stage, which is typically around 5k iterations.

\subsection{Optional sky masks}

Previous research studying multi-view neural reconstruction \citep{sun2022neural-recon-w,li2022voxsurf} or novel view synthesis \citep{rematas2022urban,tancik2022blocknerf} on outdoor scenes typically require mask annotation or segmentation that marks sky pixels in order to penalize non-occupied areas.
Thanks to the disentangled design of our close-range and distant-view models, our approach works effectively with or without sky mask inputs.

\begin{figure}
\centering
\setlength\tabcolsep{1pt}
\begin{tabular}{c@{\hskip 0.1cm}|@{\hskip 0.1cm}c@{\hskip 0.1cm}c@{\hskip 0.1cm}c}
	
\includegraphics[width=0.3\textwidth]{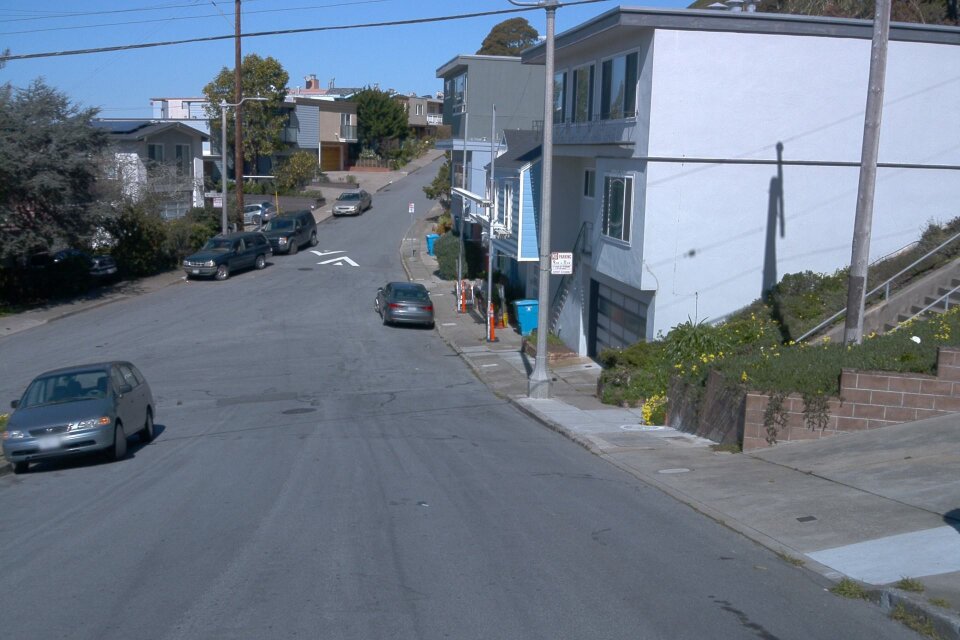}
& 
\raisebox{0.093\textwidth}[0pt][0pt]{\rotatebox[origin=c]{90}{\small{With mask and sky}}}
&
\includegraphics[width=0.3\textwidth]{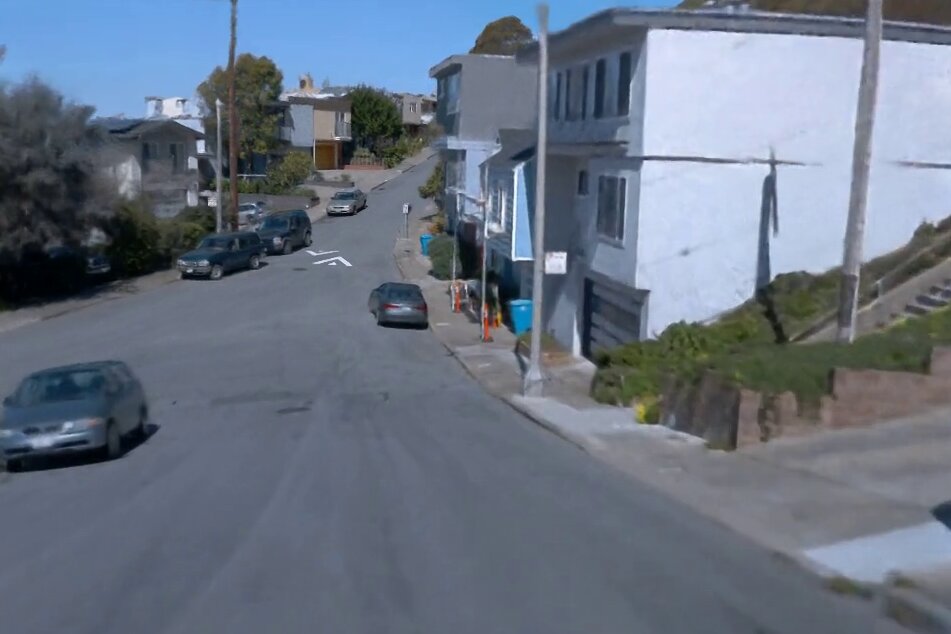}
& 
\includegraphics[width=0.3\textwidth]{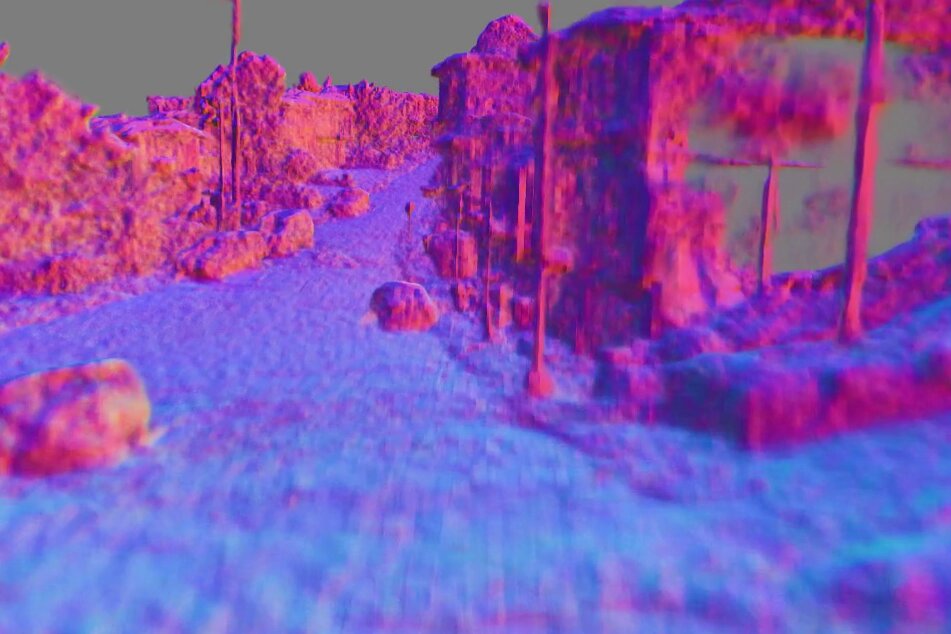}
\\
GT image & & Rendered image & Rendered normals \\

& & & \\

\includegraphics[width=0.3\textwidth]{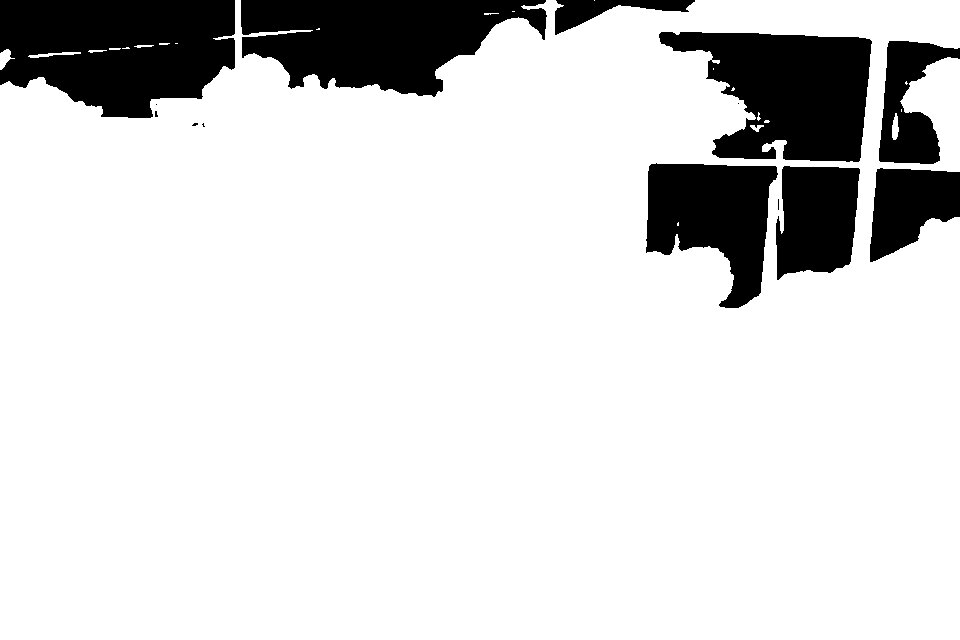}
& 
\raisebox{0.093\textwidth}[0pt][0pt]{\rotatebox[origin=c]{90}{\small{Without mask or sky}}}
&
\includegraphics[width=0.3\textwidth]{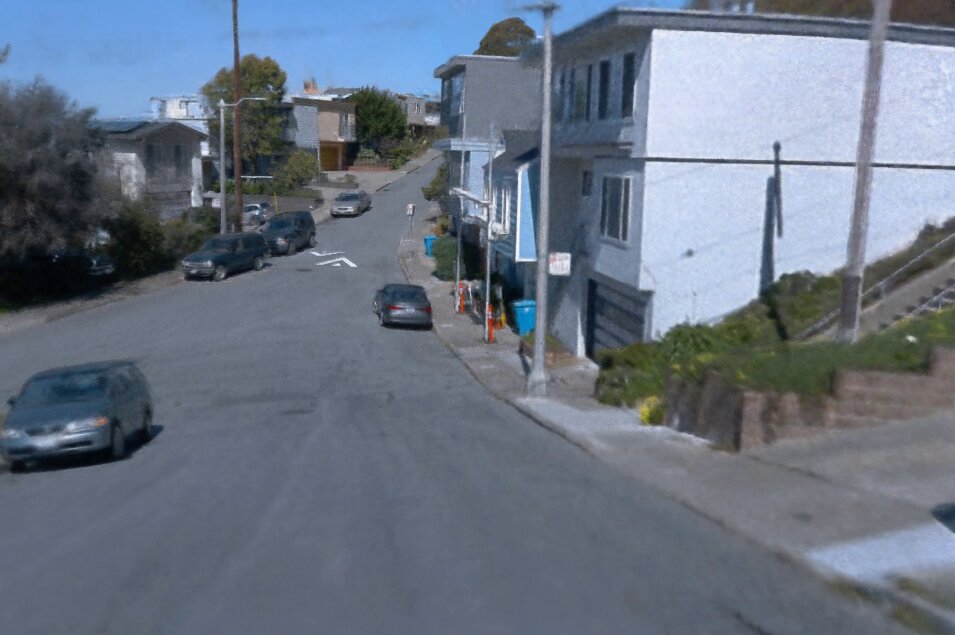}
& 
\includegraphics[width=0.3\textwidth]{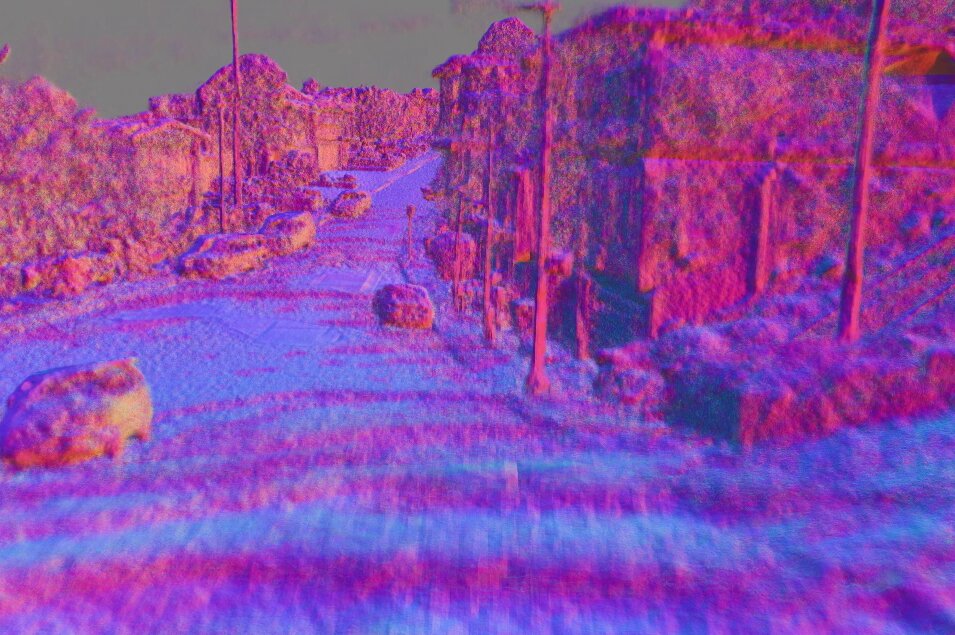}
\\
Estimated occupancy mask & & Rendered image & Rendered normals \\
\end{tabular}
\caption{Example illustration of the influence of incorrectly inferred sky masks on reconstruction (top row) and the benefit of reconstruction without masks (bottom row) to avoid such segmentation errors. Experiments are conducted on Waymo Open Dataset~\citep{waymo} seg1664636\dots. The bottom-left figure demonstrates an incorrect occupancy mask inferred using Segformer~\citep{xie2021segformer}, which has mistakenly classified the white wall as sky.}
\label{fig:supp:nomask}
\end{figure}

As shown in Fig.~\ref{fig:supp:nomask}, getting rid of mask dependency is beneficial because the sky masks produced by segmentation models are often noisy and may incorrectly include foreground elements.

If sky masks $M^{(\text{sky})}$ are provided, we explicit distinguish the sky model out of distant-view. We use $M^{(\text{sky})}$ to supervise the rendered occupancy $\widehat{O}^{(\text{cr,dv})}$ of the joint close-range and distant-view models 
as shown in Equ.~\ref{equ:mask_loss}, 
where BCE stands for the binary cross entropy loss.
It is worth noting that the distant-view model is still necessary even with the use of sky masks and the sky model. This is because the close-range scene we defined only applies to objects within a few hundred meters, while a distant-view model is required to represent landscapes as well as objects that extend beyond the end of the capture trajectory.

\begin{equation}
\label{equ:mask_loss}
\mathcal{L}_{\text{mask}}=
\begin{cases}
	\text{BCE}\left( \widehat{O}^{(\text{cr,dv})}(\mathbf{r}), 1-M^{(\text{sky})}(\mathbf{r}) \right) & \text{with sky} \\
	0 & \text{without sky}
\end{cases}
\end{equation}

If sky masks are not provided, disentanglement of distant view and the sky is found to be difficult. Therefore, we let the distant-view model to inclusively represent the sky by setting a large step at the end of marching interval for each ray. In this case, the close-range and distant-view models can still be disentangled without extra supervision. 


In practice, distinguishing the sky model out of the distant-view model is preferred due to its better performance. 
If sky annotations are not provided, we use SegFormer~\citep{xie2021segformer} to efficiently and automatically extract sky masks.

%

\subsection{Ray marching with multi-stage hierarchical sampling based on occupancy information}

\begin{figure}[htbp]
	\includegraphics[width=\textwidth]{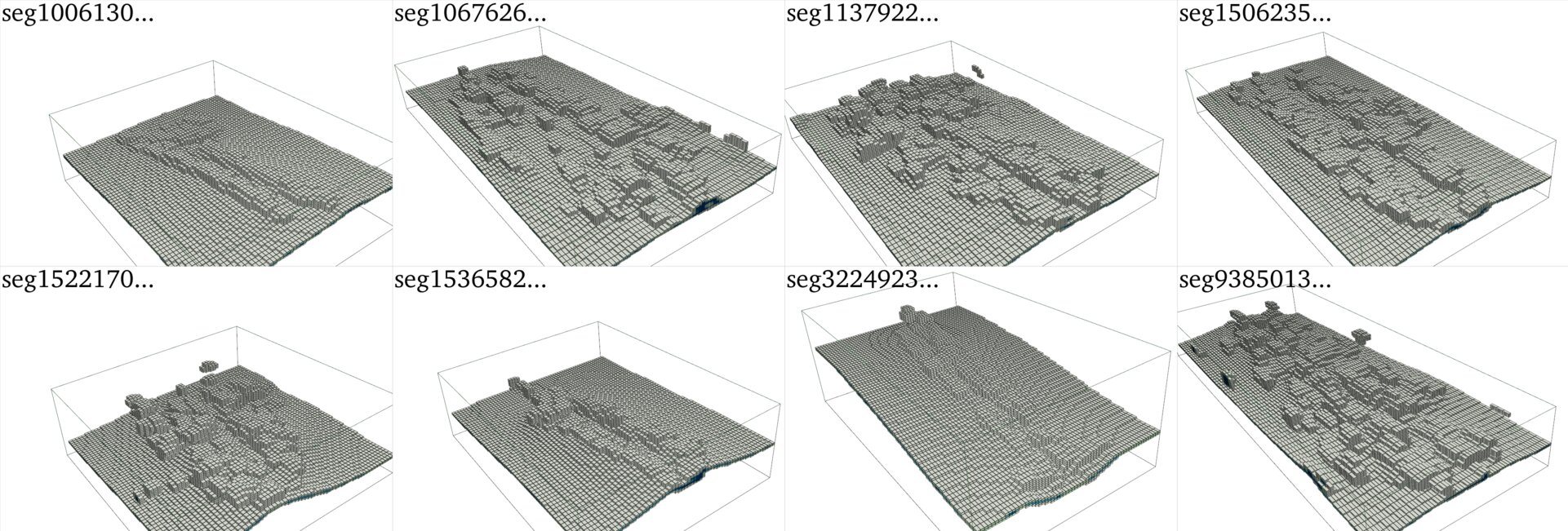}
	\caption{Illustration of the updated occupancy information for ray marching at the end of the training step of each data sequence. The cuboid wire-frame indicates the AABB~(Axis-Aligned-Bounding-Box) of the close-range scene.}
	\label{fig:supp:occgrid}
\end{figure}

We propose a ray marching strategy that apply multi-stage hierarchical sampling to the marched ray samples of the occupancy grid. 
Examples of the updated occupancy information at the end of training stage are shown in Fig.~\ref{fig:supp:occgrid}.

\begin{figure}[htbp]
	\centering
	\setlength\tabcolsep{1pt}
	\begin{tabular}{cc@{\hskip 0.1cm}c@{\hskip 0.1cm}c@{\hskip 0.1cm}c@{\hskip 0.1cm}c}
		\raisebox{0.07\textwidth}[0pt][0pt]{\rotatebox[origin=c]{90}{seg1137922\dots}}
		& \includegraphics[width=0.19\textwidth]{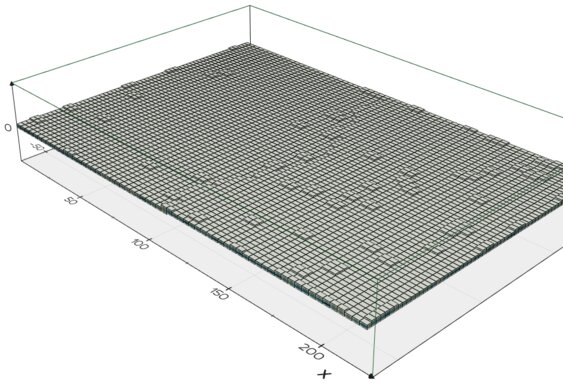}
		& \includegraphics[width=0.19\textwidth]{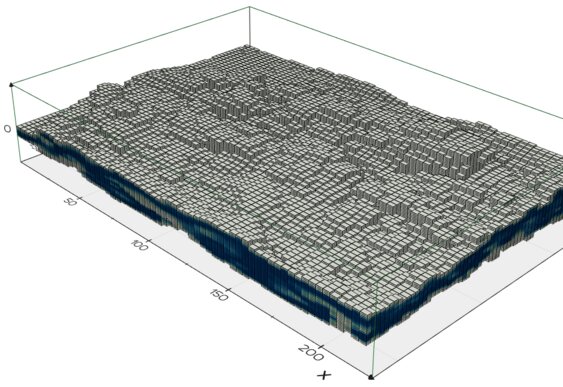}
		& \includegraphics[width=0.19\textwidth]{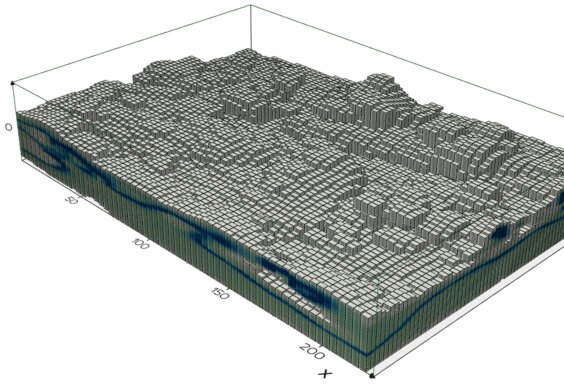}
		& \includegraphics[width=0.19\textwidth]{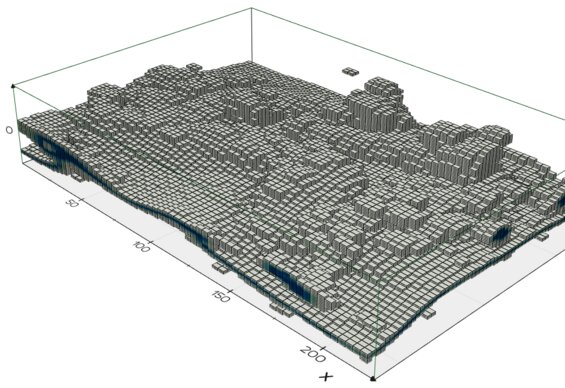}
		& \includegraphics[width=0.19\textwidth]{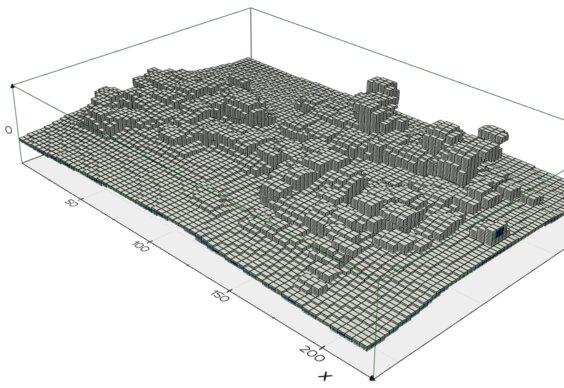}
		\\
		& 0 iter & 400 iter & 1.2k iter & 1.6k iter & Final \\
		
		\raisebox{0.07\textwidth}[0pt][0pt]{\rotatebox[origin=c]{90}{seg1006130\dots}}
		& \includegraphics[width=0.19\textwidth]{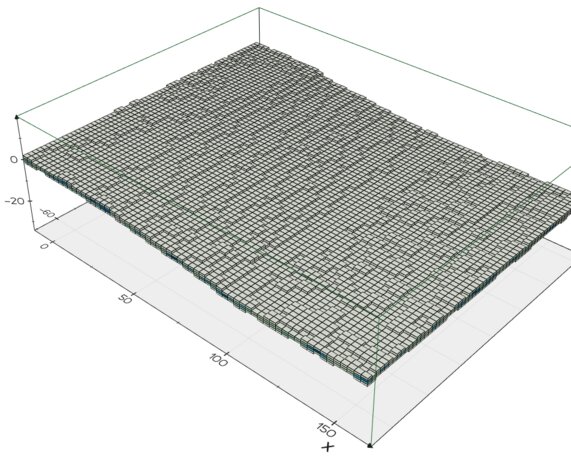}
		& \includegraphics[width=0.19\textwidth]{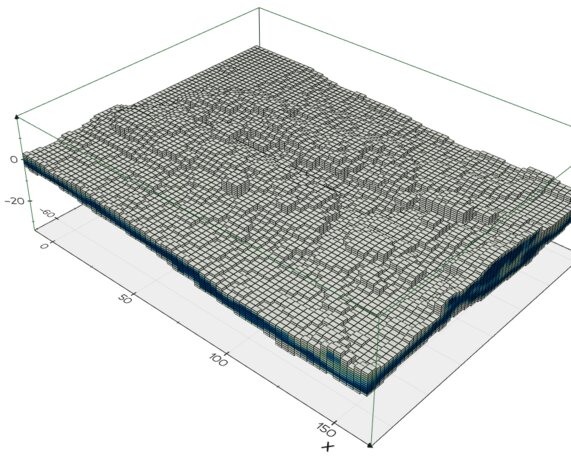}
		& \includegraphics[width=0.19\textwidth]{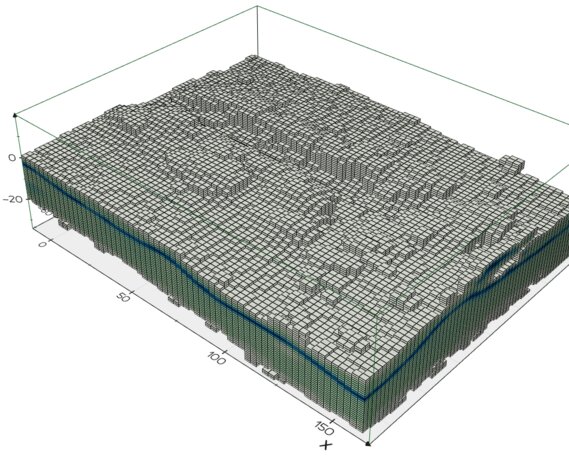}
		& \includegraphics[width=0.19\textwidth]{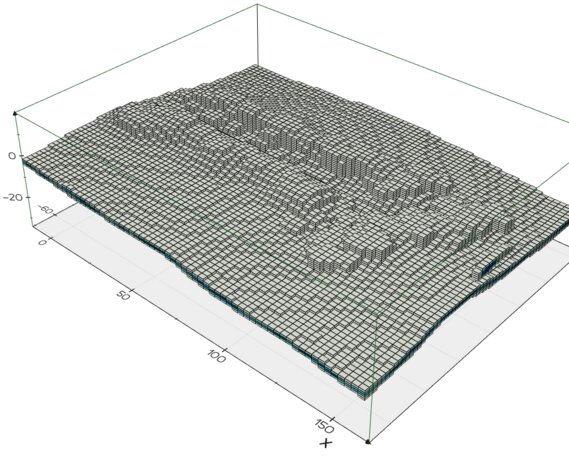}
		& \includegraphics[width=0.19\textwidth]{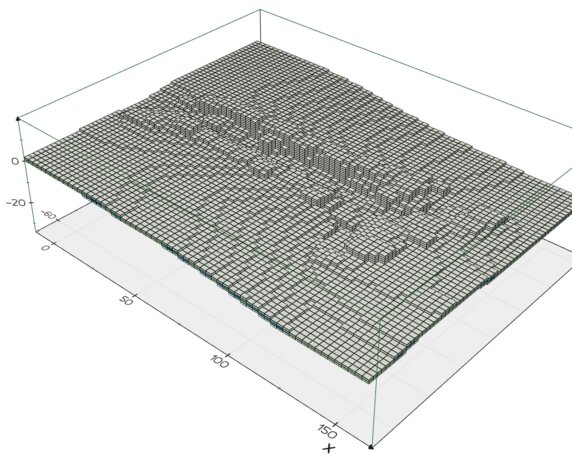}
		\\
		& 0 iter & 400 iter & 1.2k iter & 1.6k iter & Final 
		
	\end{tabular}
	\caption{Bootstrap update procedure of occupancy information for ray marching.}
	\label{fig:supp:occgrid_update}
\end{figure}

Previous approach studying multi-view reconstruction of outdoor scenes require either reconstructing the scene in advance~\citep{sun2022neural-recon-w} with COLMAP~\citep{schoenberger2016sfm} or feeding LiDAR pointclouds to pre-compute occupied voxels~\citep{li2022voxsurf}.
Following NGP~\citep{muller2022instantngp}, we update the occupancy information in a bootstrap manner and do not rely on any geometry information. Examples of the update process of occupancy information is shown in Fig.~\ref{fig:supp:occgrid_update}.

\subsection{Additional details}

We use a batch size of 8192 rays with up to $48(\text{cr})+64(\text{dv})$ meaningful samples on each ray and conduct meaningful sample reduction as~\citep{muller2022instantngp,li2022nerfacc} to ensure efficiency. The optimization takes 15k iterations for reconstruction with LiDAR data and 12k for without LiDAR data, both of which takes one hour or two on a single RTX3090 GPU.

To further stabilize the disentanglement of close-range and distant-view, we follow \citep{wang2022neus2} and introduce an annealing binary mask on different resolutions of the close-range hash-grid that gradually unleashes higher resolutions at the initial phase of training.

Image pixels are importance-sampled based on periodically accumulated and decaying error maps following NGP~\citep{muller2022instantngp}. All codes are implemented under PyTorch~\citep{pytorch} with certain atomic or pack-like operations implemented under Pytorch-CUDA extension.

\subsection{Geometry evaluation metrics}

Ground truth 3D scene geometry is often unavailable or difficult to obtain for street views.
Nonetheless, the LiDAR sensor in the autonomous driving datasets that we focus on can provide indirect 3D geometry information.
Despite the potential for noise and inaccuracy in LiDAR sensors, having a metric is still preferable to having none at all. In practice, we find that the LiDAR sensor data in Waymo-perception dataset~\citep{waymo} is of relatively high quality and can serve as a reliable evaluation basis.
Therefore, we propose a set of metrics to evaluate the reconstruction accuracy based on the comparison of the ground truth LiDAR sensor data with the rendered results on the provided LiDAR beams.

LiDAR sensors can be viewed as sparse depth sensors if we only use the first range return.
Following common practice of depth estimation~\citep{godard2019digging}, we use the usual metrics RMSE~(Root mean square error).

Depth metrics are sensitive for LiDAR beams closing to the edge of object boundaries. 
Hence, we also use the Chamfer Distance (C-D)\footnote{We utilize the implementation of Pytorch3D~\citep{pytorch3d}.} between the rendered and original LiDAR point clouds $\widehat{G},G$:

\begin{equation}
	\label{equ:chamfer_dis}
	\text{C-D}(\widehat{G},G) = 
	\frac{1}{|\widehat{G}|} \sum_{\mathbf{x} \in \widehat{G}} \min_{\mathbf{y} \in G} \lVert \mathbf{x}-\mathbf{y} \rVert_2^2 +
	\frac{1}{|G|} \sum_{\mathbf{y} \in G} \min_{\mathbf{x} \in \widehat{G}} \lVert \mathbf{y}-\mathbf{x} \rVert_2^2
\end{equation}

In practice, we have found that LiDAR sensors may produce some outlier observation points that significantly affect the calculation of the average chamfer distance. 
Therefore, we limit the calculation of chamfer distance to the first closest 97\% of ground truth points and discard the remaining 3\%. 


\end{document}